\documentclass[10pt,journal,compsoc]{IEEEtran}
\ifCLASSOPTIONcompsoc
  \usepackage[nocompress]{cite}
\else
  \usepackage{cite}
\fi

\ifCLASSINFOpdf
\else
\fi

\usepackage{xcolor}
\usepackage{times}
\usepackage{epsfig}
\usepackage{graphicx}
\usepackage{amsmath}
\usepackage{amssymb}
\usepackage{bbm}
\usepackage{caption}
\usepackage{subcaption}

\usepackage{amssymb}
\usepackage{pifont}
\newcommand{\cmark}{\ding{51}}%
\newcommand{\xmark}{\ding{55}}

\usepackage{arydshln}
\usepackage{xcolor}
\usepackage{tcolorbox}

\usepackage{float}

\usepackage{stfloats} 

\usepackage{dsfont}
\usepackage{subcaption}
\usepackage{caption}
\usepackage{nicefrac}
\usepackage{mathrsfs}
\usepackage{xcolor}
\usepackage{enumitem}
\usepackage{colortbl}
\usepackage{multirow}

\usepackage{marvosym}
\usepackage[export]{adjustbox}

\usepackage{pifont}

\definecolor{sh_gray}{rgb}{0.84,0.84,0.84}
\definecolor{sh_gray2}{rgb}{1,0.89,0.75}
\definecolor{color3}{rgb}{0.95,0.95,0.95}
\definecolor{color4}{rgb}{0.94,0.94,1}
\definecolor{color6}{rgb}{1,1,1}
\definecolor{color5}{rgb}{1,0.96,0.88}
\begin{document}

\title{Spatially-Attentive Patch-Hierarchical Network with Adaptive Sampling for Motion Deblurring}
%
%
%
%

\author{Maitreya Suin,
        Kuldeep Purohit,
        and  A. N. Rajagopalan,~\IEEEmembership{Senior Member,~IEEE}
}

\markboth{Journal of \LaTeX\ Class Files,~Vol.~14, No.~8, August~2015}%
{Shell \MakeLowercase{\textit{et al.}}: Bare Demo of IEEEtran.cls for Computer Society Journals}
\IEEEtitleabstractindextext{%
\begin{abstract}
This paper tackles the problem of motion deblurring of dynamic scenes. Although end-to-end fully convolutional designs have recently advanced the state-of-the-art in non-uniform motion deblurring, their performance-complexity trade-off is still sub-optimal. Most existing approaches achieve a large receptive field by increasing the number of generic convolution layers and kernel size. In this work, we propose a pixel adaptive and feature attentive design for handling large blur variations across different spatial locations and process each test image adaptively. We design a content-aware global-local filtering module that significantly improves performance by considering not only global dependencies but also by dynamically exploiting neighboring pixel information. We further introduce a pixel-adaptive non-uniform sampling strategy that implicitly discovers the difficult-to-restore regions present in the image and, in turn, performs fine-grained refinement in a progressive manner. Extensive qualitative and quantitative comparisons with prior art on deblurring benchmarks demonstrate that our approach performs favorably against the state-of-the-art deblurring algorithms.
\end{abstract}

}

\maketitle
\IEEEdisplaynontitleabstractindextext

\IEEEpeerreviewmaketitle

\IEEEraisesectionheading{\section{Introduction}\label{sec:introduction}}
\IEEEPARstart{I}{mage deblurring} is the task of learning a mapping function that transforms input blurred images into enhanced images devoid of blur-related artifacts and deformities. It is an ill-posed inverse problem due to the existence of many possible solutions. Relative motion during the sensor exposure period forms motion-blurred images. Such degradations impact not only human perceptibility but also deteriorate many standard computer vision applications, such as detectors, trackers, autonomous driving, etc. Blind motion deblurring aims to recover a sharp image from a given image degraded due to motion-induced smearing of texture and high-frequency components. Due to its diverse applications in surveillance, remote sensing, and cameras mounted on hand-held and vehicle-mounted cameras, deblurring has gathered substantial attention from computer vision and image processing communities in the past two decades.

Majority of traditional deblurring approaches are based on variational models, whose key component is the regularization term. The restoration quality depends on the selection of the prior, its weight, as well as tuning of other parameters involving highly non-convex optimization setups\cite{nimisha2017blur}. Non-uniform blind deblurring for general dynamic scenes is a challenging computer vision problem as blurs arise from various sources, including moving objects, camera shake, and depth variations, causing different pixels to capture different motion trajectories. Such hand-crafted priors struggle while generalizing across different types of real-world examples, where blur is far more complex than modeled \cite{gong2017motion}.

Recent works based on deep convolutional neural networks (CNN) have examined the benefits of replacing the image formation model with a parametric model trained to emulate the non-linear relationship between blurred-sharp image pairs. Such works~\cite{nah2017deep} directly regress to deblurred image intensities and overcome the limited representative capability of variational methods in describing dynamic scenes. These methods can handle the combined effects of camera and dynamic object motion and achieve impressive results on the single image deblurring task. 

Existing CNN-based methods have two significant limitations:
a) Weights of the CNN are content-agnostic and spatially invariant, which may not be optimal for different pixels in different dynamically blurred scenes (e.g., homogeneous sky vs. moving car pixels). This issue is generally tackled by learning a highly non-linear mapping by stacking a large number of filters. But this drastically increases the computational cost and memory consumption. 
b) Traditional CNN layers usually result in a geometrically uniform and limited receptive field that is sub-optimal for the task of deblurring. Large image regions tend to be used to increase the receptive field even though the blur is small. This inevitably requires a network with a considerable number of layers with a high computation footprint or large downsampling operations leading to loss of finer pixel details.

Our work focuses on designing efficient and interpretable filtering modules that offer a better accuracy-speed trade-off than a simple cascade of convolutional layers. Since motion blur is inherently directional and different for each image instance, a deblurring network can benefit from adapting to the blur present in each input test image. We investigate motion-dependent adaptability within a CNN to directly address the challenges in single image deblurring. We deploy content-aware modules which adjust the filter to be applied and the receptive field at each pixel. Our analysis shows that the benefits of these dynamic modules for the deblurring task are two-fold:
i) Cascade of such layers provides a large and dynamically adaptive receptive field, which a normal CNN cannot achieve within a small number of layers.
ii) It efficiently enables spatially varying restoration, since changes in filters and features occur according to the blur in a particular region. With such a design, our network learns to exploit the similarity in the motion between different pixels within an image and is also sensitive to the position-specific local context.
\newline To fully exploit the capabilities of pixel-dependent operations in restoring the degraded regions, we upgrade our dynamic modules with a non-uniform adaptive sampling ability. This allows our network to densely sample and process pixels from the blurry areas explicitly, resulting in a significant improvement in pixelwise accuracy. Our design is based on the observation that for dynamic scene motion deblurring, the significant distortion occurs only at a few specific regions, for example, moving object boundaries or shaky edges. In contrast, homogeneous/low-texture areas are minimally affected. Intuitively, instead of exhausting the computational ability of the adaptive modules uniformly all over the image, the non-uniform adaptive sampling step allows us to `zoom into' the demanding regions and perform a much granular refinement. This strategy non-uniformly distributes the overall computational ability depending on the need of different areas.

We adopt a multi-patch hierarchical design with multiple stages as our backbone. At the end of each stage, we introduce GT image supervision, resulting in progressive deblurring of the image. At each stage, depending on the previous output, we perform pixel-adaptive refinement of the image, focusing on a finer refinement of the regions that are yet to be fully restored. This strategy first solves the easier areas and gradually shifts to more challenging ones in a coarse-to-fine manner, utilizing the knowledge gained in solving the easier regions.

\begin{figure}
   \centering
\begin{tabular}{ccc}
\hspace{-3.5mm} \includegraphics[width=0.16\textwidth, height = 0.09\textwidth]{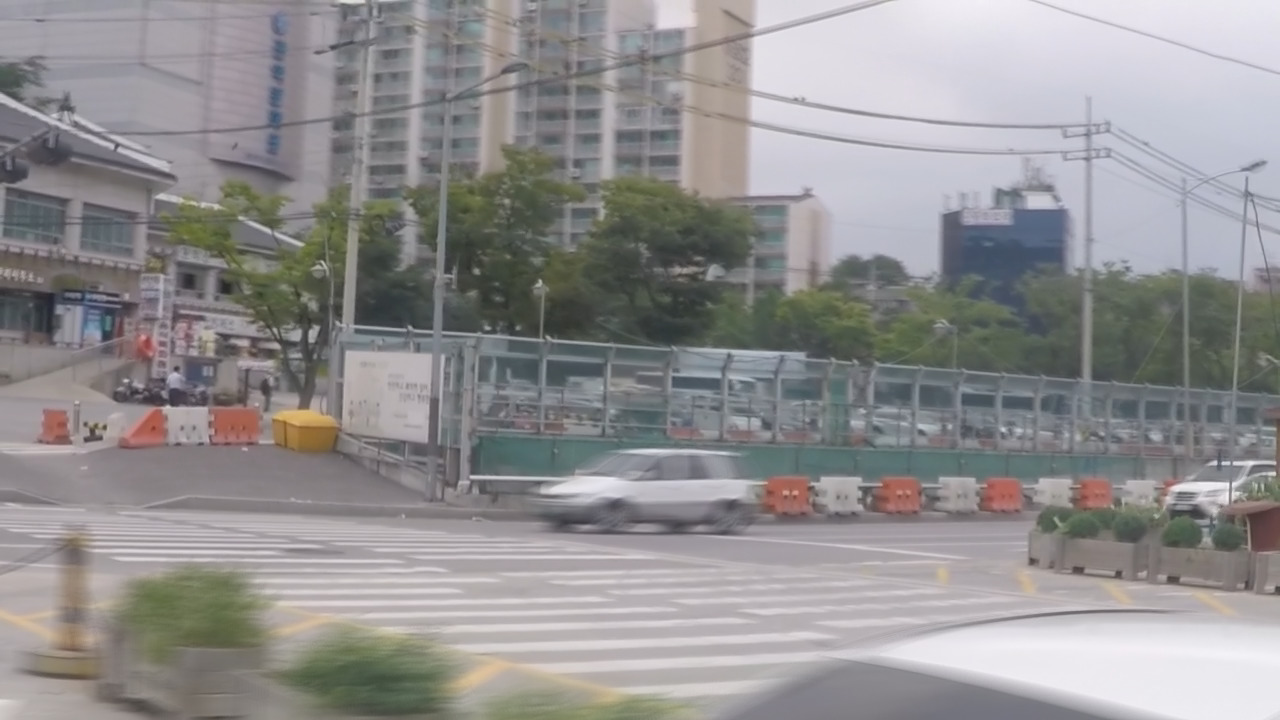} \hspace{-4.5mm} &
\includegraphics[width=0.16\textwidth, height = 0.09\textwidth]{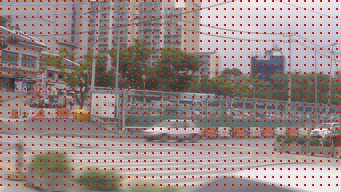} \hspace{-4.5mm} &%
\includegraphics[width=0.16\textwidth, height = 0.09\textwidth]{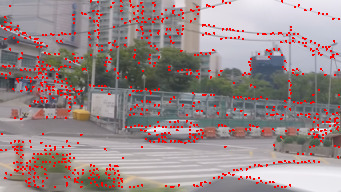} \\
Blurry Image & Uniform sample & Adaptive sampling
\end{tabular}

    \caption{Example of non-uniform sampling. Sampled pixels are shown in red. Best viewed when zoomed-in.}
    \label{fig:moti}
\end{figure}

The efficiency of our architecture is demonstrated through a comprehensive evaluation on two benchmarks and comparisons with the state-of-the-art deblurring approaches. Our model achieves superior performance while being computationally more efficient. The major contributions of this work are:
\renewcommand{\labelenumi}{\Roman{enumi}}
\begin{itemize}

    \item We propose an efficient deblurring design built on content-adaptive modules that learn the transformation of features using global attention and adaptive local filters. We show that these two branches complement each other and result in superior deblurring performance. 
    
    \item We further enhance the utility of the adaptive modules by introducing a non-uniform pixel-adaptive sampling step. This allows the modules to focus on a finer refinement of the heavily blurry areas while maintaining the computational load.
    
    \item We also demonstrate the efficacy of learning cross-attention between encode-decoder as well as different levels in our design for improved feature propagation.
    
    \item We provide extensive analysis and evaluations on multiple dynamic scene deblurring benchmarks, demonstrating the efficacy of our approach.
\end{itemize}

Compared with our original conference version \cite{suin2020spatially}, the following improvements are introduced: (1) We introduce GT supervision at each stage to progressively restore the image. (2) We enhance the deblurring ability of the adaptive modules by introducing a non-uniform pixel-adaptive sampling step (Sec. \ref{sec:pas}). Although progressive restoration was explored in MPR \cite{Zamir_2021_CVPR_mprnet}, our work demonstrates the utility of the adaptive sampling module coupled with progressive restoration to significantly improve the performance. The operation of the updated attention module and the filtering block are mentioned in Eqs. \ref{eq:xnu_att}, \ref{eq:att_p}, \ref{eq:nu_filter}. (3) We have analyzed the behavior of the adaptive sampling step in Tables \ref{table:ablation}, \ref{table:sampling}, Figs. \ref{fig:visualization}, \ref{fig:sampling}. (4) We include extensive experiments on additional RealBlur dataset \cite{rim_2020_realblur} to verify the effectiveness of our approach for varying scenarios. The rest of this paper is organized as follows. We first review related work in Sec. \ref{sec:rel_works} and describe the architecture of our network in Sec. \ref{sec:method}. In Sec. \ref{sec:exp} experimental results are given, and in Sec. \ref{sec:ablation}, behavior of different modules are analyzed. Sec. \ref{sec:conc} concludes the work.
\vspace{-3mm}
\section{Related Works}
\label{sec:rel_works}
Dynamic Scene deblurring is a highly ill-posed problem as blurs can arise from various intertwined factors. Conventional image deblurring studies often make different assumptions, involving expensive non-convex optimization for estimating image and blur kernel. \cite{fergus2006removing} exploited the sparse nature of blur kernel, \cite{cho2009fast} utilized image gradients inside a multi-scale framework, \cite{gong2016blind} used a gradient activation method to select a subset of gradients. The role of edge maps was explored in \cite{xu2010two}. A framework based on the conditional random field was used in \cite{hu2012good}. \cite{hyun2013dynamic} jointly produced motion segments along with the kernel and the restored image, where each segment shared a kernel. Such methods typically struggle to handle depth variations and complex motion. \cite{hyun2014segmentation} designed a segmentation-free algorithm.  Most of the conventional methods impose different constraints with handcrafted regularization. Such as low-rank approximation \cite{ren2016image}, total variation based prior \cite{pan2014motion}, dark channel prior \cite{pan2016blind}, extreme channel priors \cite{yan2017image},  discriminative prior \cite{li2018learning}, color line prior \cite{lai2015blur}, etc. Such methods typically suffer from high time and memory consumption and may even lead to artifacts for complex scenarios where the priors used might not hold.

\par  With the advancement of deep neural networks, many learning-based approaches were proposed for single image deblurring. Some preliminary works like \cite{sun2015learning} estimated non-uniform patch-wise blur kernels using deep neural networks and then resorted to non-blind deblurring algorithms to produce the restored sharp image. It suffers from a lack of high-level global information, and the conventional non-blind deblurring step is time-consuming. \cite{gong2017motion} assumed linear blur kernels and used CNN to estimate the motion flow.  \cite{xu2017learning} deployed a generative adversarial network (GAN) to jointly deblur and super-resolve blurry text and face images. But, methods designed for restoring text, face, license plate, etc. (\cite{xu2017learning, hradivs2015convolutional,svoboda2016cnn}) fail to handle general dynamic scenes. \cite{nimisha2017blur} proposed a sparse-coding-based framework consisting of both variational and adversarial learning. \cite{kupyn2018deblurgan} proposed a conditional GAN-based framework. Recently, multi-scale coarse-to-fine methods achieved huge performance improvements. Methods like \cite{gao2019dynamic,nah2017deep,tao2018scale} gradually restore a sharp image on different resolutions. \cite{cho2021rethinking}  proposed a  coarse-to-fine network with asymmetric feature fusion between different scales. \cite{suin2024diffuse,purohit2021spatially} also proposed spatially-varying netwoeks for a diverse range of tasks. \cite{zhang2019deep} proposed a multi-patch hierarchical network where images are divided into multiple patches at each stage.  \cite{Zamir_2021_CVPR_mprnet} also followed a multi-patch hierarchical network that progressively performs the deblurring. We also use a progressive restoration strategy. But, as discussed in Sec. \ref{sec:ablation}, only progressive refinement marginally improves the performance over our earlier work. But, when used together with the pixel-adaptive sampling strategy, it significantly enhances the deblurring accuracy.

\par In general dynamic scenes, blur is essentially spatially variant and also varies from one image to another. Thus, using the same filtering operation to handle all cases is inefficient. A more natural choice would be to use a network that can adapt its behavior depending on the nature of the blur present at a particular pixel. \cite{zhang2018dynamic} is one of the first works that deployed a spatially-varying operation using RNNs. But, the expressibility of their design is much inferior to ours. Typically, the RNN structure is more restrictive than self-attention. The RNN accumulates pixel information from four directions and cannot directly emphasize certain pixels while suppressing others (adaptive weightage). In contrast, our self-attention design can dynamically decide the relative weightage of all the pixels using a set of spatially-varying attention maps. Furthermore,  RNN only focuses on long-range information accumulation. In contrast, we empirically demonstrate that both global and local adaptations are crucial, and the requirement can vary from pixel to pixel. A few concurrent super-resolution works also explored the utility of spatially-varying operations. \cite{xu2020unified} predicted per-pixel kernels, \cite{niu2020single} used a simplistic mask-multiplication-based spatial and channel attention mechanism, \cite{yi2019progressive} used standard self-attention but on a highly downsampled feature map to handle the computational burden. None of these works explored the computational aspect of self-attention for restoration tasks or the utility of both global and local information together. Our original work previously demonstrated the advantages of adaptive global-local pixel-refinement operations. In the current extension, we further introduce an adaptive sampling strategy for image deblurring. To our knowledge, ours is the first restoration work to utilize adaptive sampling to focus more on the difficult-to-restore pixels, increasing the pixelwise accuracy while maintaining the overall computational cost.

\vspace{-4mm}
\section{Method}
\label{sec:method}
To date, the driving force behind performance improvement in deblurring has been the use of a large number of layers and larger filters, which assist in increasing the ``static'' receptive field and the generalization capability of a CNN. However, these techniques offer suboptimal design since network performance does not always scale with network depth and the effective receptive field of deep CNNs is much smaller than the theoretical value (investigated in \cite{luo2016understanding}).

\begin{figure}
    \centering
    \includegraphics[width = 0.49\textwidth]{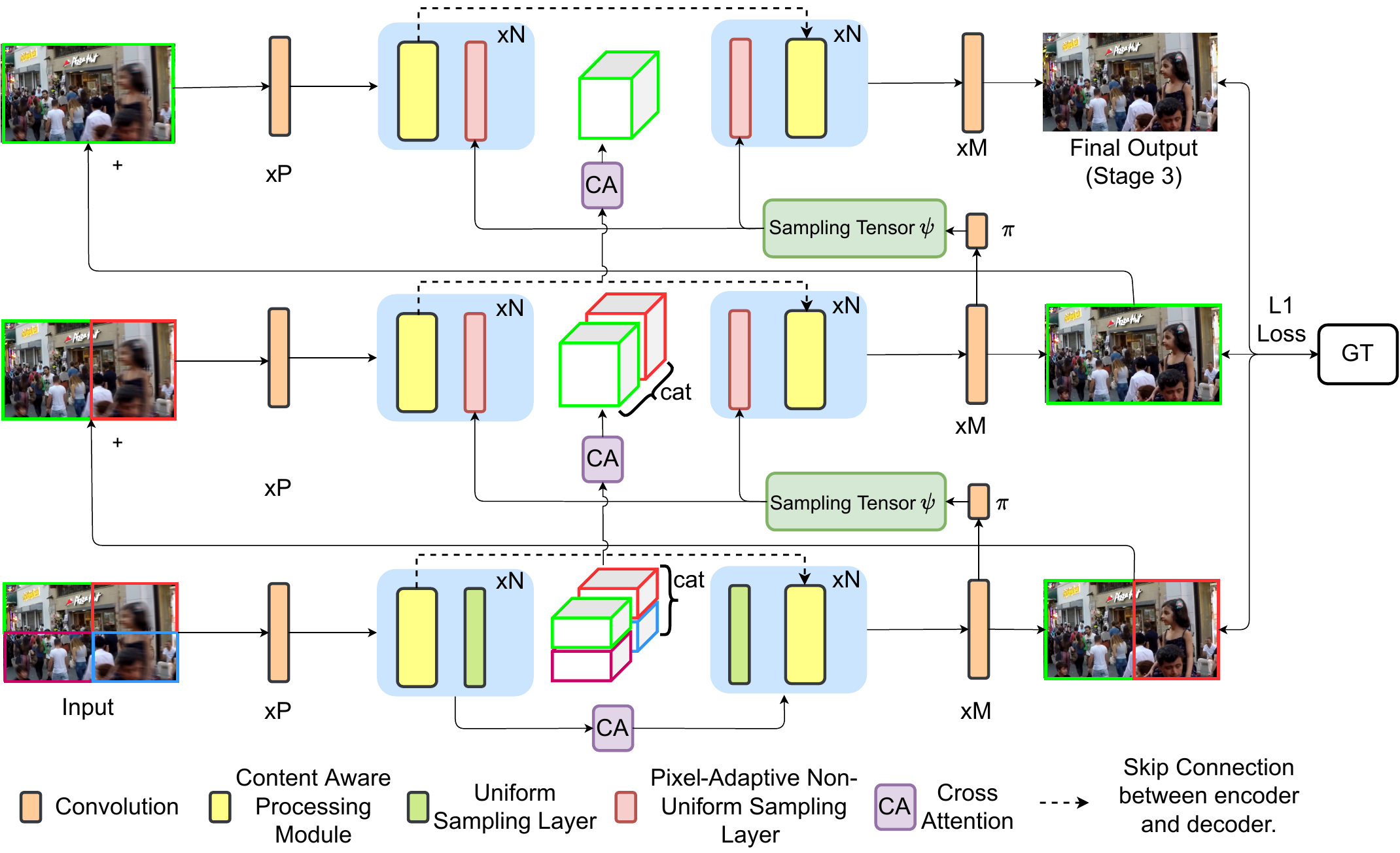}
    \caption{Overall architecture of our proposed network. CA block represents cross attention between different levels of encoder-decoder and different levels. All the resblock contains one content aware processing module. `+' denotes elementwise addition. We include skip connection between the encoder and decoder, which is described in Fig. \ref{fig:content_aware} for the non-uniform sampling.}
    \label{fig:arch}
\end{figure}

We claim that a superior alternative is a dynamic framework wherein the filtering and the receptive field change across spatial locations and different input images. Our experiments show that this approach is a considerably better choice due to its task-specific efficacy and delivers consistent performance across diverse magnitudes of blur.

Although previous multi-scale and scale-recurrent backbones have shown good performance in removing non-uniform blur, they suffer from expensive inference time and performance bottlenecks while simply increasing model depth. Instead, inspired by DMPHN \cite{zhang2019deep}, we adopt a more advantageous multi-patch hierarchical structure as our base model. The hierarchical architecture facilitates residual connections between different levels as the resolution (combining all patches) is the same across levels, unlike \cite{nah2017deep,tao2018scale}. The residual learning significantly eases the learning compared to using a huge stack of convolution layers. Also, dividing the image into multiple small patches instead of a single downscaled image results in the lower levels being exposed to more training data. Ultimately, the multi-patch architecture can use much smaller filters at every level while producing better performance. 
Inheriting these benefits, coupled with parallel processing of multiple patches, our approach achieves a much faster processing time than \cite{tao2018scale} ($\times1.5$), \cite{nah2017deep} ($\times7.6$) and even \cite{zhang2019deep} ($\times1.25$) due to the architectural adaptations described next.

The overall architecture of our proposed network is shown in Fig. \ref{fig:arch}. The network has a hierarchical structure with multiple stages stacked together. To keep a satisfactory balance between restoration performance and computational requirement, we use three stages (instead of four, as described in \cite{zhang2019deep}) in our primary model. Unlike \cite{zhang2019deep}, we also avoid cascading our network along depth, as that adds a severe computational burden. Instead, we advocate the use of content-aware processing modules which yield significant performance improvements over even the deepest stacked versions of original DMPHN \cite{zhang2019deep}.

At the bottom level, input is sliced into 4 non-overlapping patches for processing, and as we gradually move towards higher levels, the number of patches decreases. Akin to the standard encoder-decoder paradigm, each stage of our network originally had a bottleneck structure. In the encoder, we increased the number of feature maps while decreasing the spatial resolution uniformly across the spatial domain, and this process was reversed in the decoder. The pixel-adaptive processing modules were used as the primary building block of both encoder and decoder. The pixel-adaptive module comprises parallel global and local-level feature processing branches that are dynamically fused at the end. We also design cross-attention between layers at different depths and stages to effectively propagate features throughout the network.

To upgrade our network with a non-uniform pixel-adaptive sampling, we make the following changes: 
\textbf{(i)} We introduce GT image supervision at the end of each stage so that the network learns to restore the image progressively. \textbf{(ii)} We keep the first stage unchanged, i.e., it processes the whole image uniformly (spatially) like any standard encoder-decoder structure. It allows us to maintain the overall pixel-wise accuracy, especially for the regions with a small blur level. \textbf{(iii)} For the subsequent stages, we perform non-uniform pixel-adaptive sampling, depending on the previous stage's output. We deploy a learnable pixel-sampling module (Sec. \ref{sec:pas}) that adjusts the sampling rate or the `dense-ness' of pixels picked from a particular region and generates a non-uniform grid. It allows us to ``zoom" into the challenging blurry areas and performs refinement at a much finer scale. \textbf{(iv)} This adaptive sampling step is integrated into our pixel-adaptive modules. Note that, typically, the spatial correlation between neighboring pixels is crucial in a neural network, but it also depends on the operation we perform on them. For example, it is essential for convolution, which is generally used to capture local patterns. But, self-attention, a non-local operation, does not depend on the locality of a pixel but on the similarity in the content. In our case, the effectiveness of the self-attention operation after zooming in stays unchanged. For performing pixel-adaptive convolution, we first zoom into a critical pixel and then extract the neighboring patch around it for the convolution operation, thus preserving its local neighborhood. The detailed operations are described in Sec. \ref{sec:content_aware}.

To intuitively understand the scenario, let us assume a scene with significant motion blur in a small part of the image, whereas the rest of the regions have minimal or no blur. One such example is shown in Fig. \ref{fig:moti}. Although the pixel-adaptive modules, by nature, adjusts their behavior depending on the pixel content, we would under-utilize its ability in restoring the crucial degraded regions if applied on a uniformly downsampled feature map. Due to the uniform-grid-sampling in a standard bottleneck structure, the majority of the pixels may belong to the static background with minimal/no blur, and we would potentially pick only a few pixels from the severely blurry areas that we aim to restore. Uniform grids are convenient but not necessarily ideal for image restoration. A uniform grid will more often than not undersample the blurry pixels.

This phenomenon will be more prominent in the higher stages, as the network should sufficiently deblur the ``easier'' regions in the first stage itself. Also, dense sampling for all the pixels is unnecessary and is often infeasible in current hardware, incurring a massive computational load, as discussed in Sec. \ref{sec:ablation}. Instead, while keeping the computational load the same, we redistribute the overall computation non-uniformly across the spatial domain, where the density of sampled pixels is designed to be directly proportional to the difficulty level, as shown in Fig. \ref{fig:moti}. Cost-performance analysis in Sec. \ref{sec:exp} shows that our upgraded pixel-adaptive design outperforms existing works in the majority of the cases while maintaining a balance between accuracy and computational complexity. We begin with describing the pixel-adaptive sampling stage, then proceed towards the detailed description of the content-aware processing module.

\subsection{Pixel-Adaptive Sampling}
\label{sec:pas}
Let the spatial dimension of the input image be $H \times W$. Thus, the maximum grid size is $H \times W$, which denotes the maximum sampling rate for any area at any network stage. In case of uniform sampling, we produce feature maps of size $\frac{H}{r} \times \frac{W}{r}$, where $r$ is the scaling factor, like 2,4,8, etc. In the case of pixel-adaptive sampling, we still pick a total of $\frac{H}{r} * \frac{W}{r}$ pixels from the whole $H \times W$ dimensional space, but the selected pixel-positions are non-uniformly distributed depending on the degradation level of a particular area.

Let us consider the input tensor $X = \{X_{j}\} \in \mathbb{R}^{C \times HW}, j \in [1,HW]$. Assuming normalized coordinate system, all pixels will have spatial coordinates in the range of $[0, 1]$. Let $\psi = \{\psi^x, \psi^y\} \in \mathbb{R}^{L \times 2}$ be the sampling tensor, where $L \in [1,HW]$ is the number of pixels to be sampled, $\psi^x, \psi^y \in \mathbb{R}$ denote the sampling coordinates along $x$ and $y$ spatial dimensions, respectively. The sampling operator, $s$, maps a pair of input tensor $X$ and sampling tensor $\psi$ to the non-uniformly sampled output tensor $Y = \{Y_{l}\} \in \mathbb{R}^{C \times L}, l \in [1,L]$ as
\begin{equation}
    \label{eq:sample_fun}
    s: \mathbb{R}^{C \times HW} \times [0,1]^{L \times 2} \rightarrow \mathbb{R}^{C \times L}, (X,\psi) \rightarrow Y
\end{equation}
such that
\begin{equation}
\label{eq:sample_fun2}
    Y_{l} = X[\psi_{l}^x, \psi_{l}^y]
\end{equation}
This is a more generalized form of standard sampling operations used in neural networks. For example, a full-scale uniform sampling can be expressed by a sampling tensor $\psi \in \mathbb{R}^{2 \times HW}$ with the $l^{th}$ element as the normalized coordinate for the $l^{th}$ pixel, $l \in [1,HW]$. Note that $\pi(\cdot)$ is a convolutional operation that takes the input feature map and determines which pixels to sample. Depending on its output, a sampling operator $\psi$ is generated that is used to sample the pixels from the full resolution feature map (Eqs. \ref{eq:sample_fun} and \ref{eq:sample_fun2}) for further refinement.

After adaptively sampling and then processing $L$ number of pixels, we can perform the reverse operation to get back the original $H \times W$ grid as the final output.
\begin{equation}
    s': \mathbb{R}^{C \times L} \times [0,1]^{L \times 2} \rightarrow \mathbb{R}^{C \times HW}, (Y,\psi) \rightarrow X'
\end{equation}
The sampled and refined pixels are directly put back in the output grid. In the decoder, we add a direct skip connection with the encoder features of next higher resolution ($\frac{H}{r-1} \times \frac{W}{r-1}$) for the other pixel locations and pass through a convolutional layer to merge the information. Compared to standard down-up/sampling operations, the key difference in our design is that we do not have a fixed up/down sampling rate for all the locations. We preserve the crucial spatial details for the difficult-to-restore blurry regions with a denser sampling. The rest of the areas which have a minimal amount of blur or are already restored in the previous stages can be processed with a much sparser sampling.

\subsubsection{Learning $\Psi$}{Learning $\Psi$}
The sampling tensor $\psi$ is generated depending on the previous intermediate feature output of the network. To detect the pixels with significant amount of blur, that are yet to be fully restored, we generate a $HW$ dimensional probability map $G$, where the $j^{th}$ element of $G$, $G_j \in \mathbb{R}, j \in [1,HW]$ denotes the the probability of the $j^{th}$ pixel being a hard-to-restore pixel. Given the previous intermediate feature output $X_{-1} \in \mathbb{R}^{C \times HW}$, $G$ can be generated as
\begin{equation}
    G = \text{sigmoid}(\pi(X_{-1};\theta_{\pi}))
\end{equation}
where is $\pi(\cdot)$ is convolutional operation, parameterized by $\theta_{\pi}$. During inference, we select the top $L = \frac{H}{r}\frac{W}{r}$ ``hardest'' pixel-positions with the highest probabilities in $G$. While training, we select $L' \approx L$ number of pixel-positions by sampling from the pixelwise probability distributions $G$. After selecting $L$ pixels with highest probabilities, generating $\psi$ with the normalized $x$ and $y$ coordinates is straightforward. $\psi$ will be a $L \times 2$ tensor, which contains the normalized $xy$ coordinates for the selected $L$ pixels.

A technical challenge of this approach is: making `hard'-decisions (binary decisions) about picking $L'$ pixels while training, i.e., sampling from the pixelwise probability distribution $G$ is not backpropagation friendly, the gradients will not flow to $\theta_{\pi}$, and thus to the pixel coordinates ($\psi$). To learn $G$, and in turn $\psi$, we formulate this as a per-pixel decision-making problem that naturally fits into a reinforcement learning (RL) framework and addresses the gradient flowing problem with policy-gradient-based algorithms \cite{sutton1999policy,sutton2018reinforcement}. In the full-scale $H \times W$ grid, we decide for each of the $H*W$ pixel locations whether that belongs to a severely blurry area. In a straightforward solution, we can treat each location individually and make decisions using separate RL-agents in a multi-agent framework \cite{foerster2017stabilising,foerster2016learning}. But, this naive design is computationally impractical and can not handle images of arbitrary size. Instead, we exploit the convolutional structure, where given a $H\times W$ input, we produce a $H\times W$ pixelwise probability map. Intuitively we can think of it as multiple agents with shared parameters (the kernel of the convolutional layer) that compute the pixelwise decisions in parallel. 
\newline Any RL-agent is trained to maximize a reward fucnction. In our case, to provide feedback to the $\pi(\cdot)$ layer generating the probability map $G$, we design the following reward function
\begin{equation}
\label{eq:reward_def}
    R = - \alpha \cdot (|\text{OP}_2-\text{GT}| +  |\text{OP}_1-\text{GT}|) - \beta \cdot \mathbbm{1}(L' > L)
\end{equation}
where $\alpha, \beta$ are the relative weights selected empirically as $1$ and $0.5$, $R \in \mathbb{R}^{HW}$. The first term denotes the pixelwise accuracy of the outputs from different stages, by measuring the pixelwise intensity differences with the GT image. Maximizing the reward will essentially minimize the pixelwise differences, which is our ultimate goal.

The second term acts as a penalty and it is true only if $L' > L$. This helps us to maintain the computational budget by processing almost the same number of pixels as in the case of an uniform sampling. 
\par We deploy standard policy gradient based training strategy, which strives to maximize the reward $R$ (Eq. \ref{eq:reward_def}). During training, $\theta_\pi$ is updated to maximize the following function
\begin{equation}
\label{eq:rl_loss_1}
    \mathcal{L} = \mathbb{E}_{\psi \sim \pi(X_{-1},\theta_\pi)}[R]
\end{equation}
Following \cite{sutton2018reinforcement}, the expected gradient can be calculated as
\begin{equation}
\label{eq:rl_loss}
\nabla_{\theta_\pi} \mathcal{L} = \mathbb{E}[(R) \nabla_{\theta_\pi} \text{log} \pi(\cdot|X_{-1})]
\end{equation}
Practically, Eq. \ref{eq:rl_loss} is approximated by a single sample from $\pi(X_{-1},\theta_\pi)$ and the corresponding reward $R$. For a detailed analysis of the policy gradient, we encourage the reader to refer to \cite{sutton1999policy,sutton2018reinforcement}. Although the final objective of both the RL agent and the restoration network is to minimize the reconstruction loss, they operate in different ways. Specifically, the RL agent tries to minimize the loss by learning to identify the critical pixels that need to be restored. Given the critical pixels, the deblurring network learns to perform adequate refinement operations. Typically training any RL agent requires a stable environment to receive meaningful rewards. Thus, we first pre-train our deblurring network alone for a few epochs to get a good initialization and then jointly train both the networks. Note that, as we are taking intermediate feature map as the input while predicting $\psi$, $\pi(\cdot)$ is simply stack of a few convolutional layers and extremely lightweight. It adds almost negligible computational overhead, while significantly boosting the restoration accuracy.

\subsection{Content-Aware Processing Module}
\label{sec:content_aware}
In contrast to high-level problems such as classification and detection \cite{wang2018non}, which can obtain large receptive field by successively down-sampling the feature map with pooling or strided convolution, restoration tasks like deblurring need finer pixel details that can not be achieved from highly downsampled features.

Most of the previous deblurring approaches use standard convolutional layers for local filtering and stack those layers together to increase the receptive field. \cite{bello2019attention} uses self-attention and standard convolution on a parallel branch and shows that best results are obtained when both features are combined compared to using each component separately. Inspired by this approach, we design a content-aware ``global-local" processing module which depending on the input, deploys two parallel branches to fuse global and local features. The ``global" branch is made of an attention module. For the decoder, this includes both self and cross-encoder-decoder attention, whereas only self-attention is used for the encoder. We design a pixel-dependent filtering module for the local branch that determines the weight and the local neighborhood to apply the filter adaptively. We describe these two branches and their adaptive fusion strategy in detail in the following sections.
\subsubsection{Attention}
Following the recent success of transformer architecture \cite{vaswani2017attention} in the natural language processing domain, it has been introduced in image processing tasks as well \cite{parmar2018image,liu2018non}. The main building block of this architecture is self-attention which, as the name suggests, calculates the response at a position in a sequence by attending to all positions within the same sequence. Given an input tensor of shape $(C,H,W)$ it is flattened to a matrix $z \in \mathbb{R}^{HW \times C}$ and projected to $d_a$ and $d_c$ dimensional spaces using embedding matrices $W_a,W_b \in \mathbb{R}^{C \times d_a}$ and $W_c \in \mathbb{R}^{C \times d_c}$. Embedded matrices $A,B \in \mathbb{R}^{HW \times d_a}$ and $C \in \mathbb{R}^{HW \times d_c}$ are known as query, key and value, respectively. The output of the self-attention mechanism for a single head can be expressed as
\begin{equation}\label{eq:qkv}
    O = \text{softmax} \left(\frac{AB^T}{\sqrt{d_a}} \right)C
\end{equation}
The main drawback of this approach is the very high memory requirement due to the matrix multiplication $AB^T$, which requires storing a high dimensional matrix of dimension $(HW,HW)$ for the image domain. This requires a large downsampling operation before applying attention. \cite{parmar2018image} and \cite{ramachandran2019stand} use a local memory block instead of global all-to-all for making it practically usable. \cite{bello2019attention} uses attention with the smallest spatial dimension until it hits memory constraints. Also, these works typically resort to smaller batch sizes and sometimes additionally downsampling the inputs to self-attention layers. Although self-attention is implemented in recent video super-resolution work \cite{yi2019progressive}, to reduce memory requirement, it resorts to pixel-shuffling. This process is sub-optimal for spatial attention as pixels are transferred to channel domain to reduce the size.
\par Unlike others, we resort to an attention mechanism that is lightweight and fast. If we consider Eq. (\ref{eq:qkv}) without the softmax and scaling factor for simplicity, we first do a $(HW,d_a) \times (d_a,HW)$ matrix multiplication and then another $(HW,HW) \times (HW,d_c)$ matrix multiplication which is responsible for the high memory requirement and has a complexity of $\mathcal{O}(d_a(HW)^2)$. Instead, if we look into this equation differently and first compute $B^TC$ which is an $(d_a,HW) \times (HW,d_c)$ matrix multiplication followed by $A(B^TC)$ which is an $(HW,d_a) \times (d_a,d_c)$ matrix multiplication, this whole process becomes lightweight with a complexity of $\mathcal{O}(d_ad_cHW)$. We suitably introduce softmax operation at two places which makes this approach intuitively different from standard self-attention but still efficiently gathers global information for each pixel. Note that our design significantly reduces the computational overhead while slightly compromising on its expressibility, as we do not directly calculate the relation between all possible pixel pairs. Practically, there will always be a trade-off between the computational aspect and performance. A nonlinear relationship is theoritically more expressive, but it can not operate on high-resolution features. Our design focuses more on preserving pixel accuracy. We have observed that for image restoration tasks like deblurring, maintaining the resolution or pixel accuracy is more important than allowing more complex relations between pixels. Empirically we show that it performs better than standard non-linear self-attention as discussed in ablation studies. Also, due to the lightweight nature, it not only enables us to use this in all the encoder and decoder blocks across levels for self-attention but also across different layers of encoder-decoder and levels for cross attention which results in a significant increase of accuracy. 

\begin{figure*}
    \centering
    \includegraphics[width = 0.9\textwidth]{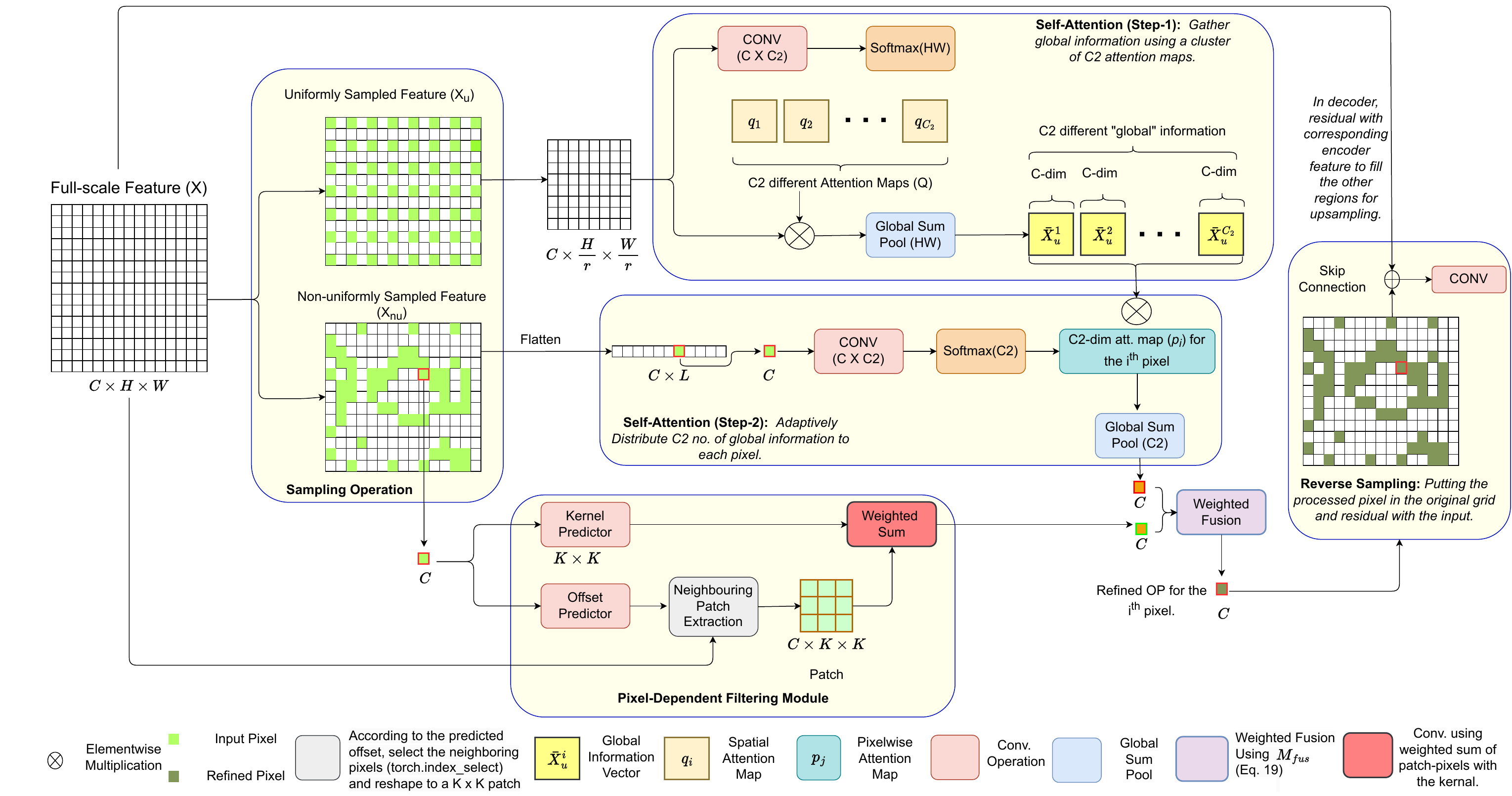}
    \caption{Illustration of our content-aware processing module. The upper and lower branch show self-attention and PDF module, respectively.
}
    \label{fig:content_aware}
\end{figure*}
\subsubsection{Self-Attention (SA)}
Let $X_u$ be the input to the attention block. As the first stage of our network is unchanged, i.e., we are using standard bottlneck structure with uniform sampling across the spatial domain, we will have $X_u \in \mathbb{R}^{C \times \frac{H}{r}\frac{W}{r}}$, where $r$ is the static down-scale factor, $H$ and $W$ are the original resolution of the image. In an encoder-decoder structure, we usually have $r = [2,4,8,...]$.

In the improved network with pixel-adaptive sampling in the subsequent stages, we generate a non-uniformly sampled feature tensor $X_{nu} \in \mathbb{R}^{C \times L}$ as
\begin{equation}
    \label{eq:xnu_att}
    X_{nu} = s(X,\psi)
\end{equation}
where sampling function $s$ is defined in Eq. \ref{eq:sample_fun}. $L = \frac{H}{r}\frac{W}{r}$ is the number of pixels to be adaptively sampled from the whole $H \times W$ space. We calculate the pixel-adaptive sampling tensor $\psi \in \mathbb{R}^{L \times 2}$ as $\psi = \pi(X_{-1})$ where $X_{-1}$ is the output of the previous stage (Fig. \ref{fig:arch}).

Having these two feature tensors $X_u$ and $X_{nu}$ as the input, we discuss the global information gathering operation. In self-attention, each pixel gathers global information from the whole image, increasing the receptive field significantly. In our efficient implementation, we perform the self-attention operation in two stages: 1. We first extract a set of global information using a cluster of spatial attention maps and global pooling. 2. Next, we adaptively distribute the required information from this set to each output pixel. While extracting the global information, we are expected to cover the whole scene and not any particular object or area. Thus, for stage 1, we utilize the uniformly sampled feature $X_u$ as described next.
\newline \textbf{Step-1:} Given the feature map $X_{u}$, we pass it through a convolutional layer $f_q$ followed by spatial-softmax (along $HW$) operation to generate $Q = \text{softmax}_{HW}(f_q(X_{u})), Q \in \mathbb{R}^{C_2 \times \frac{H}{r}\frac{W}{r}}$. As we apply spatial-softmax, $Q$ can be considered a set or cluster of $C_2$ spatial attention maps where each feature channel will represent a different map. Intuitively, a single spatial attention map will capture one aspect of the blurred image. However, there can be multiple pertinent properties like edges, textures, etc., in the image that can be equally crucial for removing the blur. Therefore, we deploy a cluster of learnable attention maps to effectively gather $C_2$ different key features. Next, we element-wise multiply the uniformly sampled feature map $X_{u}$ with each attention map ($q_i \in \mathbb{R}^{\frac{H}{r}\frac{W}{r}}, i = 1,2,...,C_2$) to generate as $X^i_{u}$
\begin{equation}
\label{eq:spat_cluster}
    X^i_{u} = q_i \odot X_{u} ~ ~ ~ , \text{with}  \sum_{j=1}^{\frac{H}{r}\frac{W}{r}} q_{ij} = 1
\end{equation}
where $X^i_{u} \in \mathbb{R}^{C \times \frac{H}{r}\frac{W}{r}}$. As $X^i_{u}$ is generated after multiplication with normalized weights (output of softmax), we can extract a global feature by global-sum-pooling (GSP) along spatial $HW$ dimension. The $i^{th}$ `global' feature can be calculated as 
\begin{equation}
\label{eq:att-gsp-hw}
    \bar{X}^i_{u} = GSP_{HW}(X^i_{u}) ~ ~ ~ ~ ~ ~(i = 1,2,...,C_2)
\end{equation}
where $\bar{X}^i_{u} \in \mathbb{R}^C$. If we repeat this process (i.e., multiplication followed by weighted summation) for all the spatial attention maps in $Q$, we will have a set of $C_2$ different global information. Specifically, we will have $\bar{X}_{u} = \{\bar{X}^1_{u},\bar{X}^2_{u},...,\bar{X}^{C_2}_{nu}\}, \bar{X}_{u} \in \mathbb{R}^{C \times C_2}$. Note that, instead of performing the multiplication and weighted summation operation for each attention map ($q_i$) separately, we can represent the whole operation as a matrix-matrix multiplication that can be efficiently executed using highly optimized modern deep learning libraries. $X_{u} \in \mathbb{R}^{C \times \frac{H}{r}\frac{W}{r}}$ can be considered a matrix where each row will contain all the pixels in a particular feature map and there are $C$ such rows (channels). Similarly, we can have $Q^T \in \mathbb{R}^{\frac{H}{r}\frac{W}{r} times C_2}$, where each column will contain normalized attention weights for all the $\frac{H}{r}\frac{W}{r}$ pixels and there are $C_2$ such columns (different attention maps). Thus, $\bar{X}_{u}$ can be generated as
\begin{equation}
\label{eq:att-step1-matmul}
    \bar{X}_{u} = (X_{u})(Q^T)
\end{equation}
where all the spatial pixels will be automatically multiplied with the corresponding pixel-weights and aggregated as per the properties of matrix multiplication.

For extracting a global feature, the overall scene content is more crucial than finer details of any particular object or area. Thus, for step-1, we can use $X_u \in \mathbb{R}^{C \times \frac{H}{r}\frac{W}{r}}$ even with large downscaling factor, as we validate in Sec. \ref{sec:ablation}. But, in the next step, where we utilize this global information to update the blurry pixels, densely sampling and refining pixels around the hard-to-restore degraded areas is critical for the final output accuracy. Although in self-attention, each pixel can extract information from the helpful regions, these pixels come from the downsampled feature map. In a standard encoder-decoder framework, the features are spatially downsampled by a large factor at the bottleneck. Thus, the efficacy of self-attention is still limited by the spatial resolution of the input feature map.
Following this intuition, we utilize the non-uniformly sampled feature tensor $X_{nu}$ in Step-2, as described next.
\newline \textbf{Step-2:} Given the non-uniformly sampled feature map $X_{nu}$, we pass it through a convolutional operation $f_p$ followed by softmax along the channels to generate $P = f_p(X_{nu}), P \in \mathbb{R}^{C_2 \times \frac{H}{r}\frac{W}{r}}$. As we apply softmax operation along the channels, $P$ can be considered the pixelwise attention maps for each sampled pixel. Specifically, we can express $P$ as $P = \{p_1,p_2,...,p_{L}\}$ where $p_j \in \mathbb{R}^{C_2}$ is the attention map for $j^{th}$ sampled pixel, $j \in [1,L]$. Intuitively, $p_j$ shows the relative importance of $C_2$ different `global' information ($\bar{X}_{u}$) for the $j^{th}$ pixel. Thus, for each pixel, we utilize $p_j$ to adaptively select a particular combination of the $C_2$ no. of global information extracted in step-1. For the $j^{th}$ non-uniformly sampled pixel, we element-wise multiply the $C_2$ global feature representations $\bar{X}_{u}$ with the corresponding attention map $p_j$, to get
\begin{equation}
\label{eq:att_p}
Y^j = p_j \odot \bar{X}_{u} ~ ~ \text{with}  \sum_{i=1}^{C_2} p_{ij} = 1 ~ ~ , (j = 1,2,...,L)
\end{equation}
where ${Y}^j \in \mathbb{R}^{C \times C_2}$. Similar to Eq. \ref{eq:att-gsp-hw}, as $Y^j$ is generated after multiplication with normalized channel weights, we extract the 
global feature representation for the $j^{th}$ pixel using global-sum-pooling (GSP) along $C_2$ as
\begin{equation}
    \bar{Y}^{j} = GSP_{C_2}({y}^j)
\end{equation}
where $\bar{Y}^{j} \in \mathbb{R}^C$ represent the accumulated global feature for the $j^{th}$ sampled pixel. Thus, each of the densely sampled pixel flexibly selects features that are complementary to the current one and accumulates a global information. Similar to Eq. \ref{eq:att-step1-matmul}, instead of repeating this process for all the pixels separately, we can express it as a matrix-matrix multiplication as $(\bar{X}_{u})(P)$. Each row of $\bar{X}_{u}$ contains $C_2$ different global information whereas each column of $P$ contains the relative weightage of those global information for a particular pixel. Thus, step-1 and step-2 together can be expressed by efficient matrix-operations as
\begin{equation}
    Y^{att} =   \left[(X_u)(Q)^T\right](P) 
\end{equation} 

To encourage the flow of most useful global information while suppressing the unnecessary ones, we introduce two sigmoid-mask multiplication steps as below
\begin{equation}
\label{eq:mask}
    Y^{att} =   M_C \odot \left[(M_S \odot X_u)(Q)^T\right](P) \\ 
\end{equation} 
where $M_C \in \mathbb{R}^{C}$, $M_S \in \mathbb{R}^{\frac{H}{r}\frac{W}{r}}$ are generated using a single convolution layer followed by sigmoid operation, while taking $X_u$ as input, respectively. $M_S$ and $M_C$ are used to emphasize the most useful information along spatial and channel dimensions, respectively. This efficient and simple matrix multiplication makes this attention module very fast, whereas the order of operation (computing $[(A)^T\text{softmax}(B)]$ in Eq. \ref{eq:qkv}) results in a low memory footprint. Note that, for the first stage in our network, we simply use $X_u$ in step-2 for estimating $P$, and $L = \frac{H}{r}\frac{W}{r}$, akin to our original design.
\vspace{-2mm}
\subsubsection{Cross-Attention (CA)}
Inspired by the use of cross-attention in \cite{vaswani2017attention}, we implement cross encoder-decoder and cross-level attention in our model. For cross encoder-decoder attention, we deploy a similar attention module where the information to be attended is from different encoder layers, and the attention maps are generated by the decoder. Similarly, for cross-stage, the attended feature is from a lower stage, and the attention decisions are made by features from a higher stage. We have observed that this helps in the propagation of information across layers and levels compared to simply passing the whole input or doing elementwise sum as done in \cite{zhang2019deep}.

\subsubsection{Pixel-Dependent Filtering Module (PDF)}
In contrast to \cite{bello2019attention}, for the local branch, we use Pixel-Dependent Filtering Module to handle spatially-varying dynamic motion blur effectively. Previous works like \cite{jia2016dynamic} generate sample-specific parameters on-the-fly using a filter generation network for image classification. \cite{li2018video} uses input text to construct the motion-generating filter weights for video generation task. \cite{zhang2018crowd} uses an adaptive convolutional layer where the convolution filter weights are the outputs of a separate “filter-manifold network” for crowd counting task. Our work is based on \cite{su2019pixel} as we use a ‘meta-layer’ to generate pixel-dependent spatially varying kernel to implement spatially variant convolution operation. Along with that, the local pixels where the filter is to be applied are also determined at runtime as we adjust the offsets of these filters adaptively. Given an input feature map $X \in  \mathbb{R}^{C \times HW}$ and a spatially varying kernel $V$, we can express the adaptive convolution operation for pixel $j$ as
\begin{equation}
    Y^{dyn}_{j,c}  = \sum_{k = 1}^{K} V_{j,j_k} W_c[j_k] x[j+j_k + \Delta j_k]
\end{equation}
where $y^{dyn}_j \in \mathbb{R}^C$, $K$ is the kernel size, $j_k \in \{(-(K-1)/2,-(K-1)/2),...,((K-1)/2,(K-1)/2)\}$ defines position of the convolutional kernel of dilation 1, $V_{j,j_k} \in \mathbb{R}^{{K^2} \times H \times W}$ is the pixel dependent kernel generated, $W_c \in \mathbb{R}^{C\times C\times K \times K}$ is the fixed weight and $\Delta j_k$ are the learnable offsets.

To extend the pixel-adaptive convolution operation with pixel-adaptive sampling step, we perform the following operations: \textbf{(i)} Given the input $X \in  \mathbb{R}^{C \times HW}$, we generate non-uniformly sampled feature tensor $X_{nu} \in  \mathbb{R}^{C \times L}$ similar to Eq. \ref{eq:xnu_att}. \textbf{(ii)} Next, we generate the pixel-adaptive offsets $\Delta' \in \mathbb{R}^{2K^2 \times L}$ as $\Delta = f_{off}(X_{nu})$. Similarly we generate the pixel-adaptive kernels $V' \in \mathbb{R}^{K^2 \times L}$ as $V = f_{ker}(X_{nu})$. $f_{off}$ and $f_{ker}$ are convolutional operations.  \textbf{(iii)} For each densely-sampled pixel $j$, we extract the neighboring patch from the original input tensor $X$ using the pixelwise offsets $\Delta'$, and generate $X_{nu}^{patch} \in \mathbb{R}^{C \times K^2 \times L}$. The filtering with adaptive weights $V'$ are performed as
\begin{equation}
\label{eq:nu_filter}
    Y^{dyn}_c = \text{sum}_2(X_{nu}^{patch} \odot V' \odot W_c)
\end{equation}
where $V$ and $W$ are expanded along $1^{st}$ and $3^{rd}$ dimensions, respectively, sum$_2$ denotes aggregation along the $2^{nd}$ dimnesion, $Y^{dyn} \in \mathbb{R}^{C \times L}$.

We set a maximum threshold $\Delta_{\text{max}}$ for the offsets to enforce efficient local processing which is important for low level tasks like deblurring.  Note that the kernels ($V'$) and offsets vary from one pixel to another, but are constant for all the channels, promoting efficiency. Standard spatial convolution can be seen as a special case of the above with adapting kernel being constant $V_{j,j_k} = 1$ and $\Delta j_k = 0$.

In contrast to \cite{bello2019attention}, which simply concatenates the output of these two branches, we design attentive fusion between these two branches so that the network can adaptively adjust the importance of each branch for each pixel at runtime. Empirically we observed that it performs better than simple addition or concatenation. Given the original input $x$ to this content-aware module, we generate a fusion mask as
\begin{equation}
    M_{fus} = sigmoid(f_{fus}(x))
\end{equation}
where $M_{fus} \in \mathbb{R}^{H \times W}$ for the first stage and $\mathbb{R}^{L}$ for the subsequent stages, $f_{fus}$ is a single convolution layer generating single channel output. Then we fuse the two branches as
\begin{equation}
    Y^{GL} = M_{fus} \odot Y^{att} + (1 - M_{fus}) \odot Y^{dyn}
\end{equation}
The fused output $y^{GL}$ contains global as well as local information distributed adaptively along pixels which helps in handling spatially-varying motion blur effectively.
\par Note that although we perform non-uniform sampling in our content-aware processing module, it does not alter the spatial correlations between pixels in the original grid. Self-attention is a non-local operation that aggregates similar global information from all over the image, irrespective of their location. Thus, it stays unaffected by the non-uniform sampling. On the other hand, convolution operation usually excels in capturing local patterns. In our PDF module, although we non-uniformly sample a critical pixel, we extract the neighboring patch around it to perform the convolution operation maintaining its local spatial correlations. Furthermore, after refinement, the pixels are restored to their original position in the uniform grid.

\vspace{-3mm}
\section{Experimental Results}
\label{sec:exp}
\subsection{Implementation Details}
\noindent \textbf{Datasets:} We follow the configuration of~\cite{zhang2019deep,kupyn2019deblurgan,tao2018scale,deblurgan,gopro2017}, which train on 2103 images from the GoPro dataset~\cite{gopro2017}. For testing, we use the following benchmarks: GoPro~\cite{gopro2017} (1111 HD images), HIDE \cite{shen2019human} (2025 HD images), RealBlur-J \cite{rim_2020_realblur} (980 images), RealBlur-R \cite{rim_2020_realblur} (980 images).

\noindent\textbf{Training Settings:} 
The hyper-parameters for our architecture are $N=3$, $M=2$, and $P=2$ (Fig. \ref{fig:arch}), and filter size in PDF modules is $5\times5$. Following \cite{zhang2019deep}, we use batch-size of $6$ and patch-size of $256\times256$. Adam optimizer~\cite{kingma2014adam} was used with initial leaning rate $10^{-4}$, halved after every $2\times10^{5}$ iterations. We use PyTorch~\cite{paszke2017automatic} library and Titan Xp GPU.

\begin{table*}[htbp] 
\centering
\caption{Performance comparisons with existing algorithms on 1111 images from the deblurring benchmark GoPro. For $^*$, the results are reported from the corresponding papers. All models are trained and tested on GoPro.}
\label{tab:quant_1}
\resizebox{\textwidth}{!}{
\begin{tabular}{|c|c|c|c|c|c|c|c|c|c|c|c|c|c|c|c|c|}
\hline
Method & \small{Xu} \cite{xu2013unnatural} & \small{Whyte} \cite{whyte2012non} &  \small{MSCNN} \cite{gopro2017} & \small{DG} \cite{deblurgan} & \small{SRN} \cite{tao2018scale} &  \small{SVRN} \cite{zhang2018dynamic} & \small{DPN} \cite{gao2019dynamic} & \small{DMPHN} \cite{zhang2019deep} & \small{DGv2} \cite{kupyn2019deblurgan} & \small{MIMO+}$^*$ \cite{cho2021rethinking} & \small{MPR}$^*$ \cite{Zamir_2021_CVPR_mprnet} & Restormer \cite{zamir2022restormer} & Uformer \cite{wang2022uformer} & Ours$_u$(a) & Ours$_u$(b) & Ours$_{nu}$ \\
\hline
PSNR (dB) & 21 & 24.6 & 29.08 & 28.7 & 30.26 & 29.19 &30.90 & 31.20 &29.55 & 32.45 & 32.66 & 32.92 & 33.06 & 31.85 &   32.02 & 32.76 \\
SSIM & 0.741 & 0.846 & 0.914 & 0.858 & 0.934 & 0.931 &0.935 &0.940 &0.934 & 0.957 & 0.959 & 0.961 & 0.967 & 0.948 & 0.953 & 0.961\\
Time (s) & 3700 & 700 & 6 & 1 & 1.2 &1 &1.0 &0.98 & 0.48 & 0.025 & 0.18 & 1.1 & 0.82 & 0.34 &  0.77 & 0.79 \\
\hline
\end{tabular}
}
\end{table*}

\begin{table}[!t]
\begin{center}
\caption{Quantitative results on test-set of HIDE~\cite{shen2019human} and RealBlur~\cite{rim_2020_realblur} datasets. All models are trained on GoPro \cite{gopro2017}.}
\label{tab:quant_2}
\setlength{\tabcolsep}{1.9pt}
\scalebox{0.9}{
\begin{tabular}{|l |c | c | c|}
\hline
 &  HIDE~\cite{shen2019human} & RealBlur-R~\cite{rim_2020_realblur} & RealBlur-J~\cite{rim_2020_realblur} \\
 \hline
 Method & PSNR~\colorbox{color6}{SSIM} & PSNR~\colorbox{color6}{SSIM} & PSNR~\colorbox{color6}{SSIM}\\
DG \cite{deblurgan}  & 24.51 \colorbox{color6}{0.871} & 33.79 \colorbox{color6}{0.903}  &  27.97 \colorbox{color6}{0.834} \\
MSCNN \cite{gopro2017}  & 25.73 \colorbox{color6}{0.874}  &  32.51 \colorbox{color6}{0.841}  &  27.87 \colorbox{color6}{0.827} \\
DGv2 \cite{deblurganv2}  & 26.61 \colorbox{color6}{0.875} & 35.26 \colorbox{color6}{0.944}  &  28.70 \colorbox{color6}{0.866} \\
SRN \cite{tao2018scale} & 28.36 \colorbox{color6}{0.915} & 35.66 \colorbox{color6}{0.947} &  28.56  \colorbox{color6}{0.867} \\
DMPHN \cite{dmphn2019}  & 29.09 \colorbox{color6}{0.924} &  35.70 \colorbox{color6}{0.948} & 28.42 \colorbox{color6}{0.860} \\
MIMO+ \cite{cho2021rethinking} & 30.00 \colorbox{color6}{0.930} & 35.54   \colorbox{color6}{0.947} & 27.63 \colorbox{color6}{0.837} \\
MPRNet \cite{Zamir_2021_CVPR_mprnet} &	30.96 \colorbox{color6}{0.939} & 35.99   \colorbox{color6}{0.952} & 28.70 \colorbox{color6}{0.873} \\
Restormer \cite{zamir2022restormer} &	31.22\colorbox{color6}{0.942} & 36.19\colorbox{color6}{0.957} & 28.96\colorbox{color6}{0.879} \\
Uformer \cite{wang2022uformer} &	30.90\colorbox{color6}{0.953} & 36.19\colorbox{color6}{0.956} & 29.09\colorbox{color6}{0.886} \\
Ours$_u$ & 29.98 \colorbox{color6}{0.930} & 35.88 \colorbox{color6}{0.950}       & 28.62 \colorbox{color6}{0.867} \\
Ours$_{nu}$ &	31.09 \colorbox{color6}{0.941} & 36.16   \colorbox{color6}{0.953} & 28.89 \colorbox{color6}{0.871} \\
\hline
\end{tabular}}
\vspace{-4mm}
\end{center}
\end{table}

\begin{table}[t]
\centering
\caption{Comparison of no. of parameters and computational complexity (GFLOPs) for a $3 \times 256 \times 256$ input.
}
\label{table:flops}
\resizebox{0.45\textwidth}{!}{
\begin{tabular}{|c|c|c|c|c|c|c|c|c|c|}
\hline
Method & DG1 & DG2 & MSCNN & DMPHN & MIMO+ & MPR & Restormer & Uformer & Ours  \\
\hline
\#params(M) & - & 60.9 & 11.7 & 21.7 & 16.1 & 20.1 & 26.12 & 50.8 & 23 \\ 
GFLOPs & 35 & 17 & 340 & 78 & 154 & 760 & 140 & 180 & 140 \\
\hline 
\end{tabular}
}
\vspace{-5mm}
\end{table}

\subsection{Performance comparisons}
The main application of our work is satisfactory deblurring of general dynamic scenes while adapting to the varying needs of different pixels interpretably. Due to the complexity of the blur present in such images, conventional image formation model based deblurring approaches struggle to perform well. Hence, we compare with only two conventional methods \cite{whyte2012non,xu2013unnatural} (which are selected as representative traditional methods for non-uniform deblurring, with publicly available implementations). We provide extensive comparisons with state-of-the-art learning-based methods, namely MSCNN \cite{gopro2017}, DG \cite{deblurgan}, DGv2 \cite{kupyn2019deblurgan}, SRN \cite{tao2018scale},  DPN \cite{gao2019dynamic} , SVRN \cite{zhang2018dynamic}, Stack(4)-DMPHN \cite{zhang2019deep}, MIMO+ \cite{cho2021rethinking}, MPR \cite{Zamir_2021_CVPR_mprnet}, Restormer \cite{zamir2022restormer}, Uformer \cite{wang2022uformer}. We use the official implementation or results from the authors.

\textbf{Quantitative Evaluation}
We show performance comparisons on three different benchmark datasets. The quantitative results on GoPro, HIDE Dataset \cite{shen2019human} and RealBlur~\cite{rim_2020_realblur} testing set are listed in Tables \ref{tab:quant_1} and \ref{tab:quant_2}. To verify the generalizability of our approach, we train our model on GoPro and test it on multiple benchmarking datasets. Recently, HIDE dataset \cite{shen2019human} was proposed, which was specifically designed for human-centric motions. The images are captured using high-framerate cameras, similar to \cite{gopro2017}. It covers a wide range of daily scenes, various human motions, sizes, etc. To handle more realistic scenarios, \cite{rim_2020_realblur} proposed RealBlur dataset. Instead of generating the blurred images from a high-frame-rate video like \cite{gopro2017,shen2019human}, the authors simultaneously captured geometrically aligned pairs of blurred and sharp images using a custom-built image acquisition system. The test images are split into two parts: RealBlur-R and RealBlur-J, generated from the captured raw and JPEG images, respectively. We compare against all existing models trained on GoPro train-set for fair comparisons.

As shown in Table \ref{tab:quant_2}, our approach shows impressive performance compared to most of the existing methods. We achieve 0.31/0.1/0.74 dB improvement on GoPro, 1.09/0.13/1.11 dB improvement on HIDE, 0.49/0.17/0.28 dB improvement on RealBlur-R, 1.26/0.19/0.27 dB improvement on RealBlur-J, against state-of-the-art MIMO+ \cite{cho2021rethinking}/MPR \cite{Zamir_2021_CVPR_mprnet}/Ours previous approach (uniform-sampling). The superiority of our model is owed to the robustness of the proposed adaptive modules coupled with the pixel-adaptive sampling strategy. A very recent work Restormer \cite{zamir2022restormer} is a pure transformer model that utilizes larger training patch sizes for a certain period of training. Thus, it is difficult to compare its performance with ours directly. It achieves 0.03 to 0.16 dB PSNR improvement while having similar computational complexity. On the other hand, Uformer \cite{wang2022uformer} is a significantly larger model with 2x more parameters and 1.3x more computation (GFlops) while achieving better accuracy. Our approach achieves a better balance between training and testing complexity, parameters, and restoration accuracy with an adaptive hybrid model. In the future, instead of a pure CNN (MPR, MIMO) or pure transformer model (Restormer, Uformer), our global-local approach should help devise better hybrid models and utilize the best from both domains. Furthermore, both type of networks can be equally benefitted by the adaptive sampling strategy introduced in this work.

To compare the efficiency, the inference times are mentioned in Table \ref{tab:quant_1}. Our previous model with uniform sampling and the upgraded model with pixel-adaptive non-uniform sampling takes similar time to process an HD image from GoPro \cite{gopro2017}. Our approach is faster than most of the existing works, except \cite{Zamir_2021_CVPR_mprnet,cho2021rethinking}. Works like MPRNet or MIMO+ are purely convolutional and can easily exploit standard accelerated PyTorch libraries. But, our design requires a few custom operations, such as pixel-adaptive convolution or adaptive sampling. As per the latest PyTorch version (1.13), such custom operations are less-optimized than the standard convolutional layer. Due to the implementation-related bottleneck, our runtime is slower. But, our GFLOPs (theoretical computational complexity) are significantly less or equivalent to existing SOTA works. Thus, in future, with the availability of more optimized deep learning libraries, our model can be executed much faster. We have also reported our model sizes in Table \ref{table:flops}, which is comparable to most SOTA works.
\newline We also compare the computational complexity, in terms of GFLOPs in Table \ref{table:flops}. We can observe that the closest competitor MPR \cite{Zamir_2021_CVPR_mprnet} is expensive (almost $5 \times $) in terms of computation. MIMO+ \cite{cho2021rethinking} shows comparable complexity, but is inferior in terms of restoration quality. Overall, our approach achieves a good balance between restoration quality, efficiency and interpretable operations. We further apply our adaptive sampling strategy to some of the previous SOTA networks, such as SRN and DMPHN. As reported in Table \ref{table:extension}, the significant boost in performance demonstrates the utility of the non-uniform sampling operation, and it can be potentially used as a plug-and-play module to boost other networks' performance as well.
\newline \textbf{Qualitative Evaluation:}
Visual comparisons on a wide range of dynamic scenes from GoPro, HIDE, RealBlur-J and RealBlur-R datasets are shown in Figs. \ref{fig:qual_gopro}, \ref{fig:qual_hide}, \ref{fig:qual_realj} and \ref{fig:qual_realr}, respectively. We observe that the results of prior works suffer from incomplete deblurring or artifacts. In contrast, our network is able to restore scene details more faithfully which are noticeable in the regions containing text, edges, texture, etc. Although the first two stages of our network divide the image into multiple non-overlapping patches, smaller patches at the lower levels constitute a bigger patch at the upper level and are processed combinedly in our hierarchical architecture. At the final level, we process the whole image/feature map. Thus, we do not observe any blocking artifacts in our final output, similar to \cite{zhang2019deep, Zamir_2021_CVPR_mprnet}. Also, an additional advantage over \cite{hyun2013dynamic,whyte2012non} is that our model does not require parameter tuning during test phase.

 \newcommand\gpw{0.16}
\begin{figure*}[htb]
\captionsetup[subfigure]{labelformat=empty}
     \centering
     \begin{subfigure}[b]{\gpw\textwidth}
         \centering
         \includegraphics[width=\textwidth]{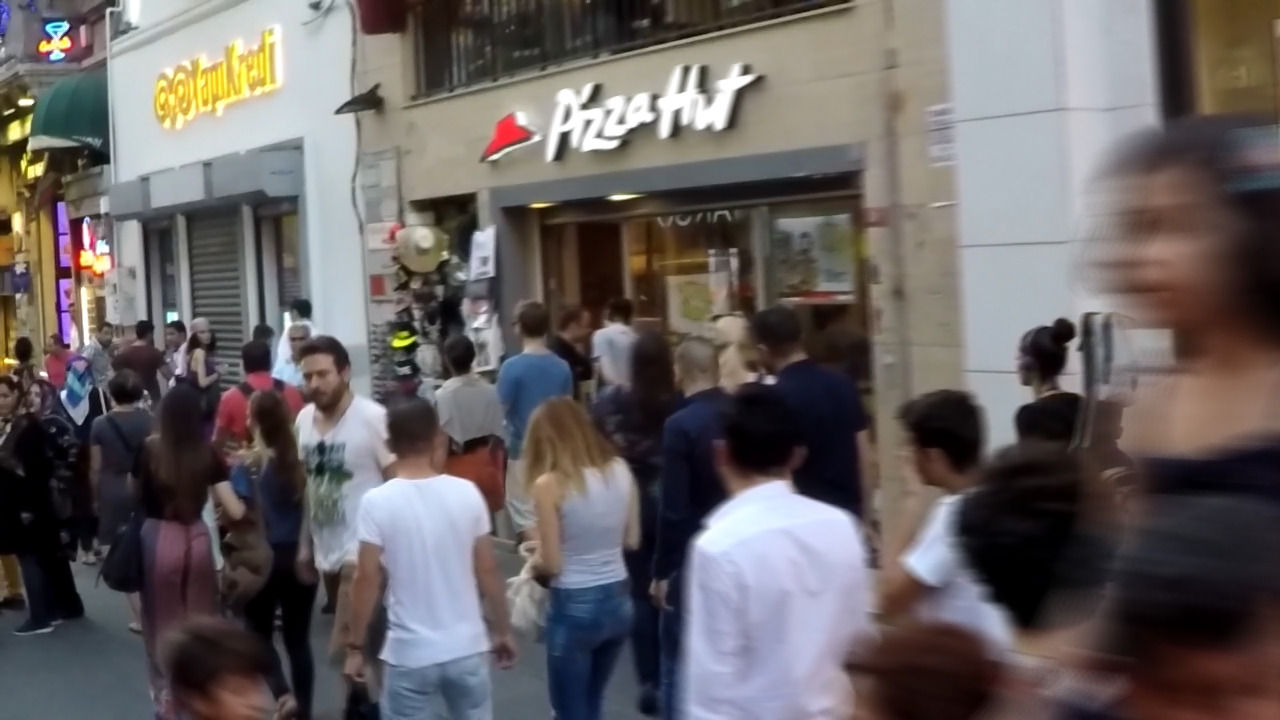}
         \caption{(a) IP}
     \end{subfigure}
     \hfill
     \begin{subfigure}[b]{\gpw\textwidth}
         \centering
         \includegraphics[width=\textwidth]{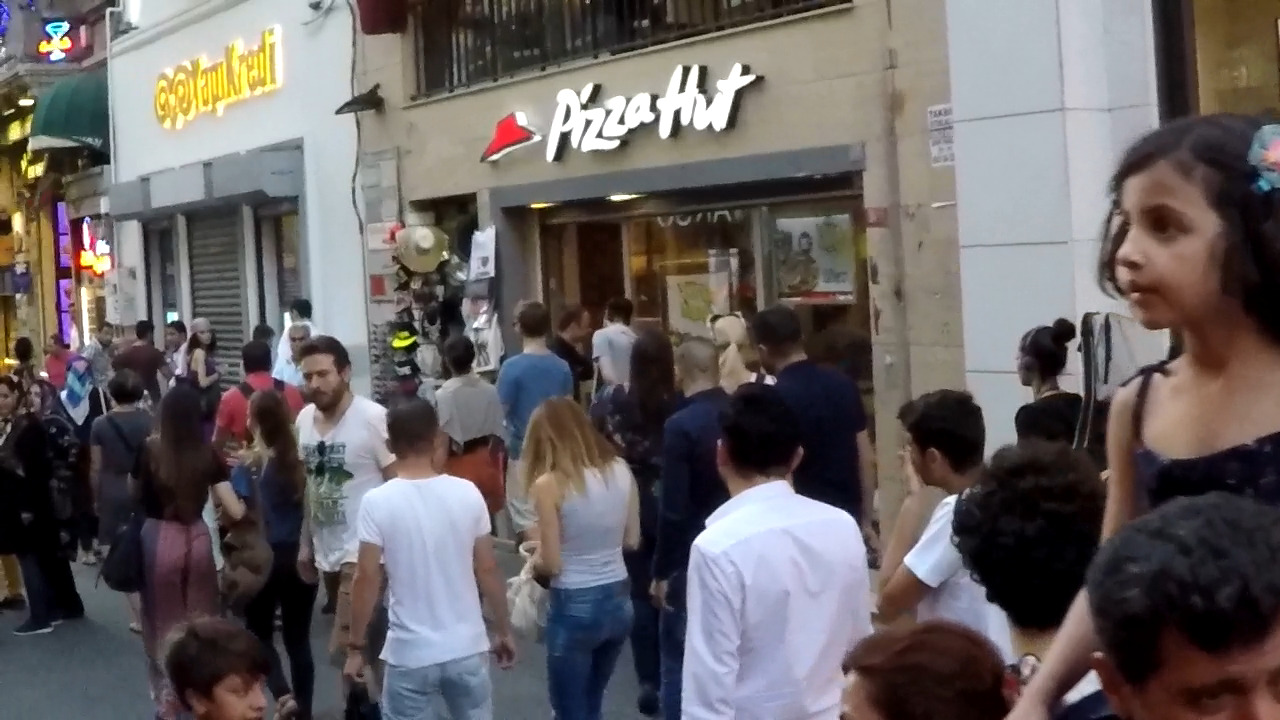}
         \caption{(a) GT}
     \end{subfigure}
     \hfill
     \begin{subfigure}[b]{\gpw\textwidth}
         \centering
         \includegraphics[bb=503 223 1280 661,clip=True,width=\textwidth]{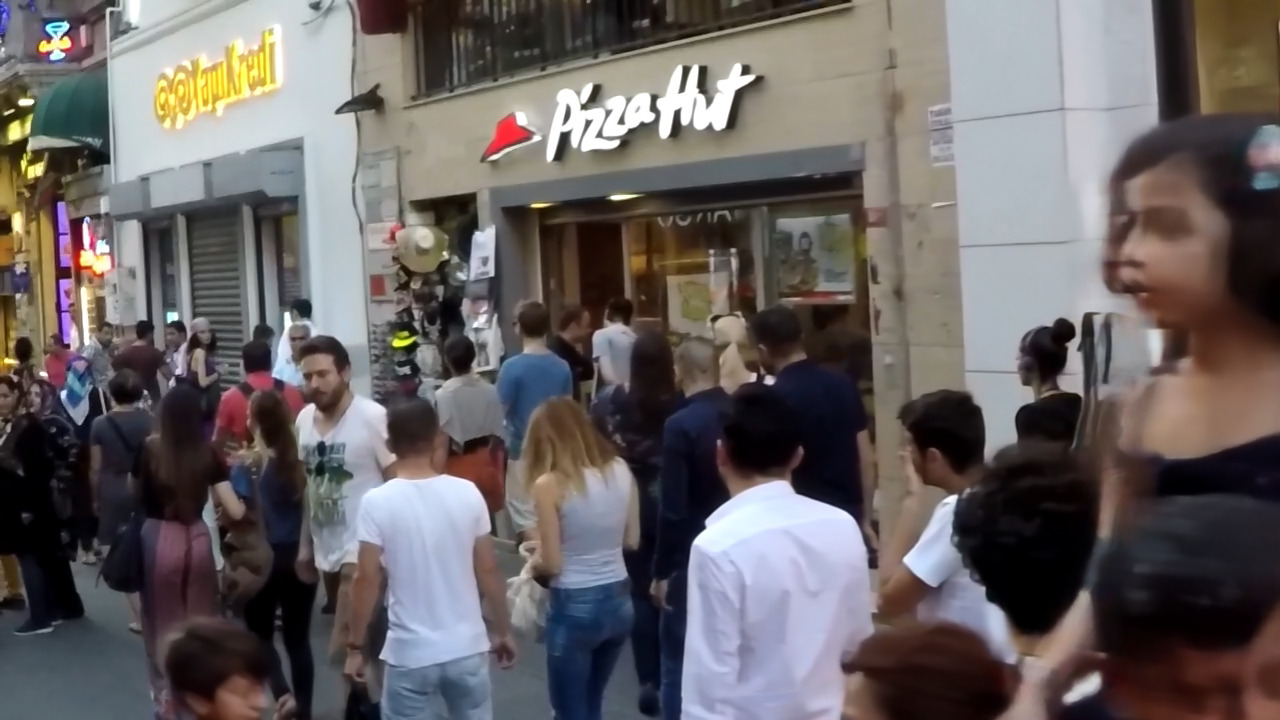}
         \caption{(a) Nah}
     \end{subfigure}
     \hfill
     \begin{subfigure}[b]{\gpw\textwidth}
         \centering
         \includegraphics[bb=503 223 1280 661,clip=True,width=\textwidth]{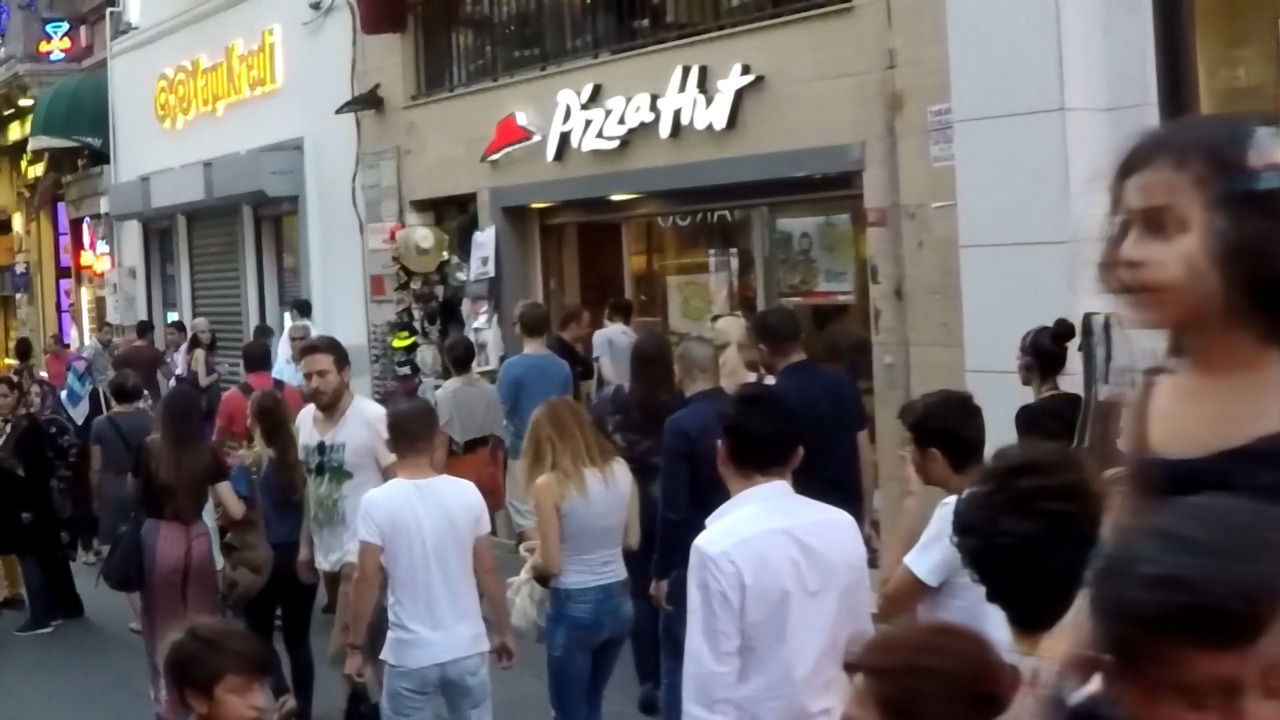}
         \caption{(a) SRN}
     \end{subfigure}
     \hfill
     \begin{subfigure}[b]{\gpw\textwidth}
         \centering
         \includegraphics[bb=503 223 1280 661,clip=True,width=\textwidth]{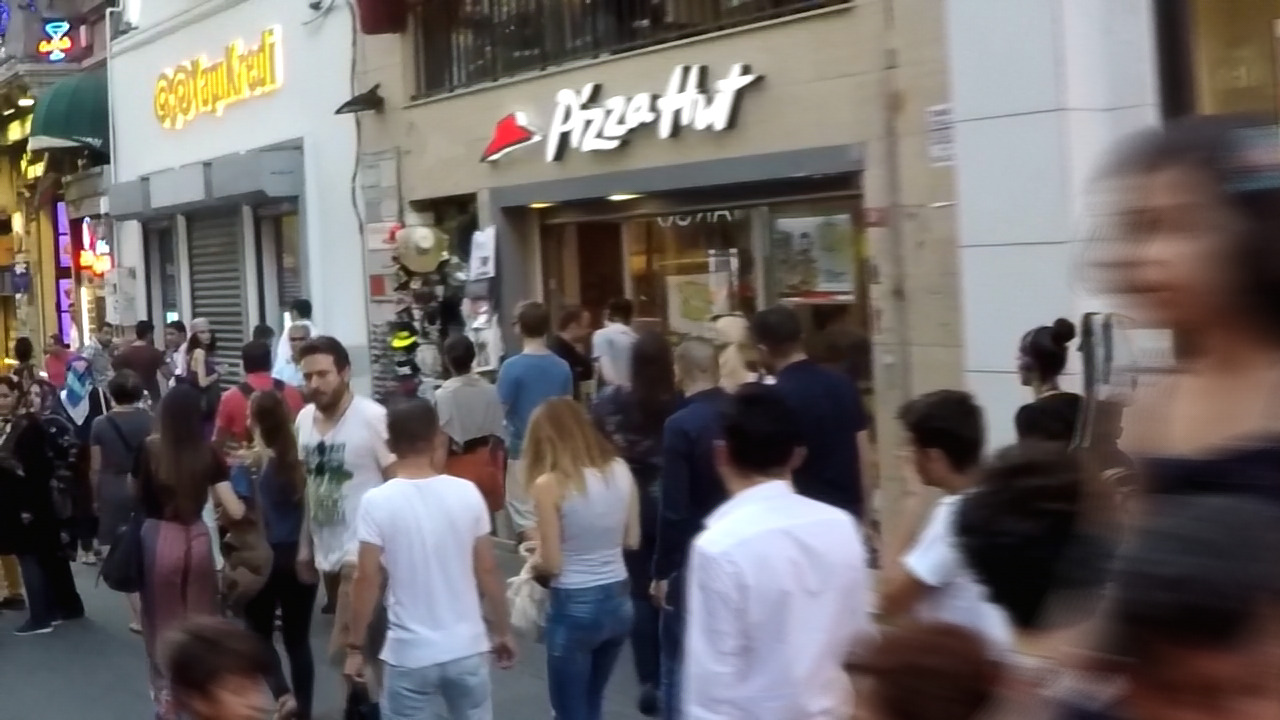}
         \caption{(a) DG1}
     \end{subfigure}
     \hfill
     \begin{subfigure}[b]{\gpw\textwidth}
         \centering
         \includegraphics[bb=503 223 1280 661,clip=True,width=\textwidth]{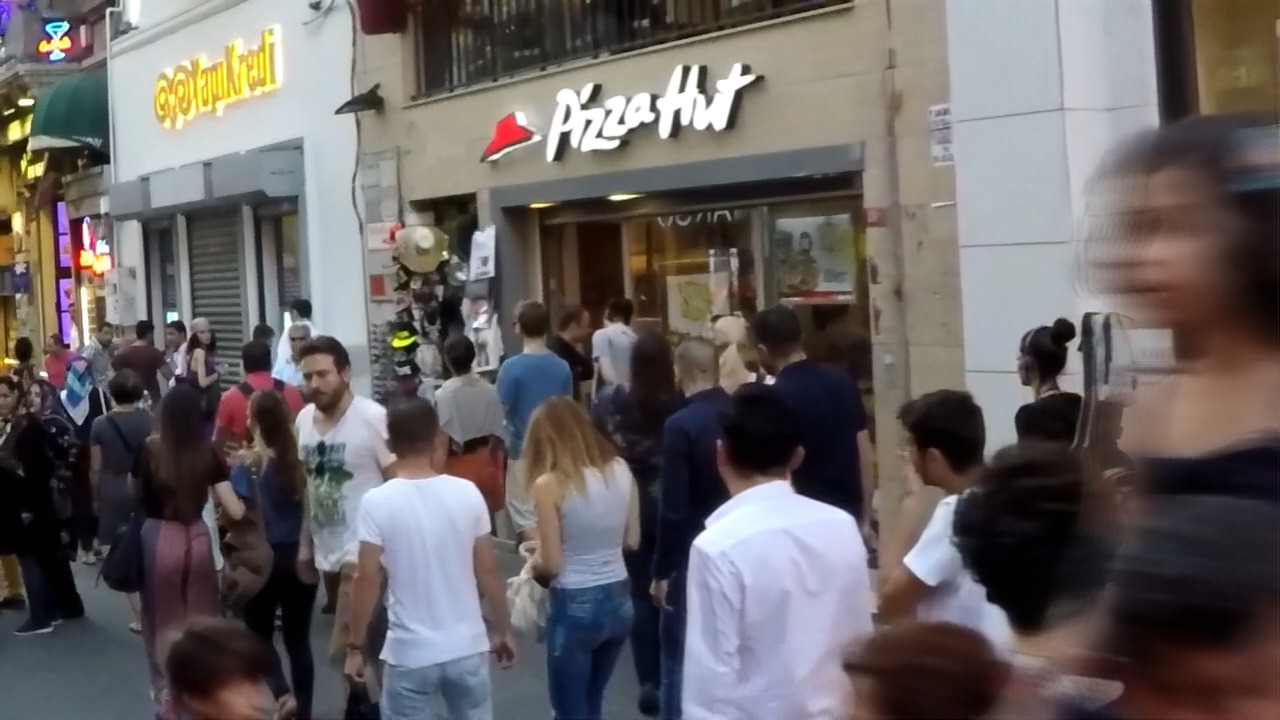}
         \caption{(a) DG2}
     \end{subfigure}

     \begin{subfigure}[b]{\gpw\textwidth}
         \centering
         \includegraphics[bb=503 223 1280 661,clip=True,width=\textwidth]{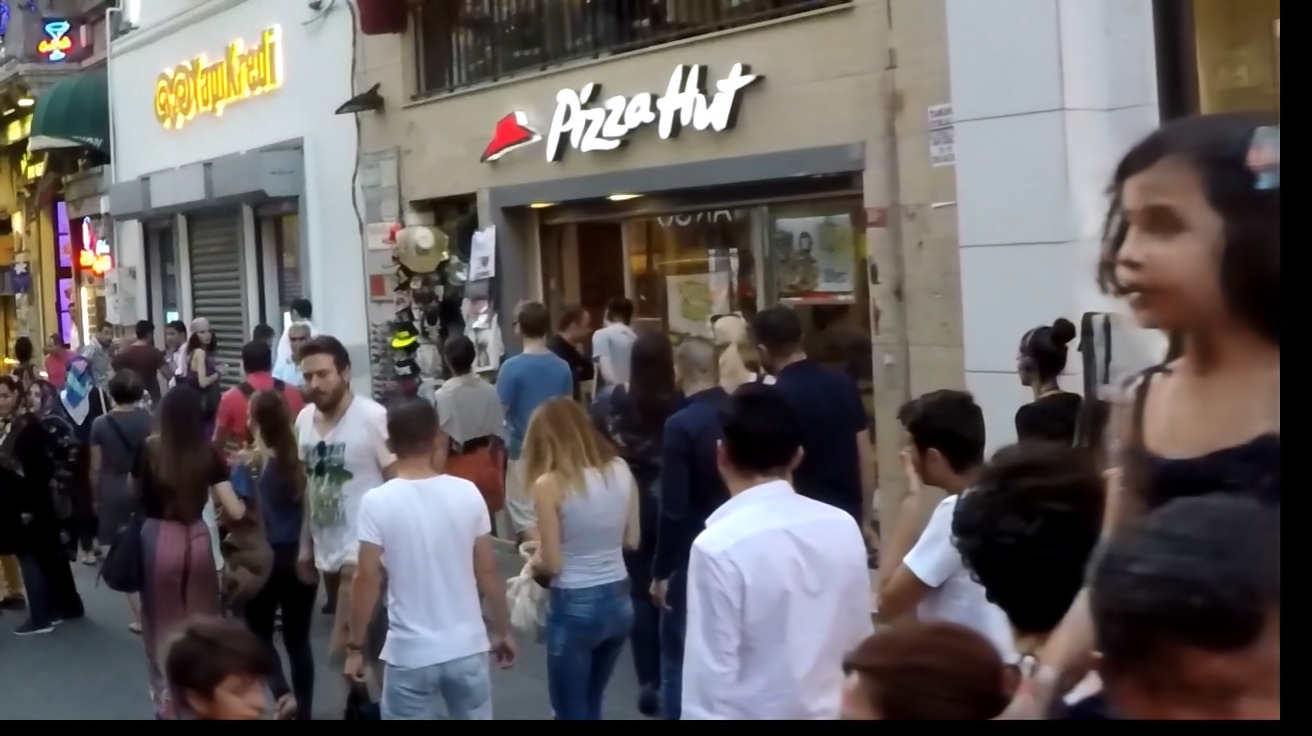}
         \caption{(a) DMPHN}
     \end{subfigure}
     \hfill 
     \begin{subfigure}[b]{\gpw\textwidth}
         \centering
         \includegraphics[bb=503 223 1280 661,clip=True,width=\textwidth]{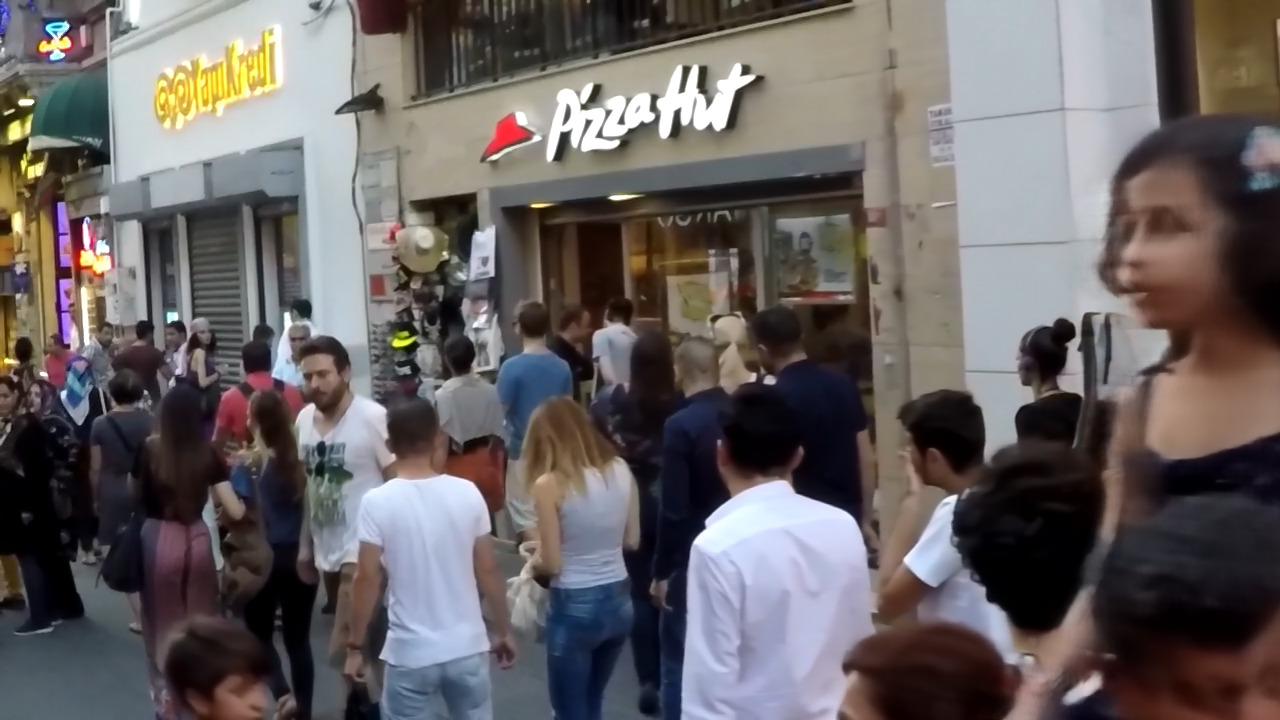}
         \caption{(a) MPR}
     \end{subfigure}
     \hfill
     \begin{subfigure}[b]{\gpw\textwidth}
         \centering
         \includegraphics[bb=503 223 1280 661,clip=True,width=\textwidth]{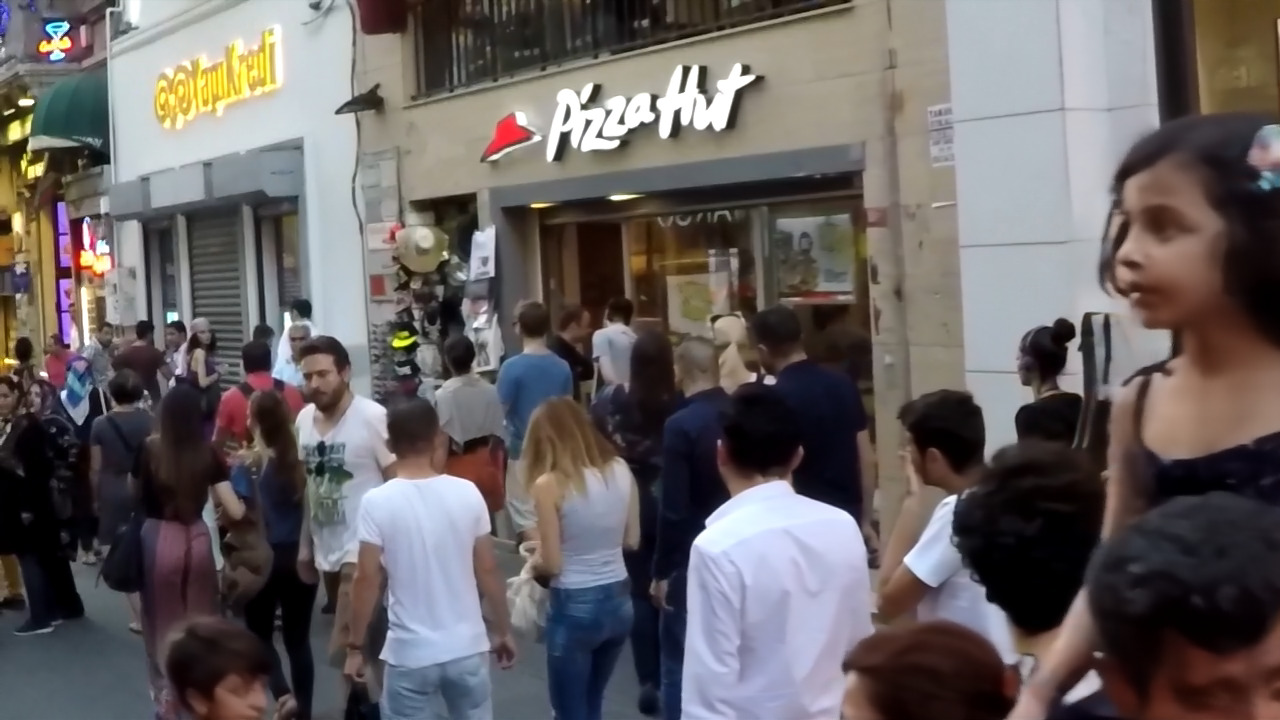}
         \caption{(a) Uform.}
     \end{subfigure}
     \hfill
     \begin{subfigure}[b]{\gpw\textwidth}
         \centering
         \includegraphics[bb=503 223 1280 661,clip=True,width=\textwidth]{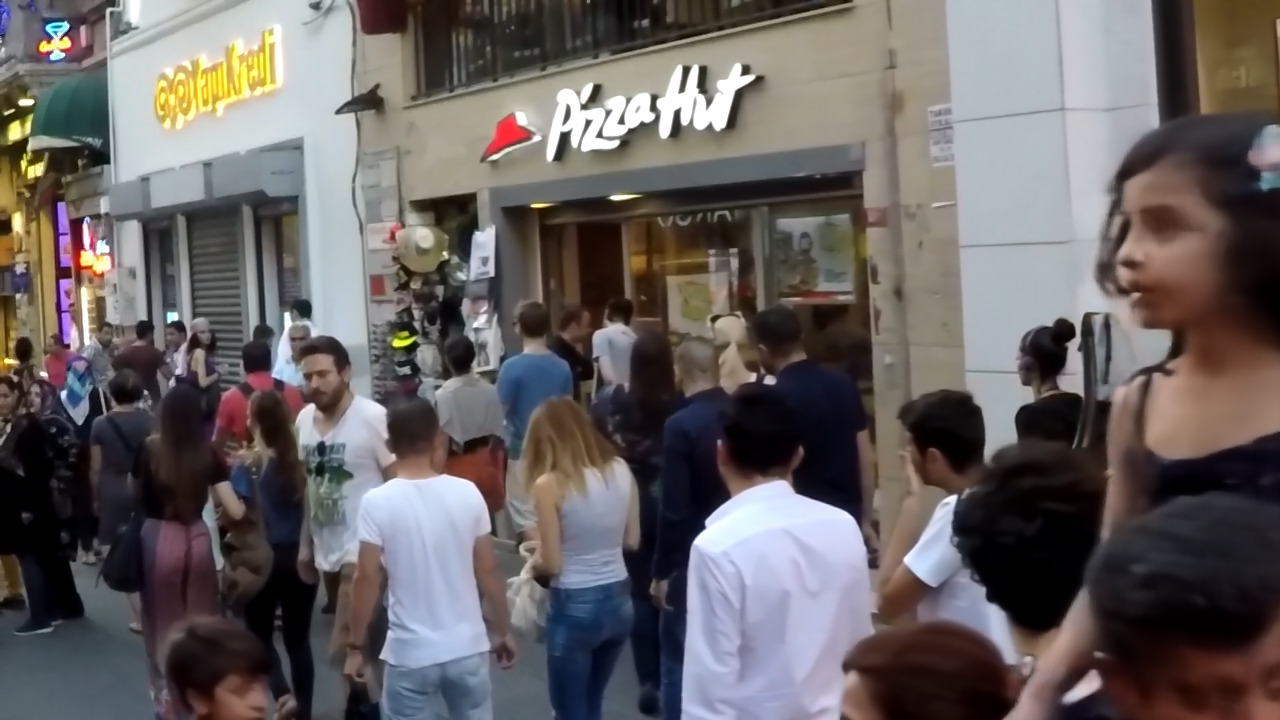}
         \caption{(a) Rest.}
     \end{subfigure}
     \hfill
     \begin{subfigure}[b]{\gpw\textwidth}
         \centering
         \includegraphics[bb=503 223 1280 661,clip=True,width=\textwidth]{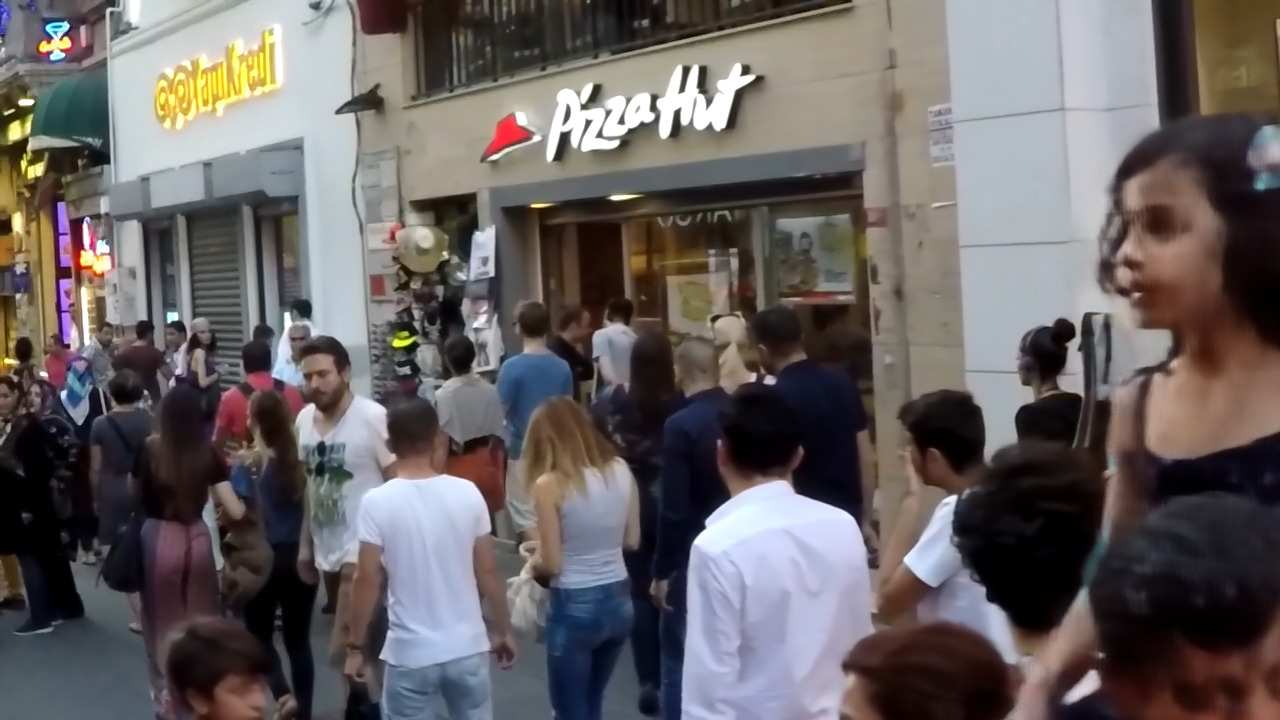}
         \caption{(a) Ours$_U$}
     \end{subfigure}
     \hfill
     \begin{subfigure}[b]{\gpw\textwidth}
         \centering
         \includegraphics[bb=503 223 1280 661,clip=True,width=\textwidth]{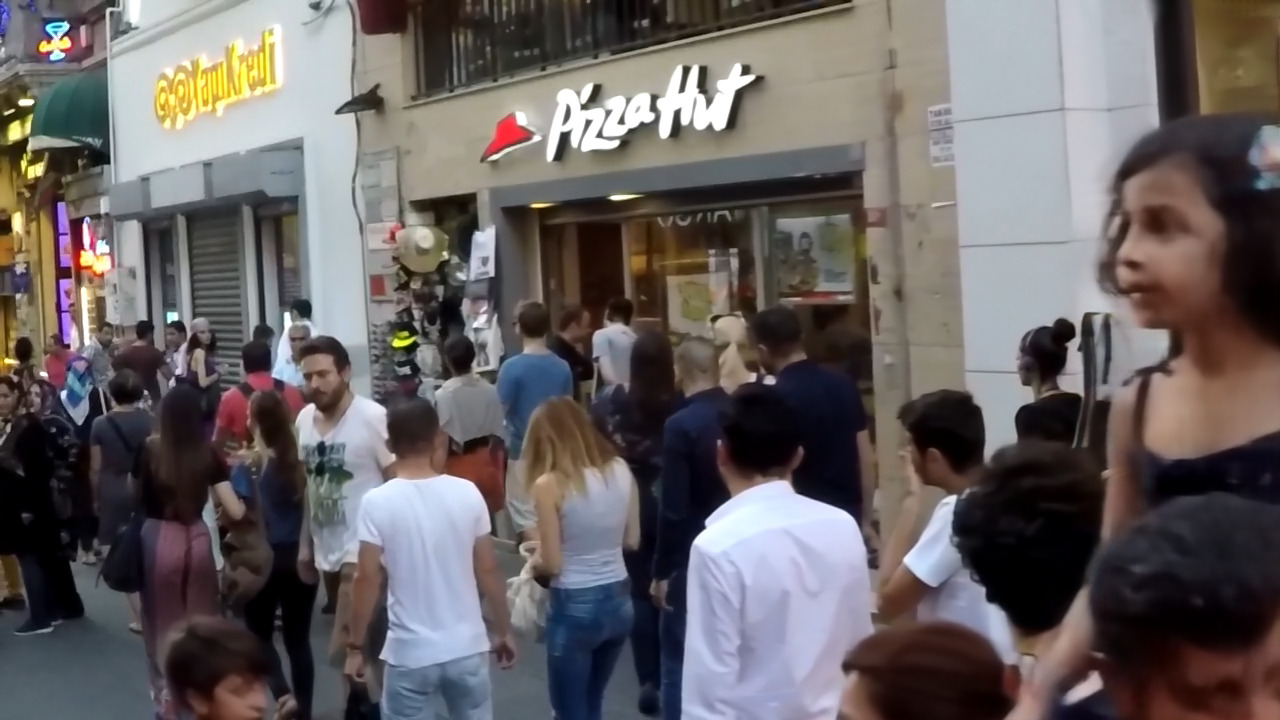}
         \caption{(a) Ours$_{NU}$}
     \end{subfigure}
     
     \begin{subfigure}[b]{\gpw\textwidth}
         \centering
         \includegraphics[width=\textwidth]{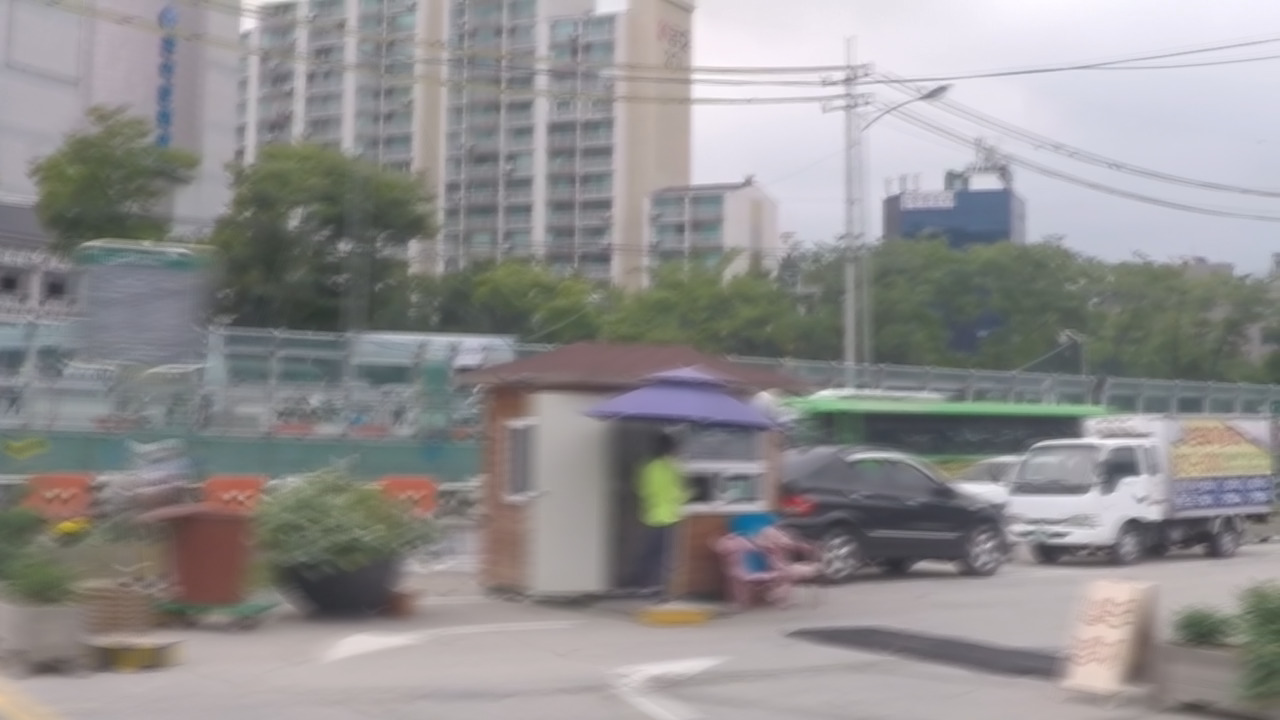}
         \caption{(b) IP}
     \end{subfigure}
     \hfill
     \begin{subfigure}[b]{\gpw\textwidth}
         \centering
         \includegraphics[width=\textwidth]{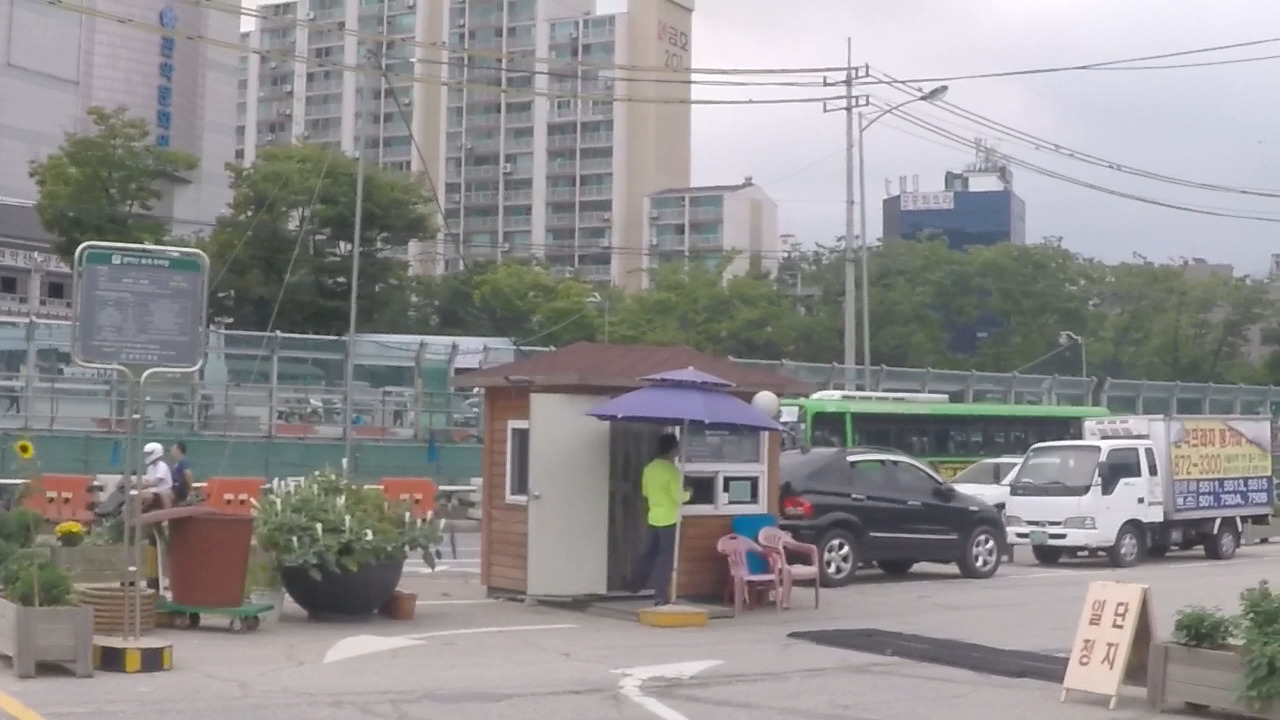}
         \caption{(b) GT}
     \end{subfigure}
     \hfill
     \begin{subfigure}[b]{\gpw\textwidth}
         \centering
         \includegraphics[bb=980 160 1280 329,clip=True,width=\textwidth]{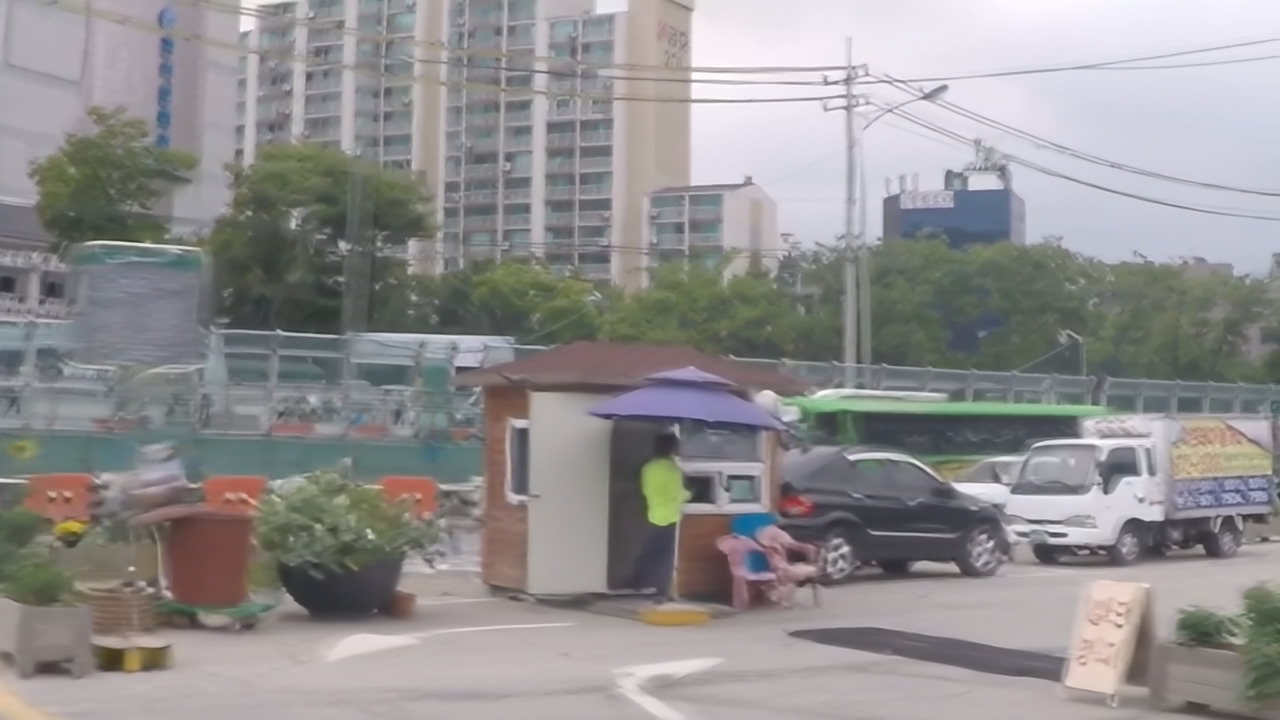}
         \caption{(b) Nah}
     \end{subfigure}
     \hfill
     \begin{subfigure}[b]{\gpw\textwidth}
         \centering
         \includegraphics[bb=980 160 1280 329,clip=True,width=\textwidth]{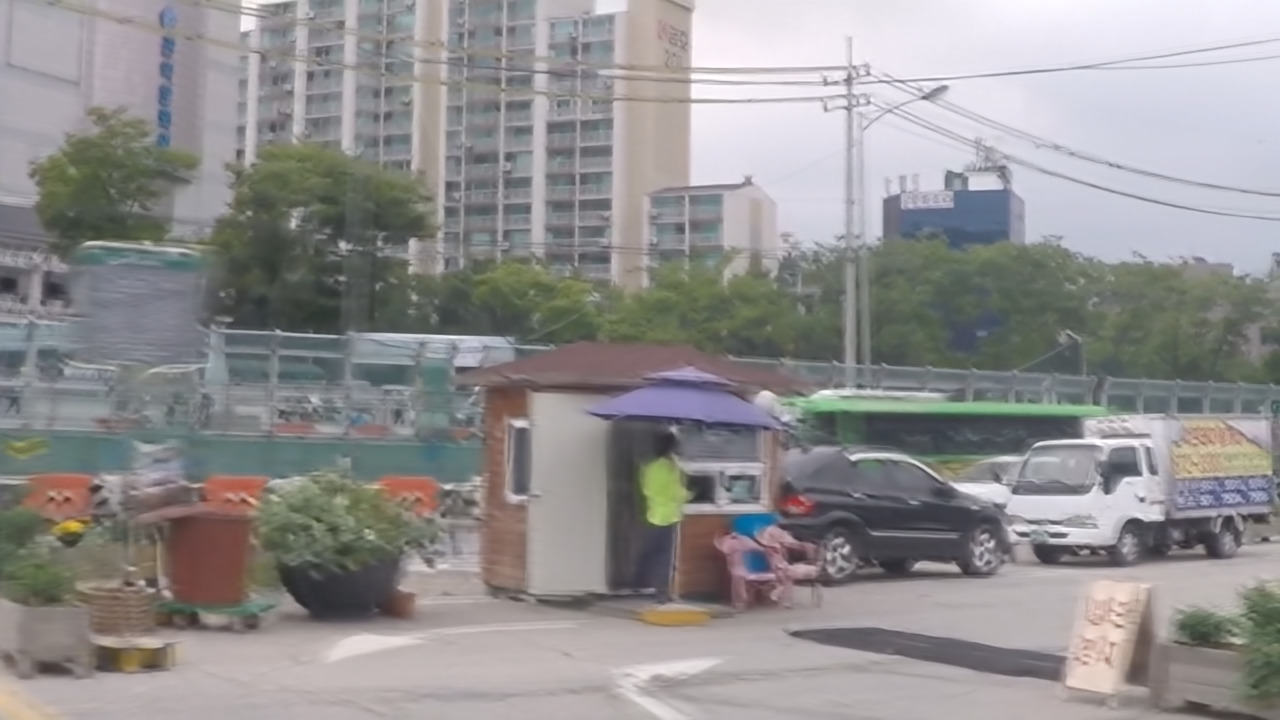}
         \caption{(b) SRN}
     \end{subfigure}
     \hfill
     \begin{subfigure}[b]{\gpw\textwidth}
         \centering
         \includegraphics[bb=980 160 1280 329,clip=True,width=\textwidth]{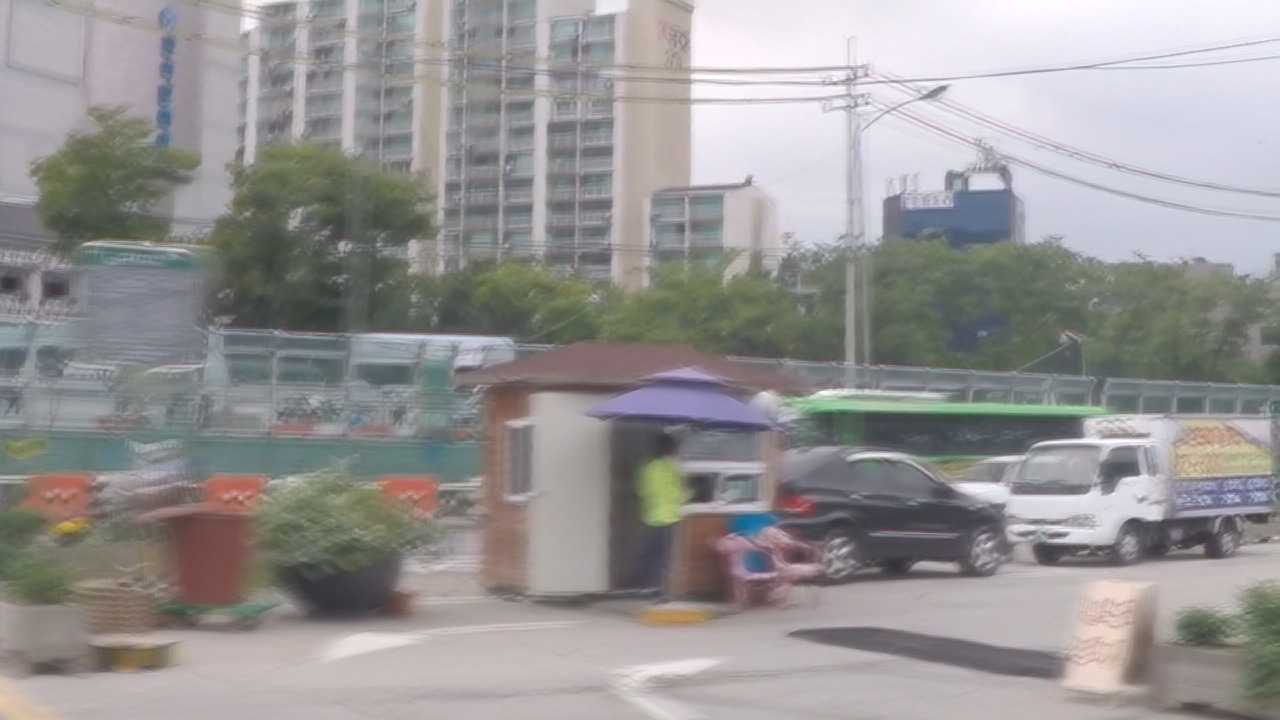}
         \caption{(b) DG1}
     \end{subfigure}
     \hfill
     \begin{subfigure}[b]{\gpw\textwidth}
         \centering
         \includegraphics[bb=980 160 1280 329,clip=True,width=\textwidth]{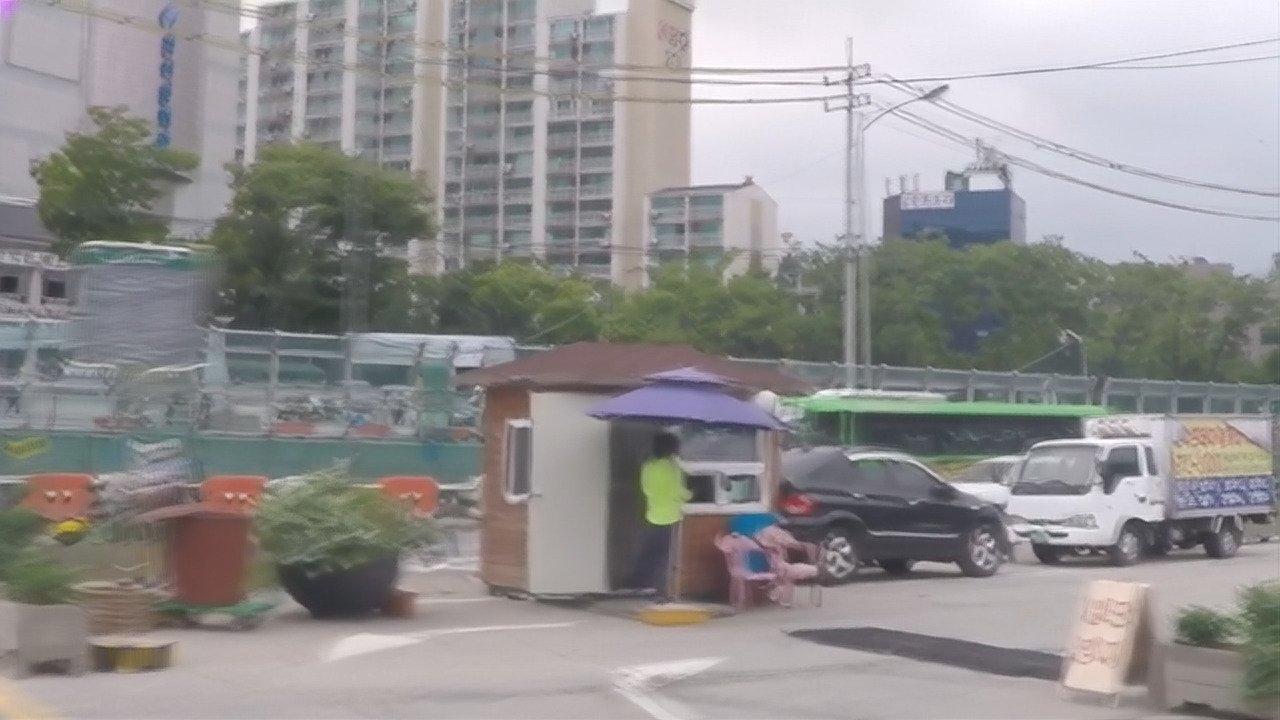}
         \caption{(b) DG2}
     \end{subfigure}
     
     \begin{subfigure}[b]{\gpw\textwidth}
         \centering
         \includegraphics[bb=980 160 1280 329,clip=True,width=\textwidth]{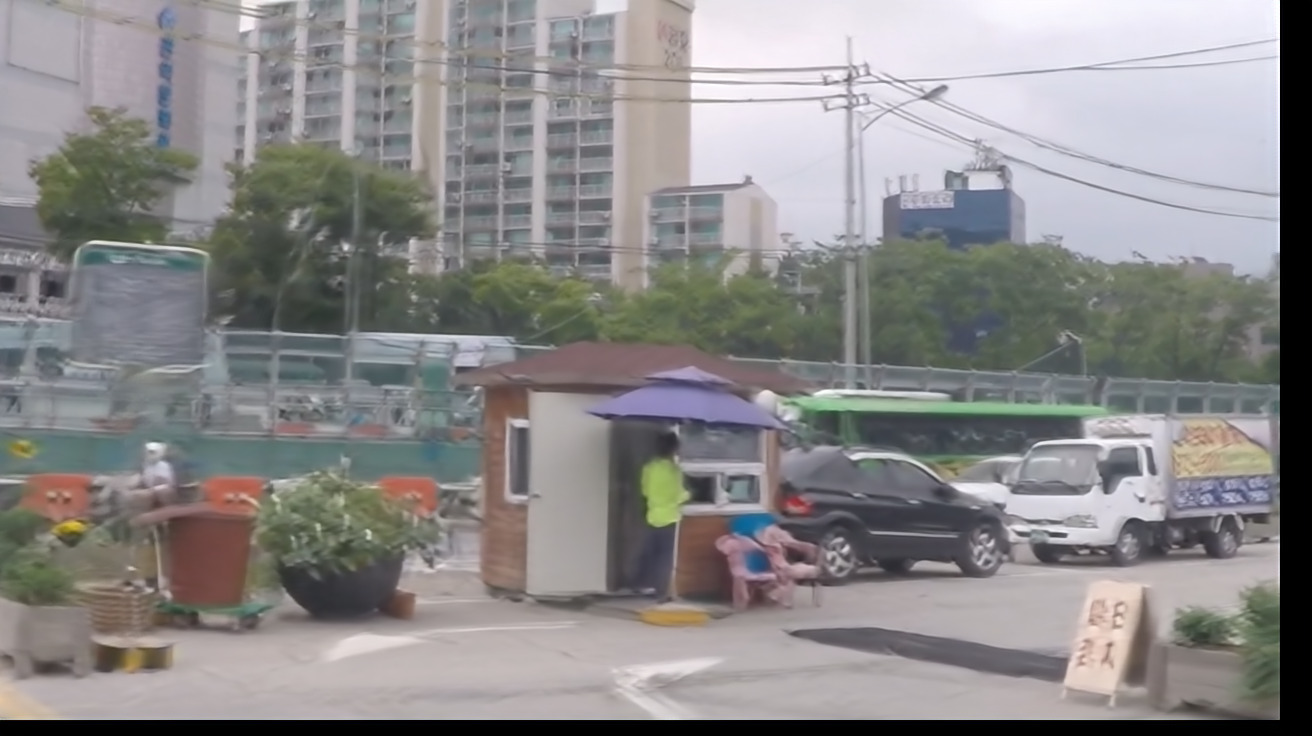}
         \caption{(b) DMPHN}
     \end{subfigure}
     \hfill 
     \begin{subfigure}[b]{\gpw\textwidth}
         \centering
         \includegraphics[bb=980 160 1280 329,clip=True,width=\textwidth]{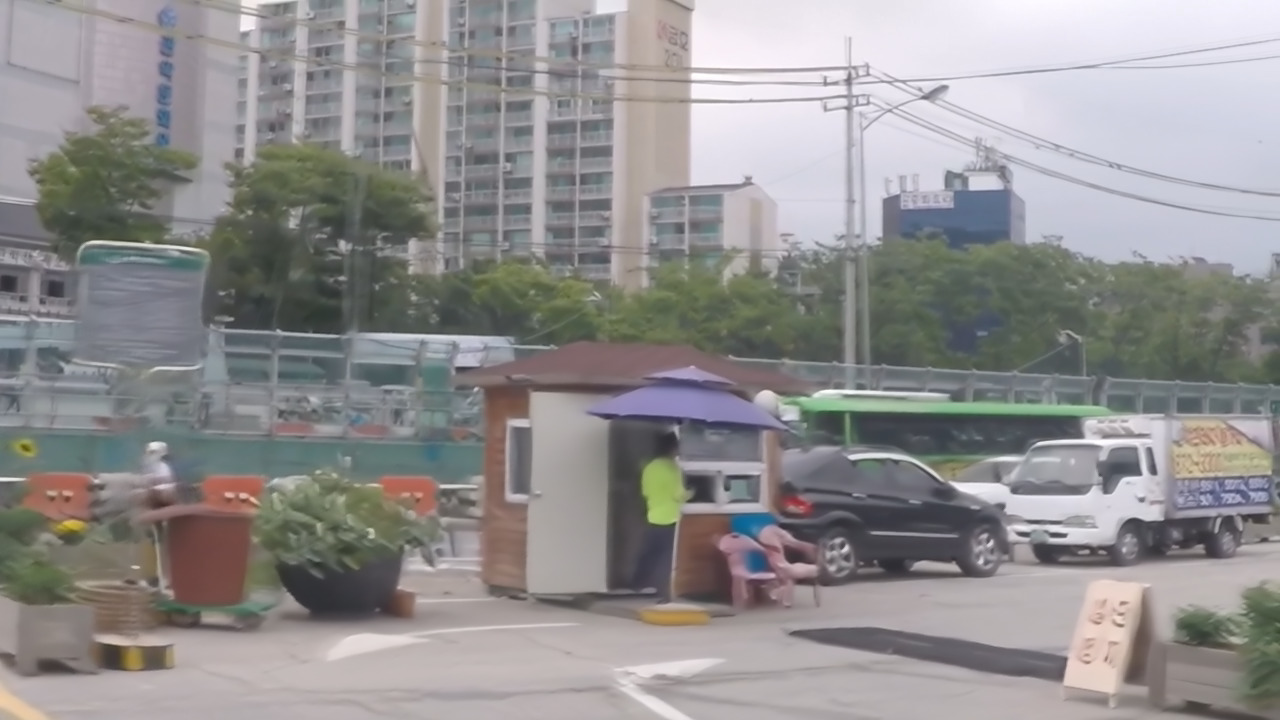}
         \caption{(b) MPR}
     \end{subfigure}
     \hfill
      \begin{subfigure}[b]{\gpw\textwidth}
         \centering
         \includegraphics[bb=980 160 1280 329,clip=True,width=\textwidth]{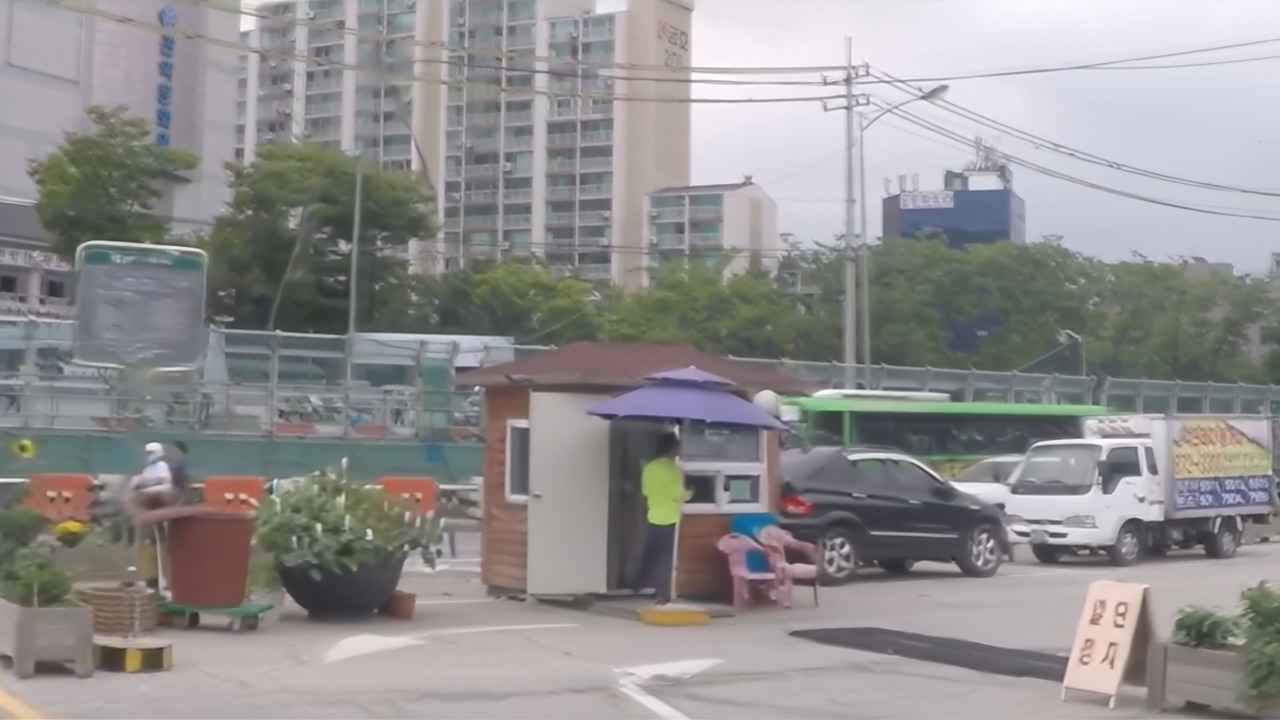}
         \caption{(b) Uform.}
     \end{subfigure}
     \hfill
      \begin{subfigure}[b]{\gpw\textwidth}
         \centering
         \includegraphics[bb=980 160 1280 329,clip=True,width=\textwidth]{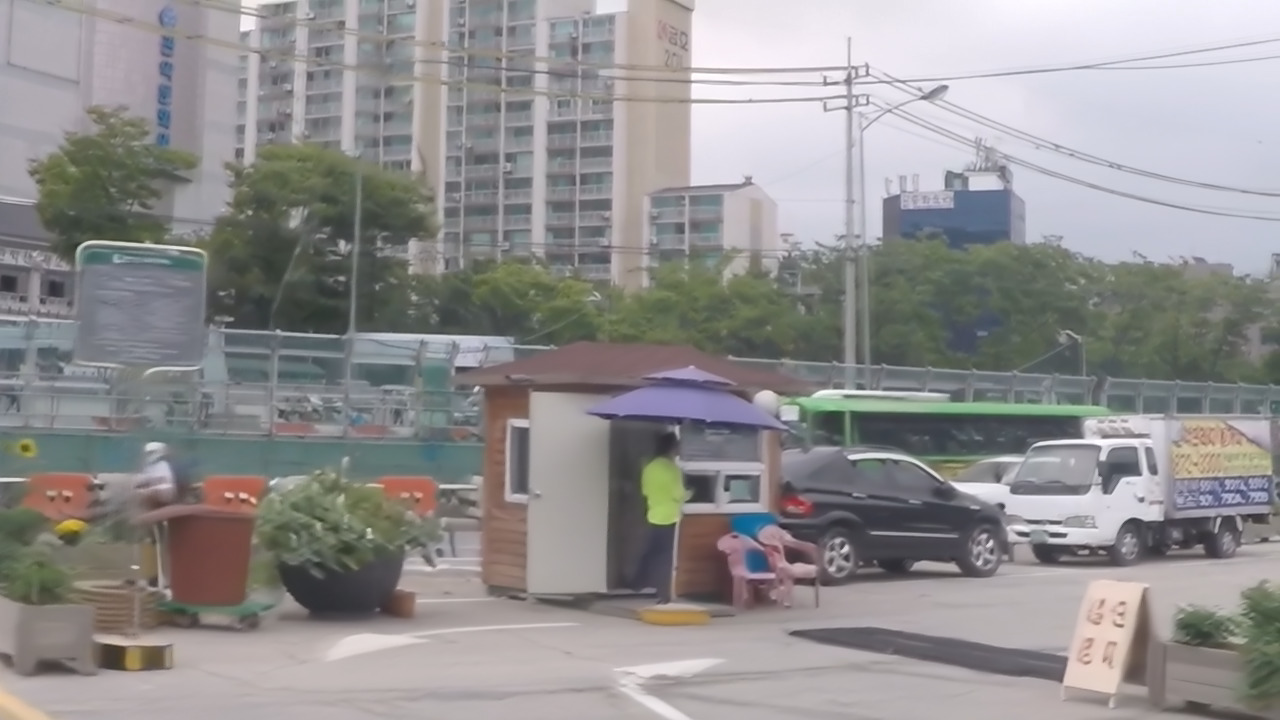}
         \caption{(b) Rest.}
     \end{subfigure}
     \hfill
     \begin{subfigure}[b]{\gpw\textwidth}
         \centering
         \includegraphics[bb=980 160 1280 329,clip=True,width=\textwidth]{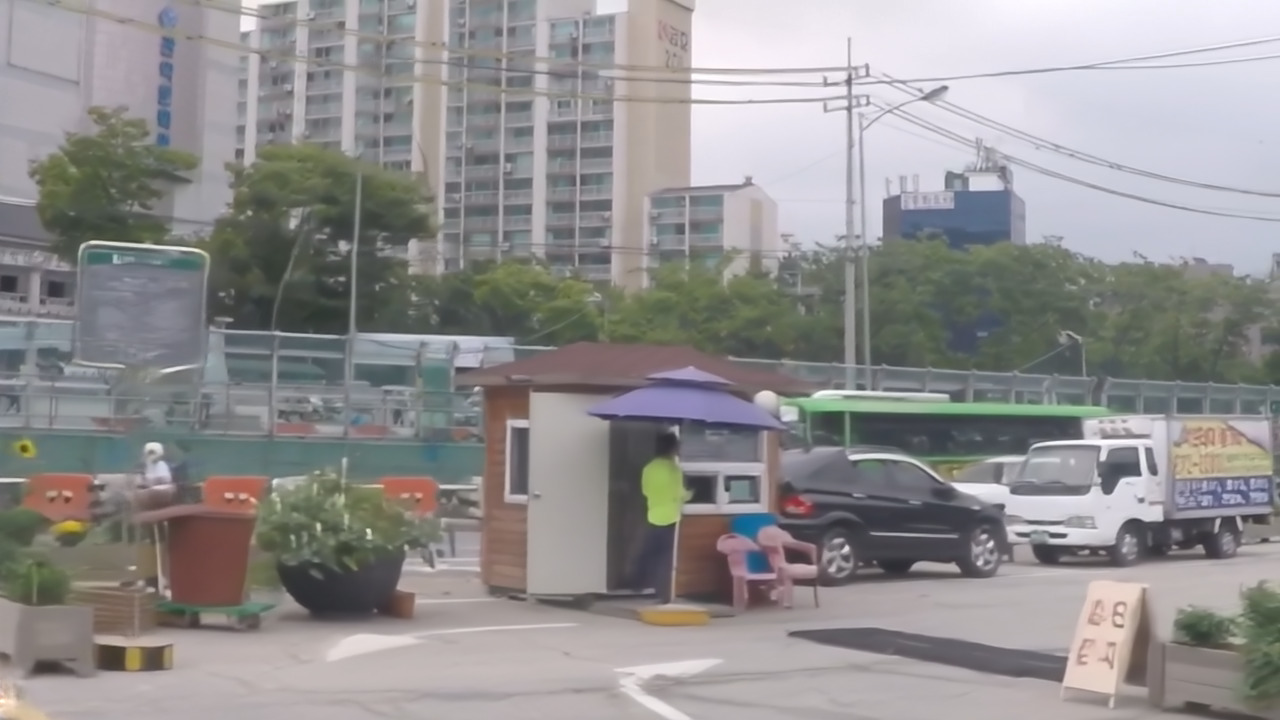}
         \caption{Ours$_U$}
     \end{subfigure}
     \hfill
     \begin{subfigure}[b]{\gpw\textwidth}
         \centering
         \includegraphics[bb=980 160 1280 329,clip=True,width=\textwidth]{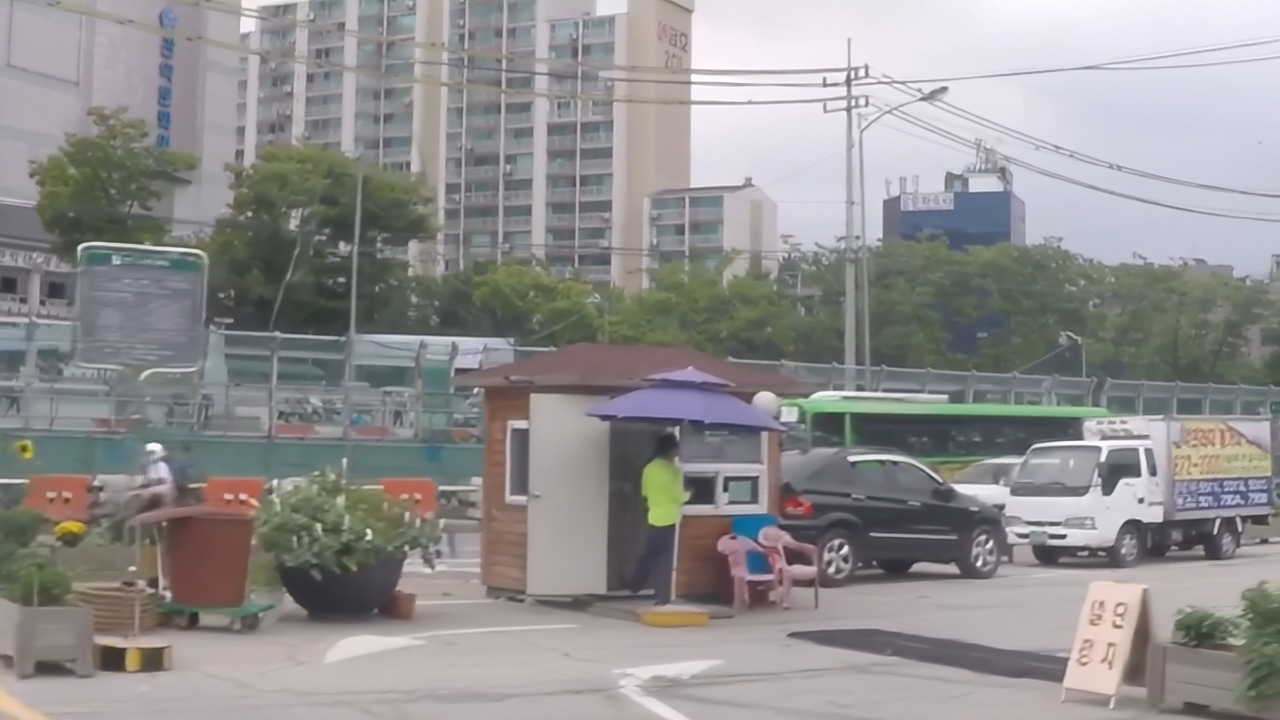}
         \caption{Ours$_{NU}$}
     \end{subfigure}
     \hfill 
     
        \caption{Visual comparisons of deblurring results on images from the GoPro test set~\cite{gopro2017}.}
        \label{fig:qual_gopro}
\end{figure*}


\begin{figure*}[!htb]
\captionsetup[subfigure]{labelformat=empty}
     \centering
     \begin{subfigure}[b]{\gpw\textwidth}
         \centering
         \includegraphics[trim={500 160 200 120},clip,width=\textwidth]{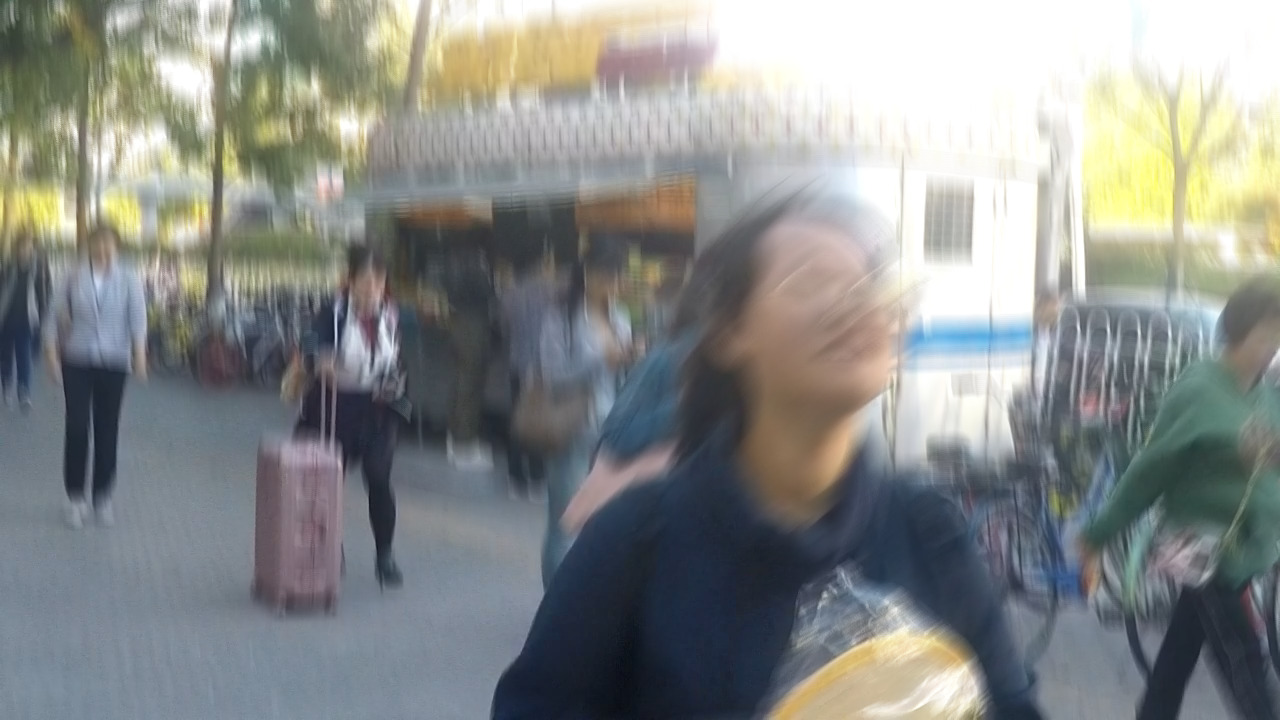}
         \caption{(a) IP}
     \end{subfigure}
     \hfill
     \begin{subfigure}[b]{\gpw\textwidth}
         \centering
         \includegraphics[trim={500 160 200 120},clip,width=\textwidth]{results/hide/set3/12fromGOPR1089.MP4_ip.jpg}
         \caption{(a) GT}
     \end{subfigure}
     \hfill
     \begin{subfigure}[b]{\gpw\textwidth}
         \centering
         \includegraphics[trim={500 160 200 120},clip,width=\textwidth]{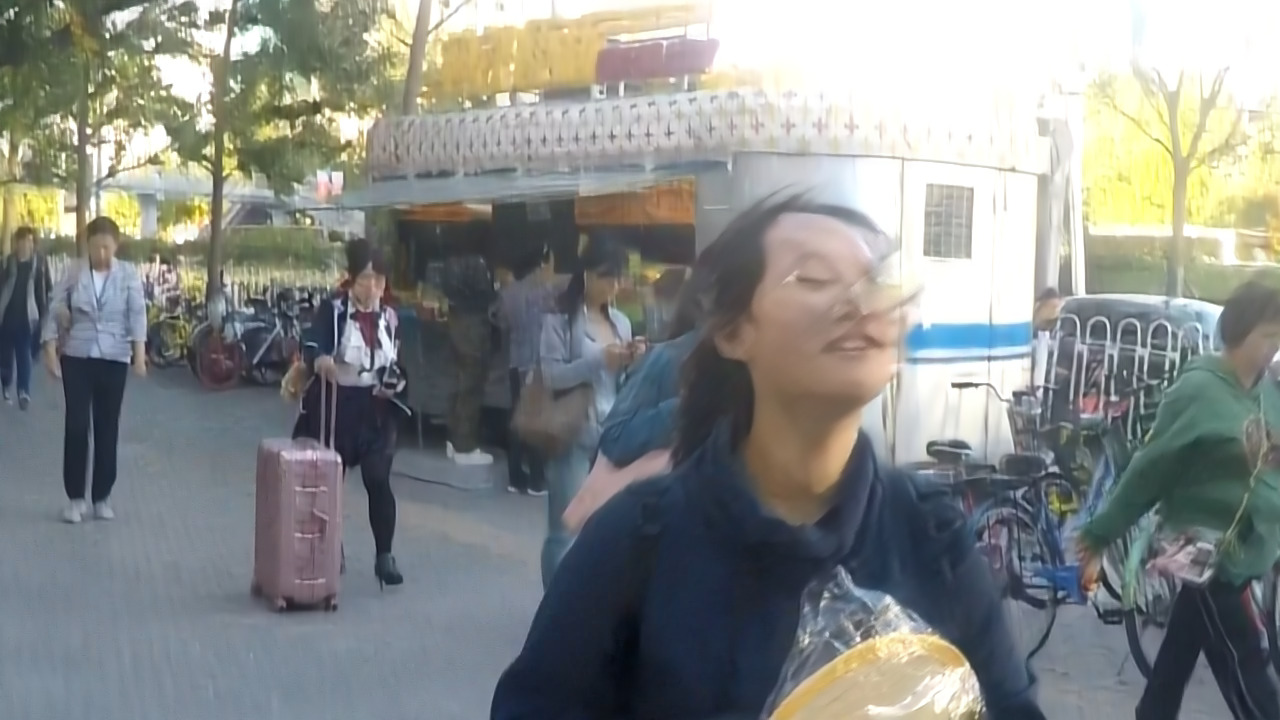}
         \caption{(a) Nah}
     \end{subfigure}
     \hfill
     \begin{subfigure}[b]{\gpw\textwidth}
         \centering
         \includegraphics[trim={500 160 200 120},clip,width=\textwidth]{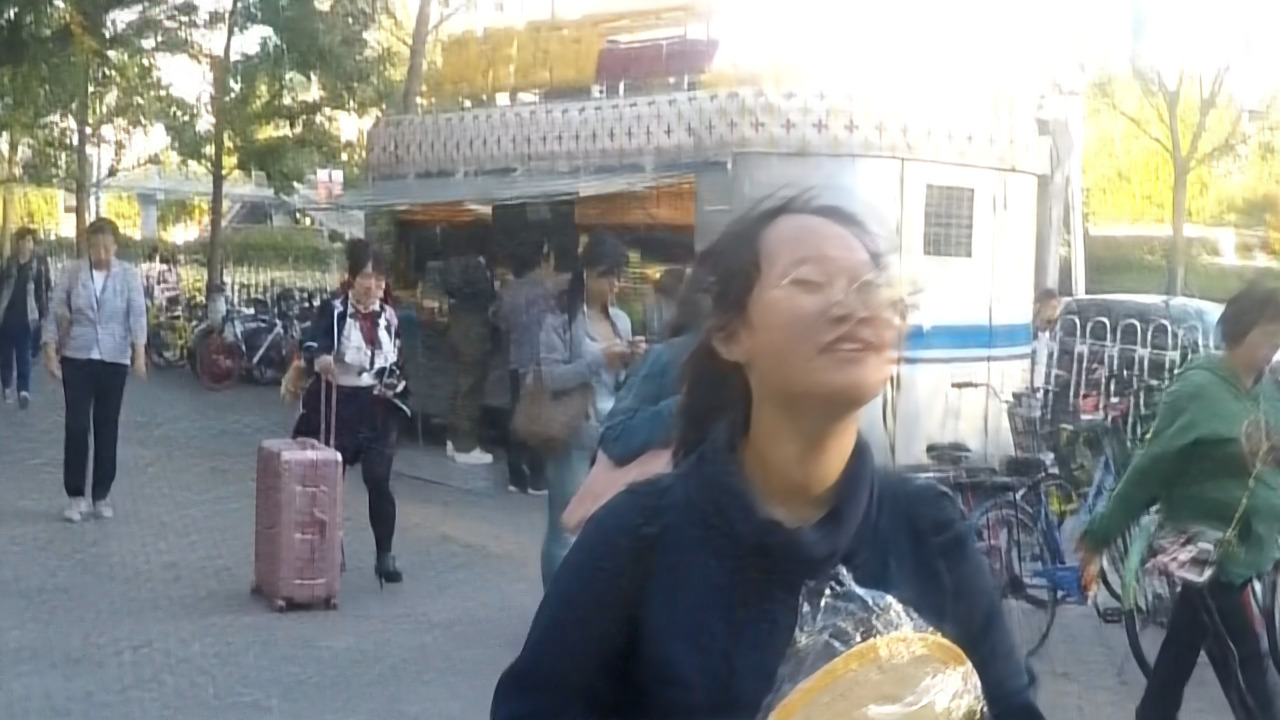}
         \caption{(a) SRN}
     \end{subfigure}
     \hfill
     \begin{subfigure}[b]{\gpw\textwidth}
         \centering
         \includegraphics[trim={500 160 200 120},clip,width=\textwidth]{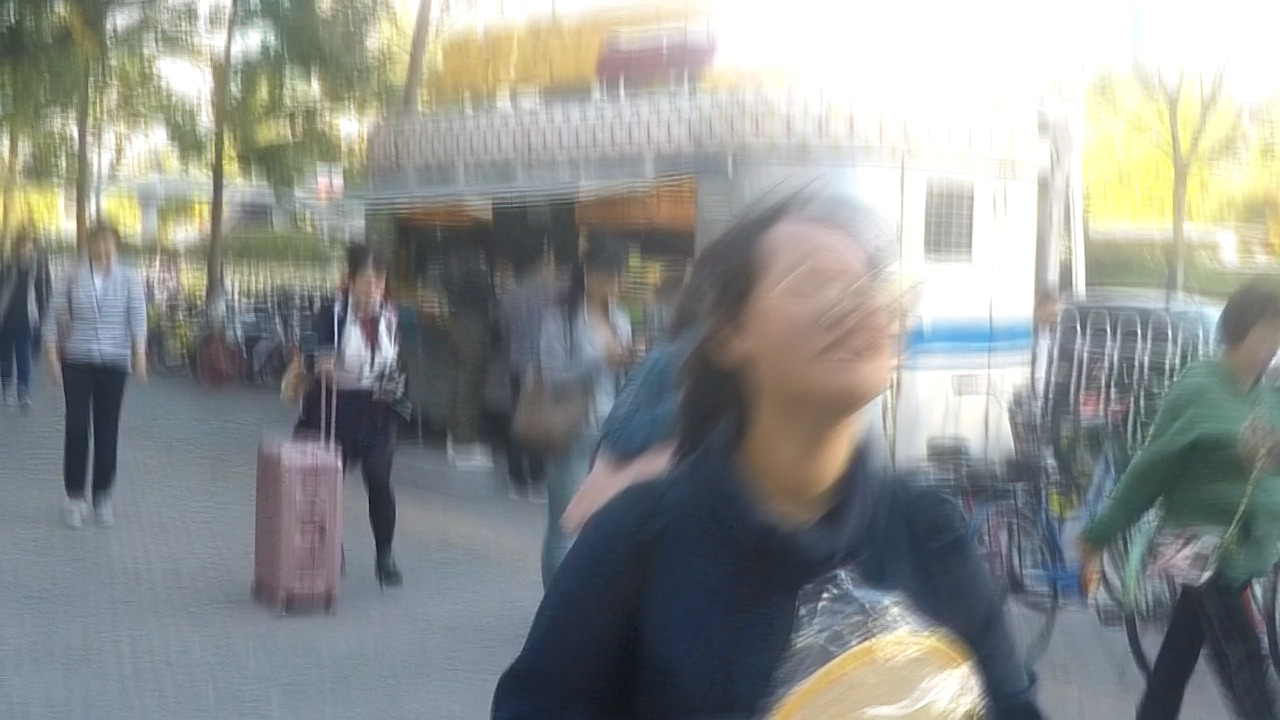}
         \caption{(a) DG1}
     \end{subfigure}
     \hfill
     \begin{subfigure}[b]{\gpw\textwidth}
         \centering
         \includegraphics[trim={500 160 200 120},clip,width=\textwidth]{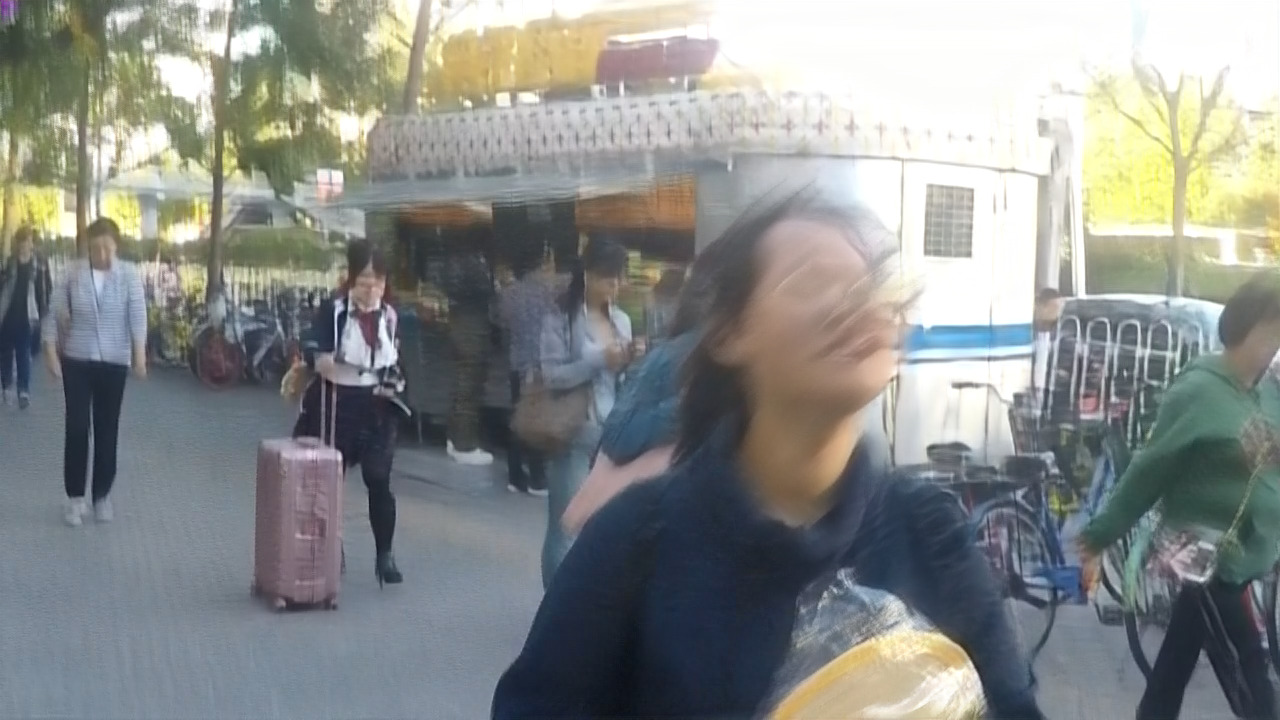}
         \caption{(a) DG2}
     \end{subfigure} \\
     \begin{subfigure}[b]{\gpw\textwidth}
         \centering
         \includegraphics[trim={500 160 200 120},clip,width=\textwidth]{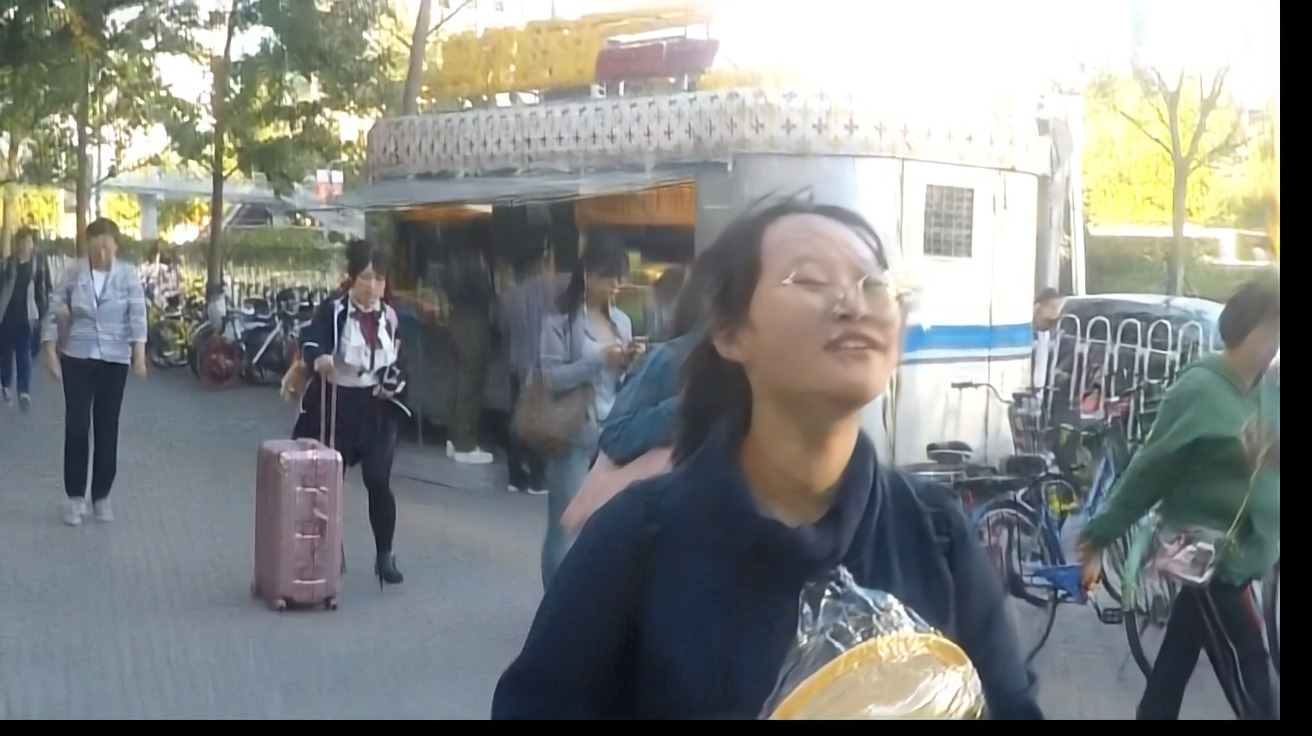}
         \caption{(a) DMPHN}
     \end{subfigure}
     \hfill 
     \begin{subfigure}[b]{\gpw\textwidth}
         \centering
         \includegraphics[trim={500 160 200 120},clip,width=\textwidth]{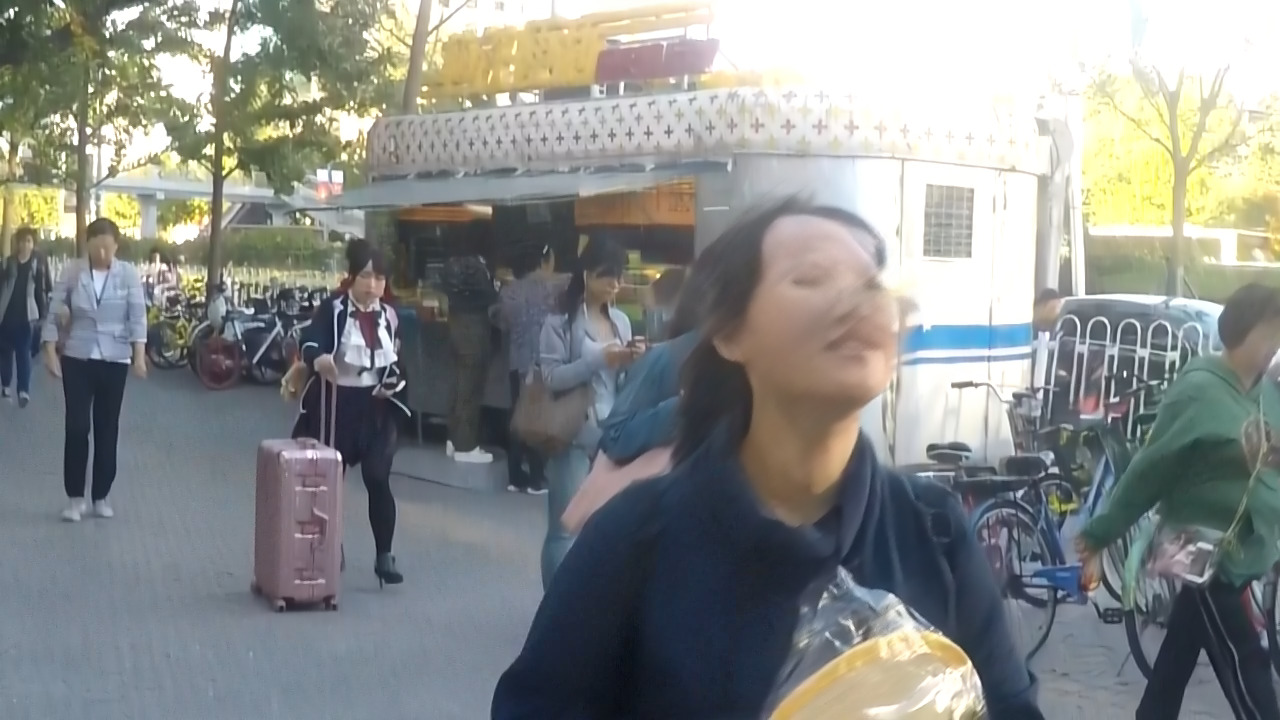}
         \caption{(a) MPR}
     \end{subfigure}
     \hfill
     \begin{subfigure}[b]{\gpw\textwidth}
         \centering
         \includegraphics[trim={500 160 200 120},clip,width=\textwidth]{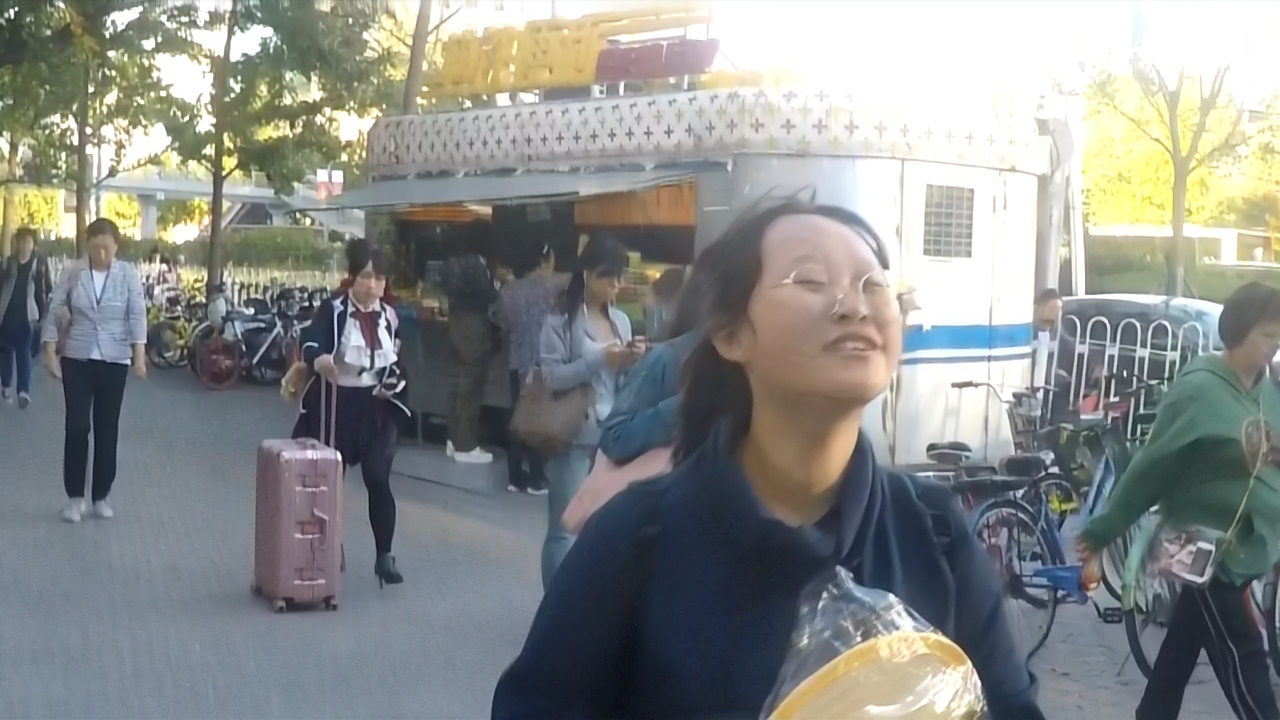}
         \caption{(a) Uform.}
     \end{subfigure}
     \hfill
     \begin{subfigure}[b]{\gpw\textwidth}
         \centering
         \includegraphics[trim={500 160 200 120},clip,width=\textwidth]{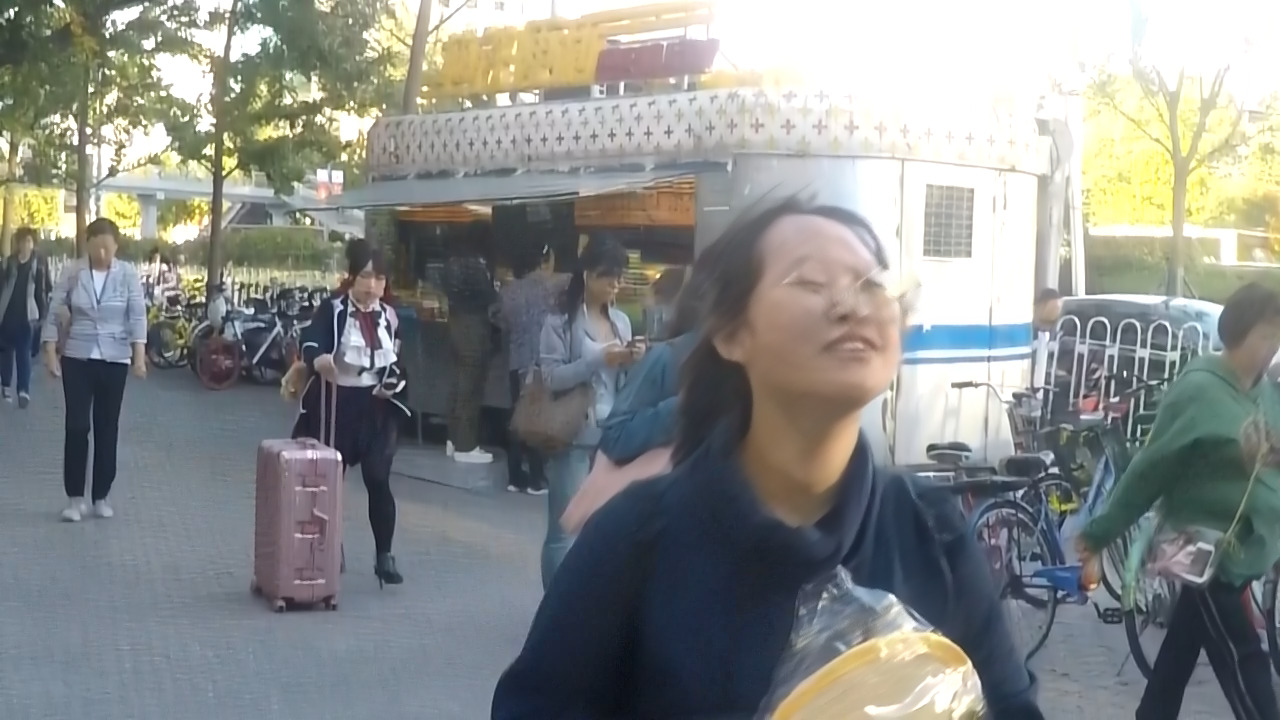}
         \caption{(a) Rest.}
     \end{subfigure}
     \hfill
     \begin{subfigure}[b]{\gpw\textwidth}
         \centering
         \includegraphics[trim={500 160 200 120},clip,width=\textwidth]{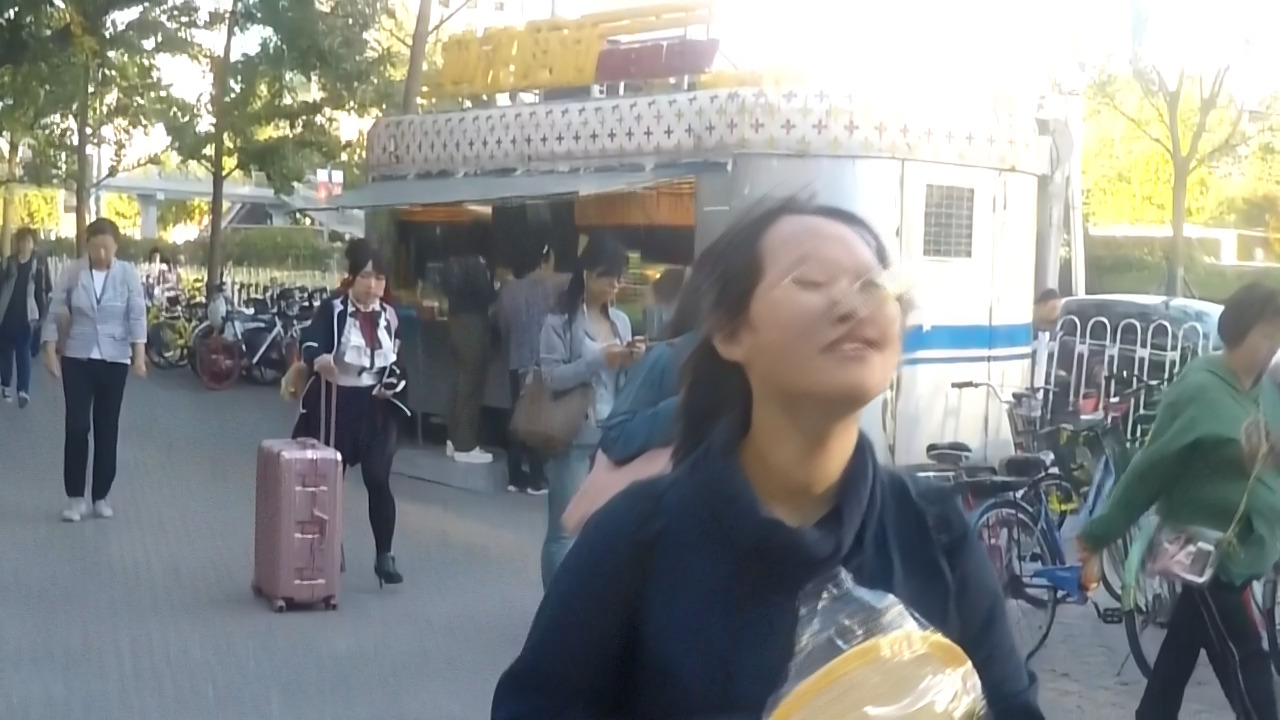}
         \caption{(a) Ours$_U$}
     \end{subfigure}
     \hfill
     \begin{subfigure}[b]{\gpw\textwidth}
         \centering
         \includegraphics[trim={500 160 200 120},clip,width=\textwidth]{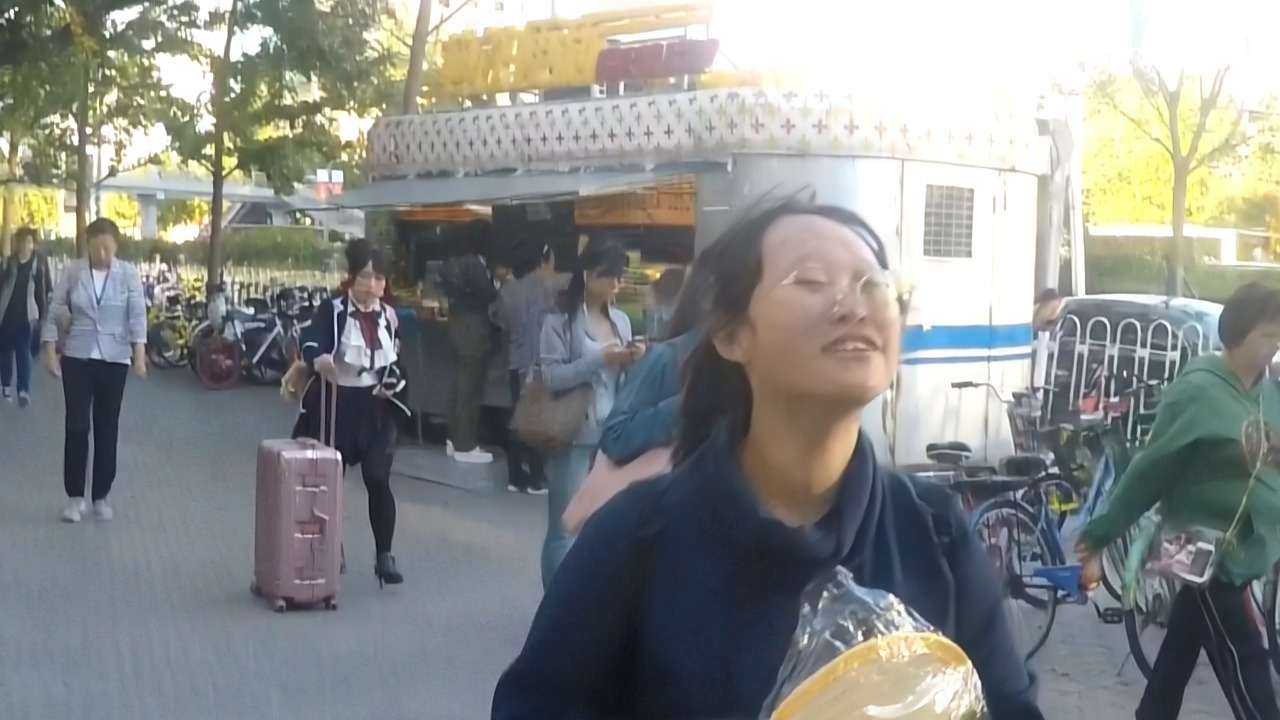}
         \caption{(a) Ours$_NU$}
     \end{subfigure}\\
     \begin{subfigure}[b]{\gpw\textwidth}
         \centering
         \includegraphics[trim={700 200 50 100},clip,width=\textwidth]{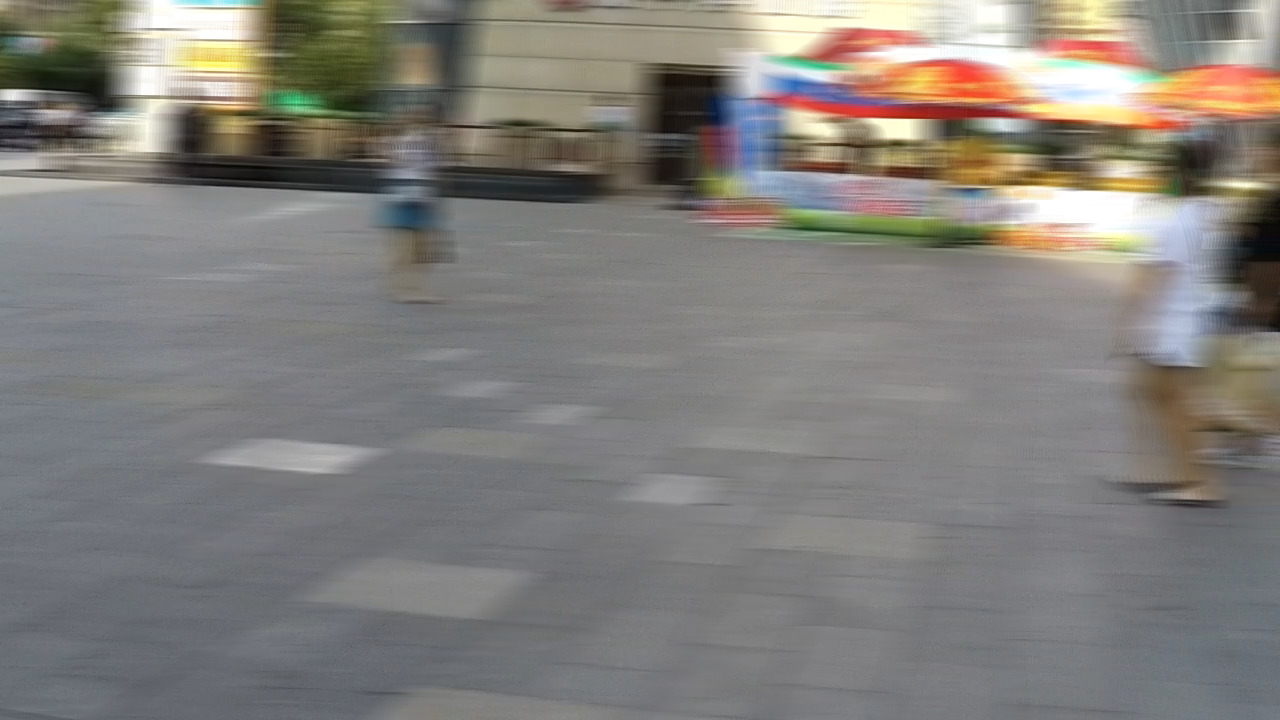}
         \caption{(b) IP}
     \end{subfigure}
     \hfill
     \begin{subfigure}[b]{\gpw\textwidth}
         \centering
         \includegraphics[trim={700 200 50 100},clip,width=\textwidth]{results/hide/set9/175fromGOPR0977_ip.jpg}
         \caption{(b) GT}
     \end{subfigure}
     \hfill
     \begin{subfigure}[b]{\gpw\textwidth}
         \centering
         \includegraphics[trim={700 200 50 100},clip,width=\textwidth]{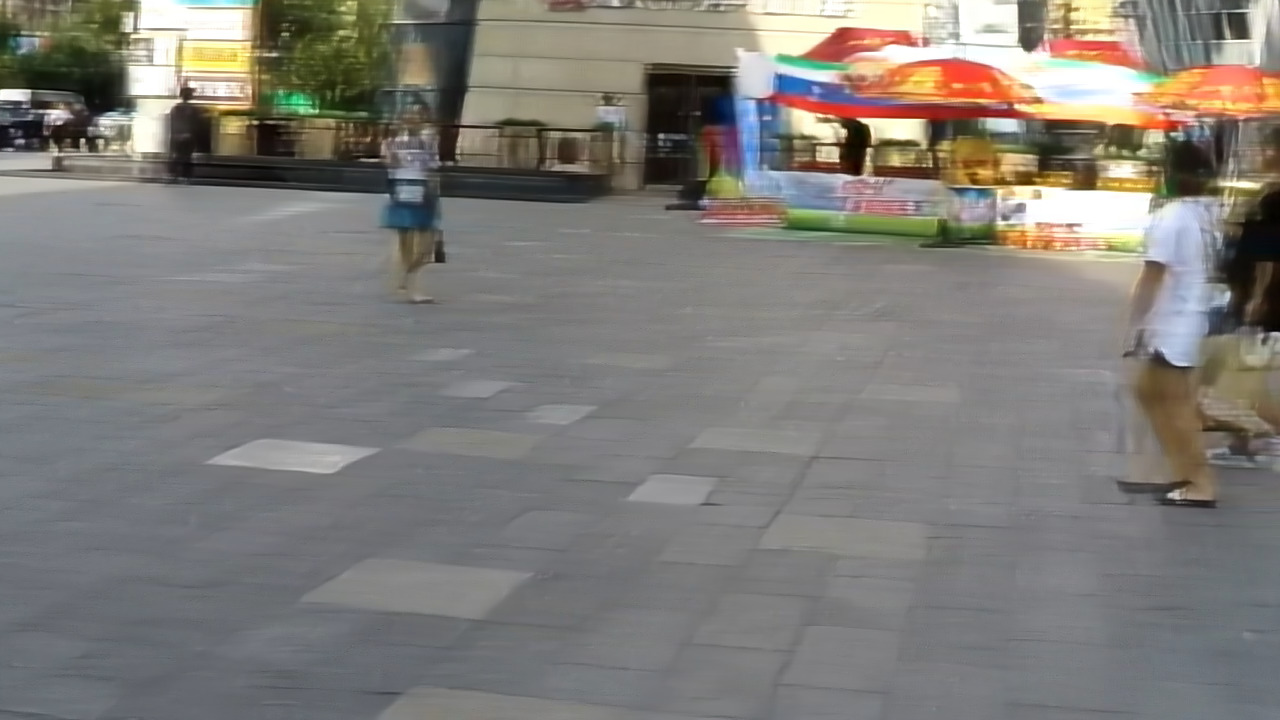}
         \caption{(b) Nah}
     \end{subfigure}
     \hfill
     \begin{subfigure}[b]{\gpw\textwidth}
         \centering
         \includegraphics[trim={700 200 50 100},clip,width=\textwidth]{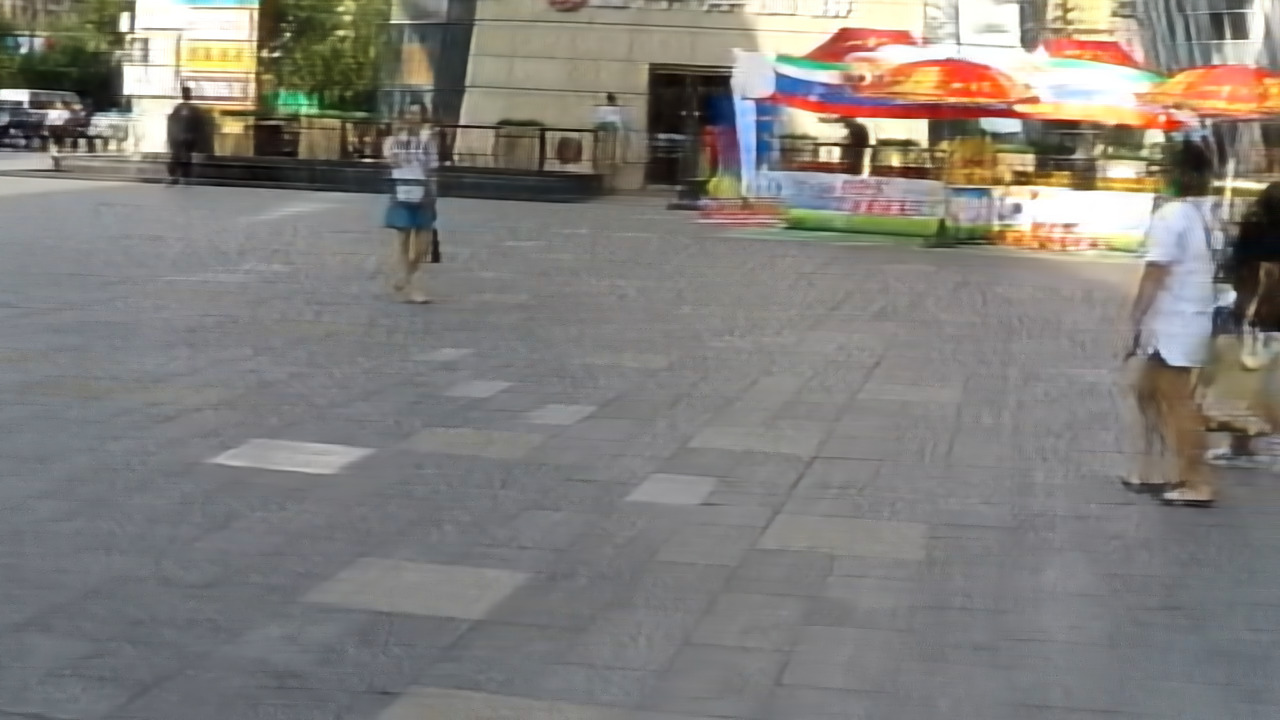}
         \caption{(b) SRN}
     \end{subfigure}
     \hfill
     \begin{subfigure}[b]{\gpw\textwidth}
         \centering
         \includegraphics[trim={700 200 50 100},clip,width=\textwidth]{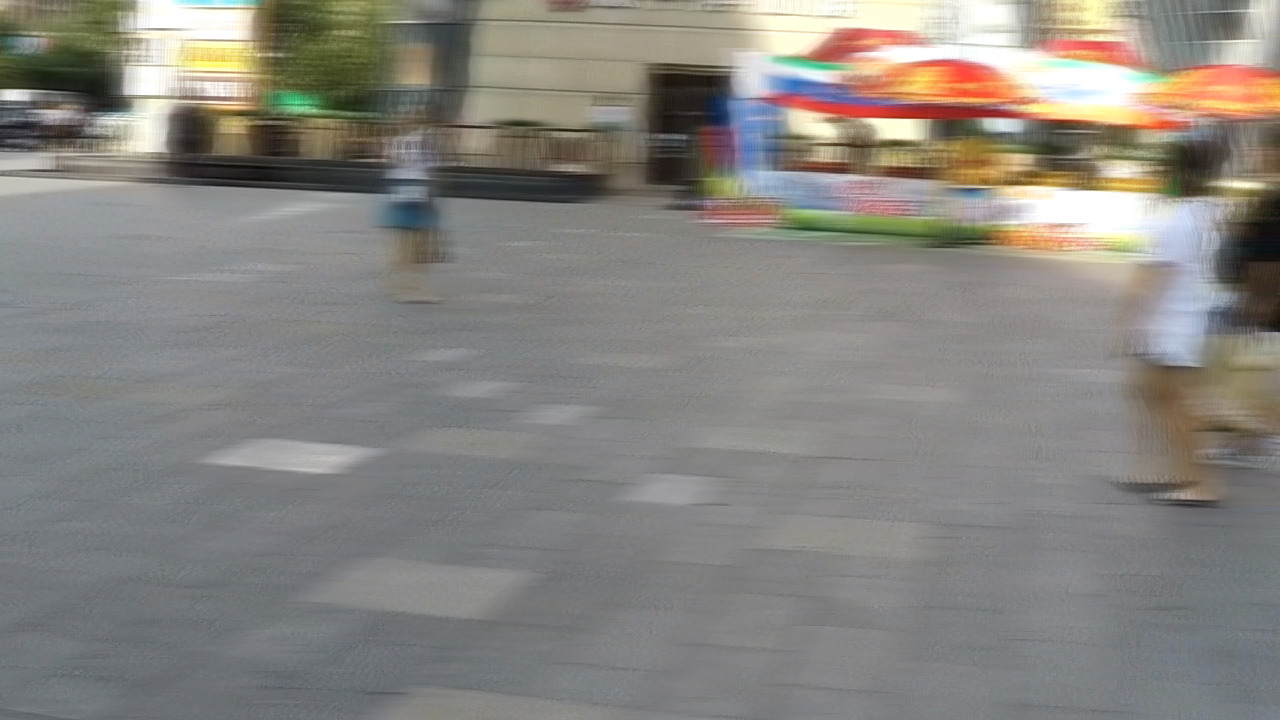}
         \caption{(b) DG1}
     \end{subfigure}
     \hfill
     \begin{subfigure}[b]{\gpw\textwidth}
         \centering
         \includegraphics[trim={700 200 50 100},clip,width=\textwidth]{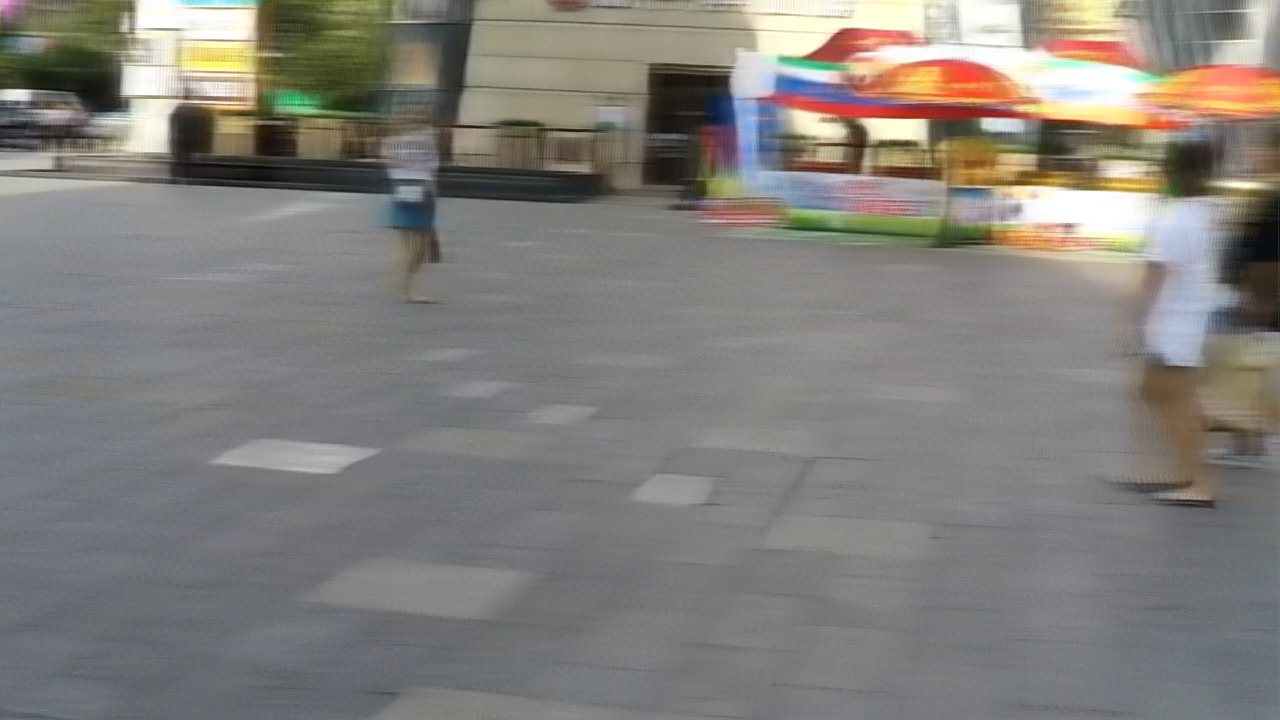}
         \caption{(b) DG2}
     \end{subfigure} \\
     \begin{subfigure}[b]{\gpw\textwidth}
         \centering
         \includegraphics[trim={700 200 50 100},clip,width=\textwidth]{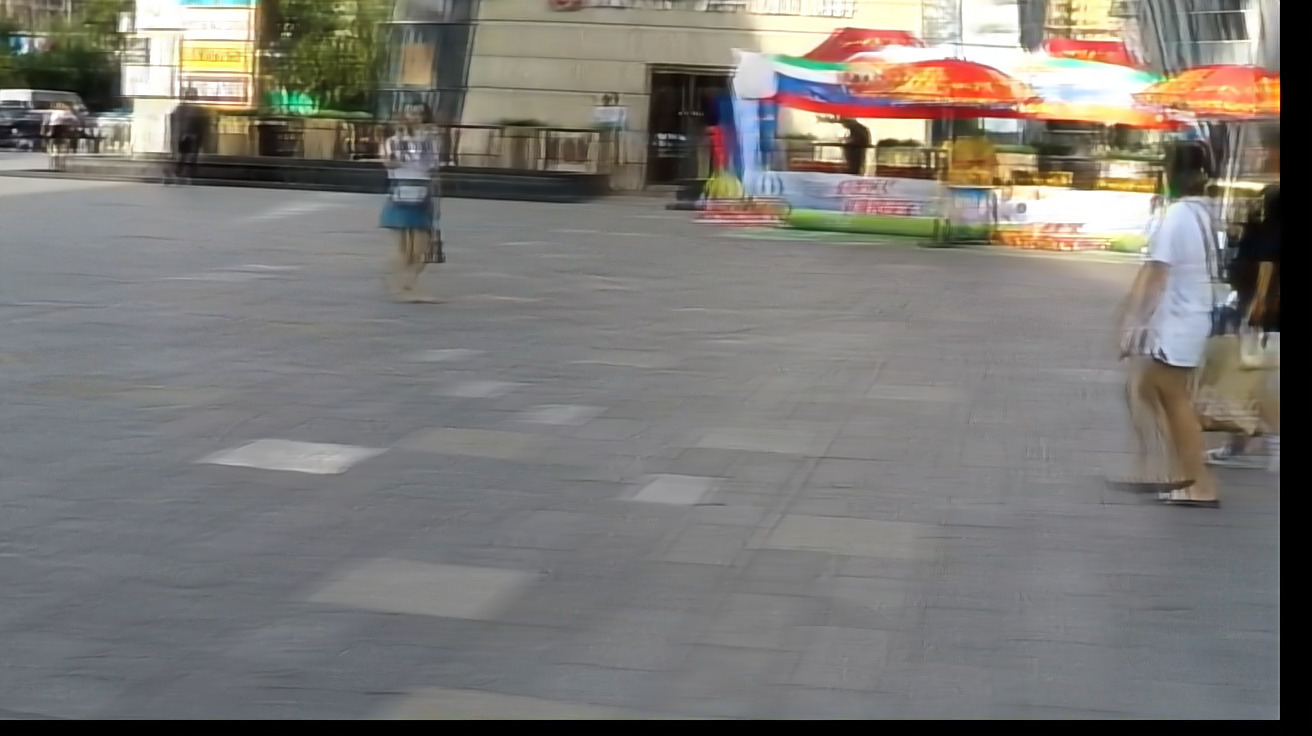}
         \caption{(b) DMPHN}
     \end{subfigure}
     \hfill 
     \begin{subfigure}[b]{\gpw\textwidth}
         \centering
         \includegraphics[trim={700 200 50 100},clip,width=\textwidth]{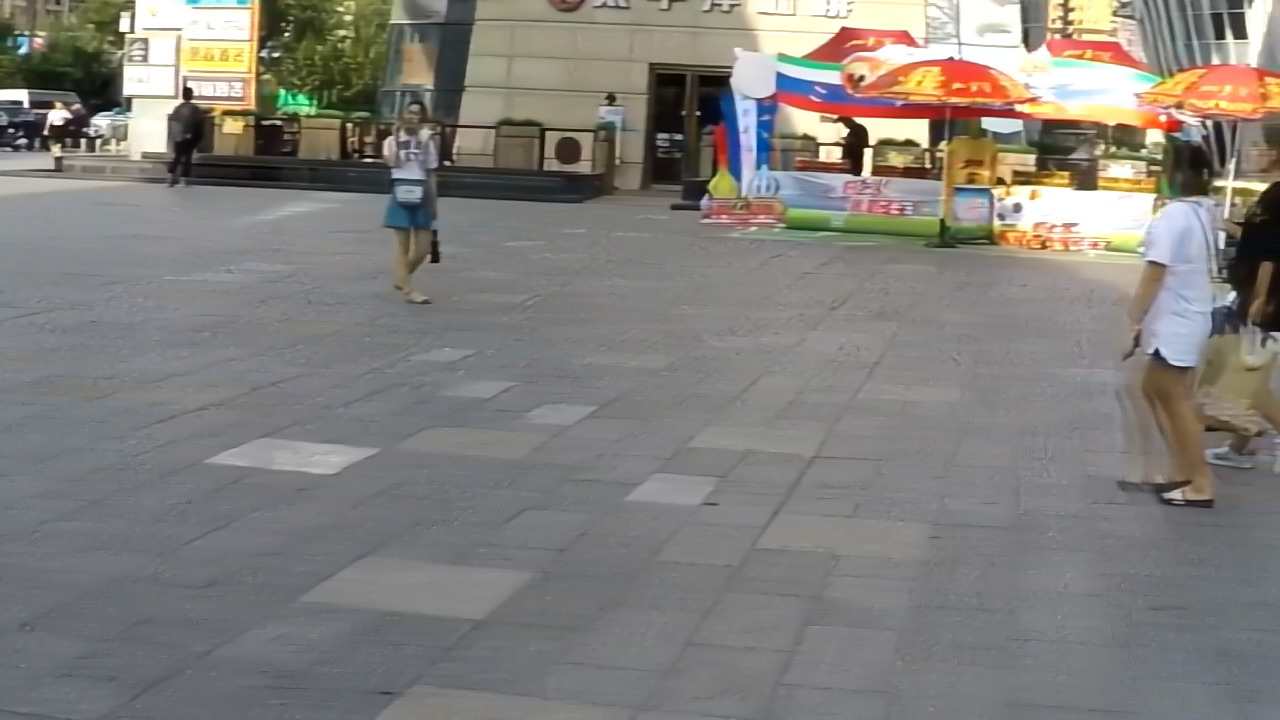}
         \caption{(b) MPR}
     \end{subfigure}
     \hfill
     \begin{subfigure}[b]{\gpw\textwidth}
         \centering
         \includegraphics[trim={700 200 50 100},clip,width=\textwidth]{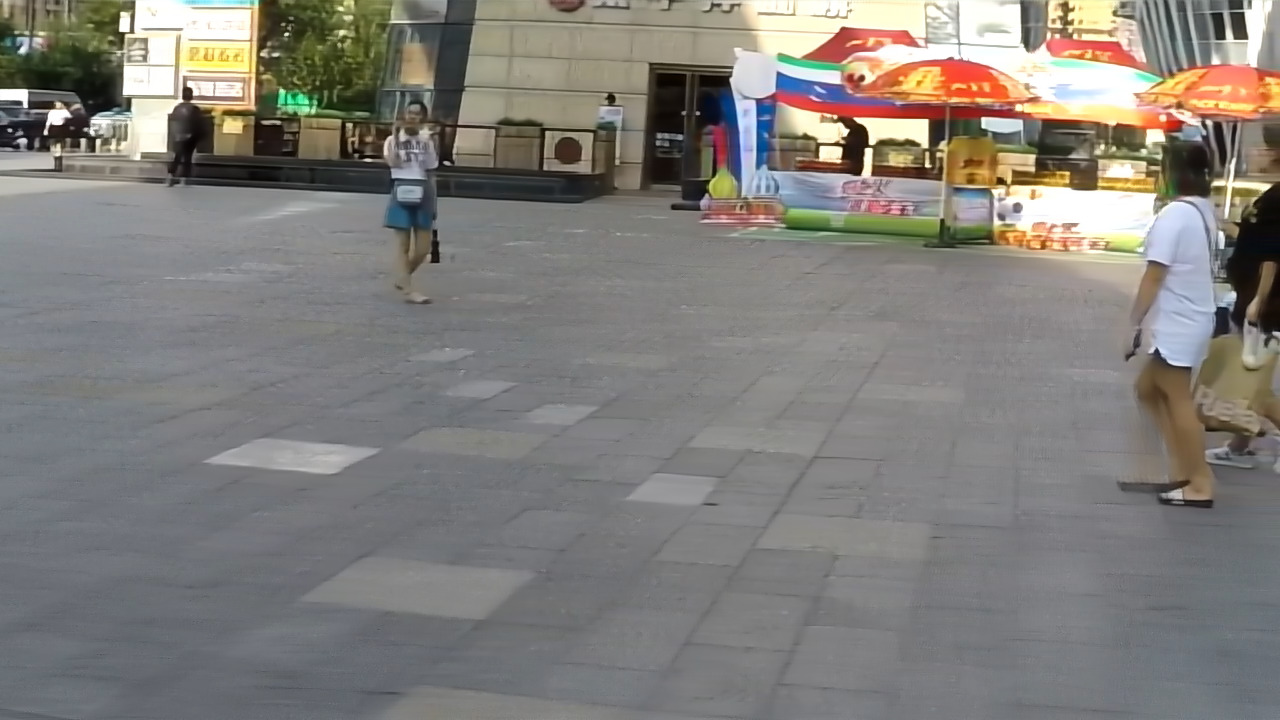}
         \caption{(b) Uform.}
     \end{subfigure}
     \hfill
     \begin{subfigure}[b]{\gpw\textwidth}
         \centering
         \includegraphics[trim={700 200 50 100},clip,width=\textwidth]{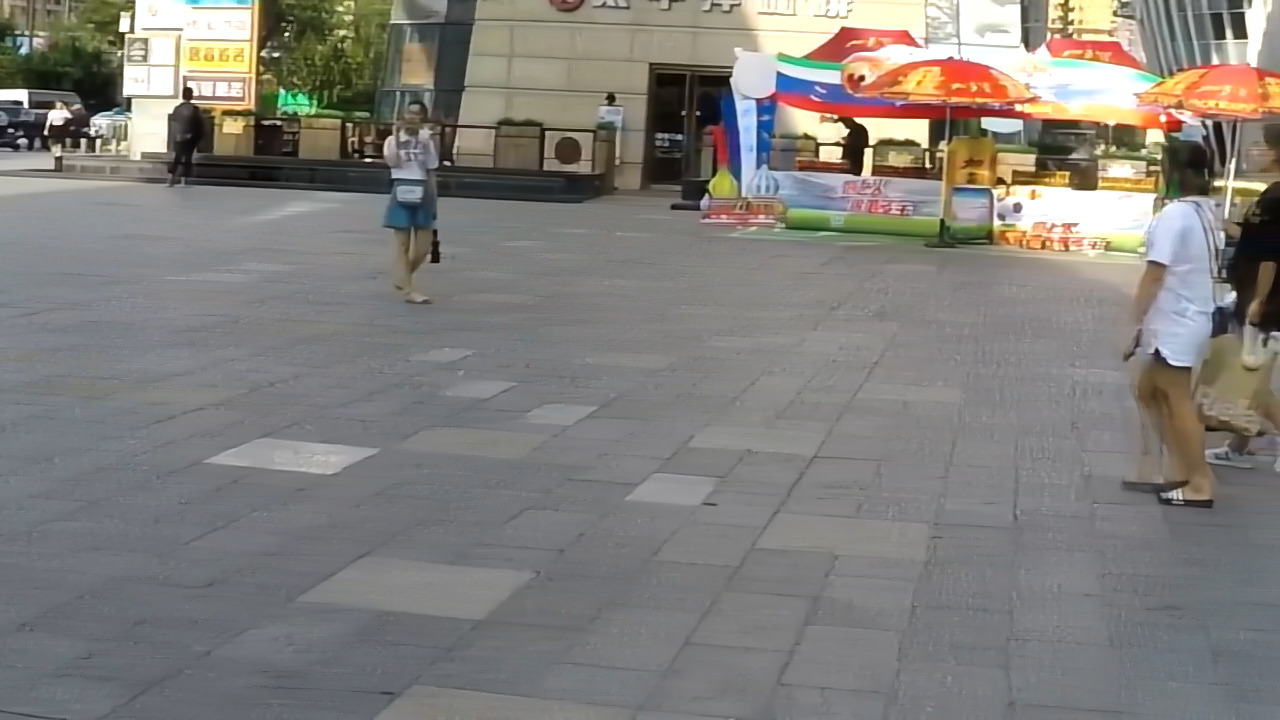}
         \caption{(b) Rest.}
     \end{subfigure}
     \hfill
     \begin{subfigure}[b]{\gpw\textwidth}
         \centering
         \includegraphics[trim={700 200 50 100},clip,width=\textwidth]{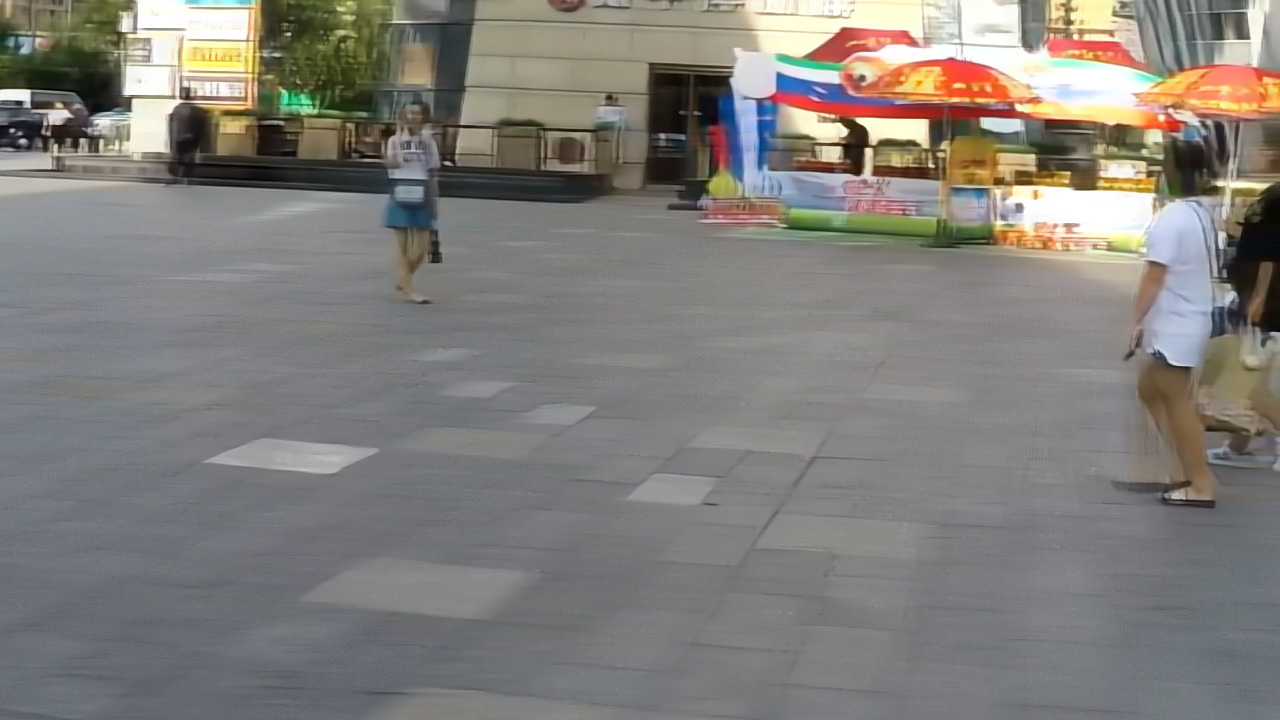}
         \caption{(b) Ours$_U$}
     \end{subfigure}
     \hfill
     \begin{subfigure}[b]{\gpw\textwidth}
         \centering
         \includegraphics[trim={700 200 50 100},clip,width=\textwidth]{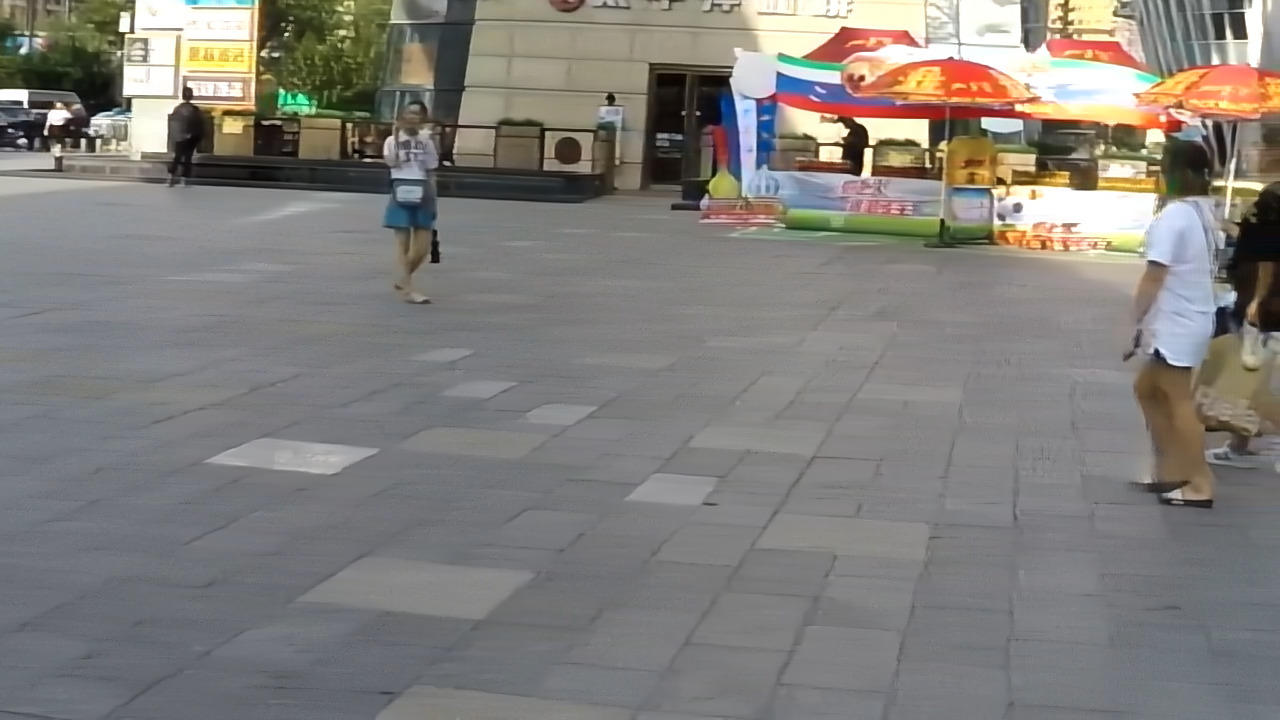}
         \caption{(b) Ours$_{NU}$}
     \end{subfigure}
        \caption{Visual comparisons of deblurring results on images from the HIDE test set.}
        \label{fig:qual_hide}
\end{figure*}

\newcommand{\gpwr}{0.105}

\begin{figure*}[htb]
\captionsetup[subfigure]{labelformat=empty}
     \centering
     \begin{subfigure}[b]{\gpwr\textwidth}
         \centering
         \includegraphics[trim={450 130 0 450},clip,width=\textwidth]{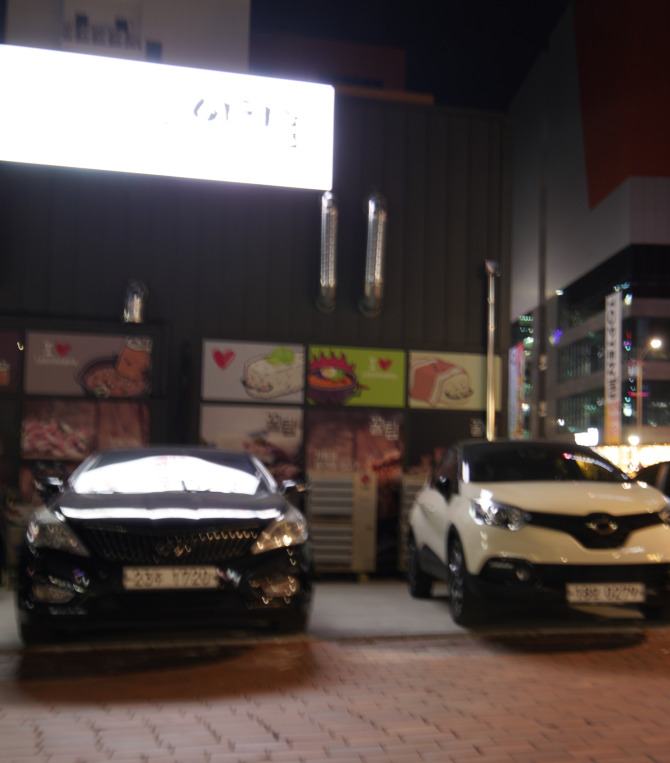}
     \end{subfigure}
     \hfill
     \begin{subfigure}[b]{\gpwr\textwidth}
         \centering
         \includegraphics[trim={450 130 0 450},clip,width=\textwidth]{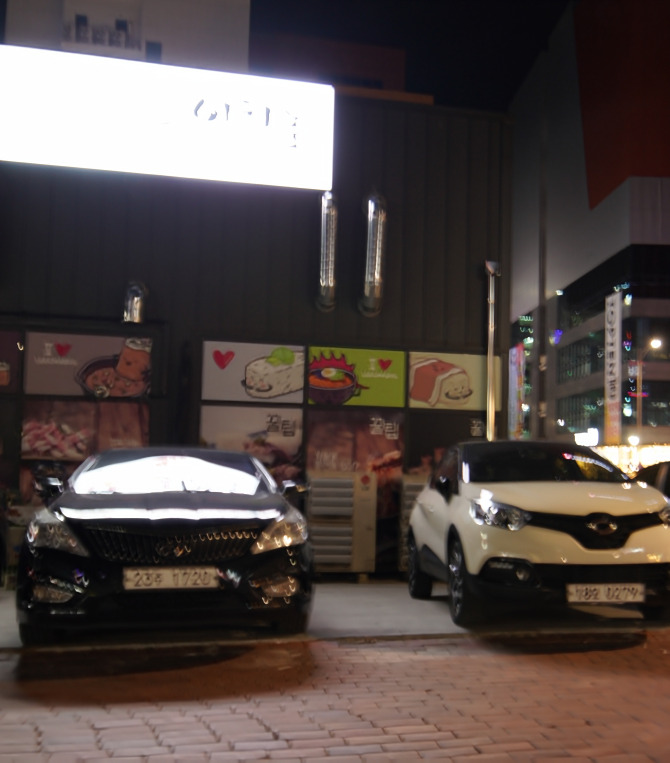}
     \end{subfigure}
     \hfill
     \begin{subfigure}[b]{\gpwr\textwidth}
         \centering
         \includegraphics[trim={450 130 0 450},clip,width=\textwidth]{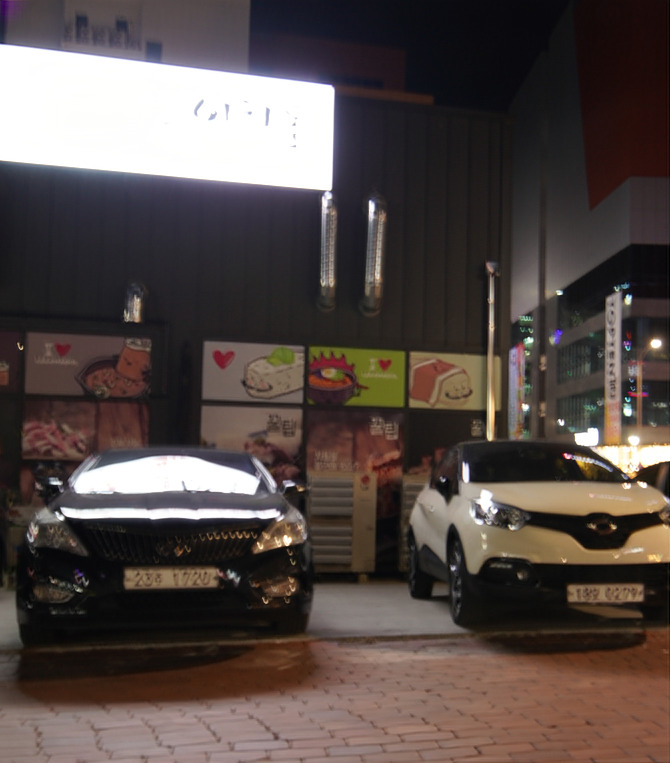}
     \end{subfigure}
     \hfill
     \begin{subfigure}[b]{\gpwr\textwidth}
         \centering
         \includegraphics[trim={450 130 0 450},clip,width=\textwidth]{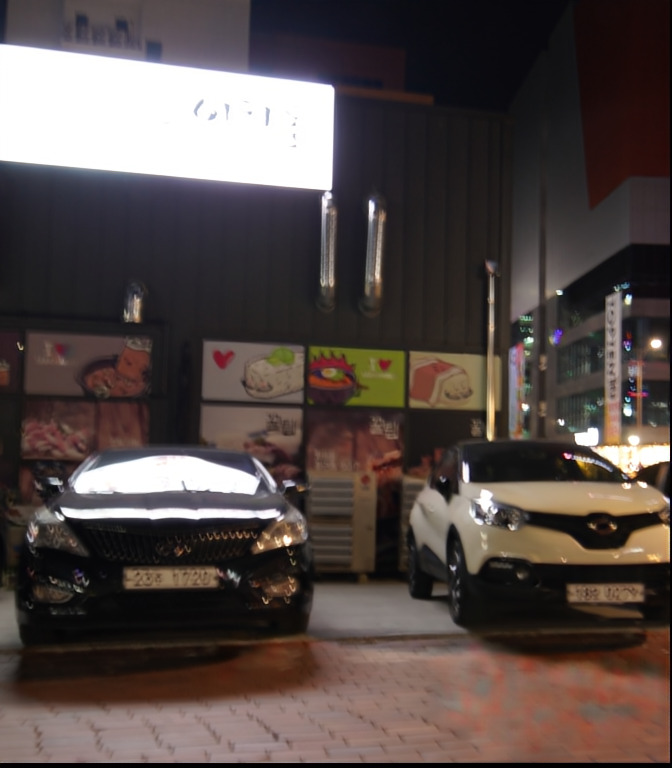}
     \end{subfigure}
     \hfill
     \begin{subfigure}[b]{\gpwr\textwidth}
         \centering
         \includegraphics[trim={450 130 0 450},clip,width=\textwidth]{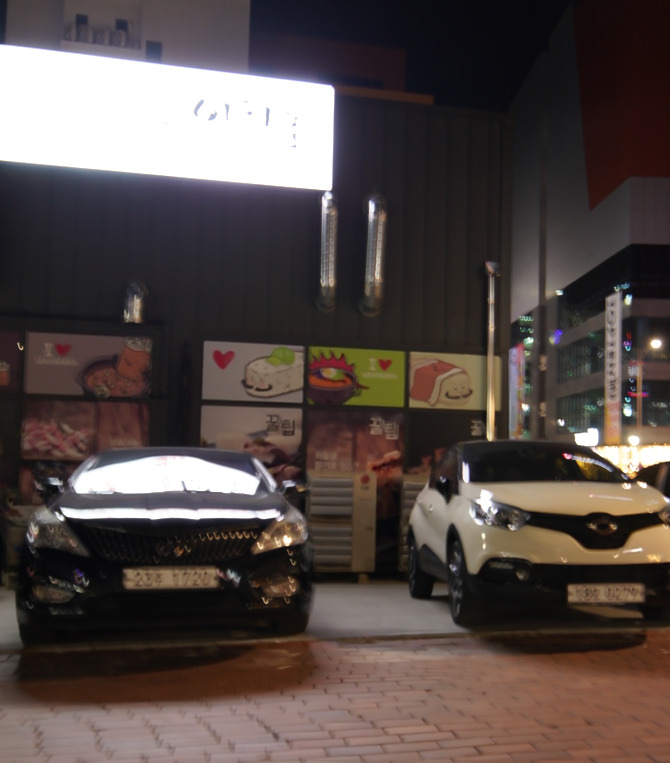}
     \end{subfigure}
     \hfill
     \begin{subfigure}[b]{\gpwr\textwidth}
         \centering
         \includegraphics[trim={450 130 0 450},clip,width=\textwidth]{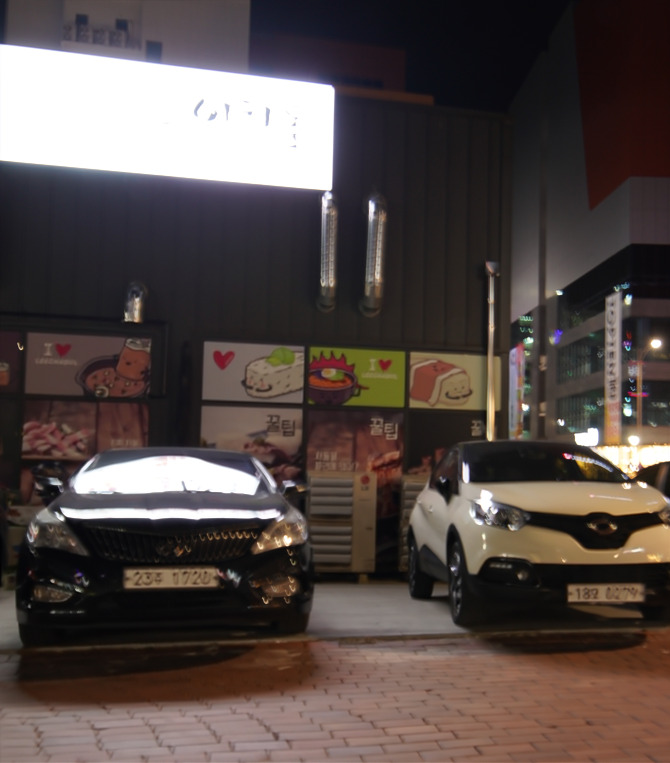}
     \end{subfigure}
     \hfill
     \begin{subfigure}[b]{\gpwr\textwidth}
         \centering
         \includegraphics[trim={450 130 0 450},clip,width=\textwidth]{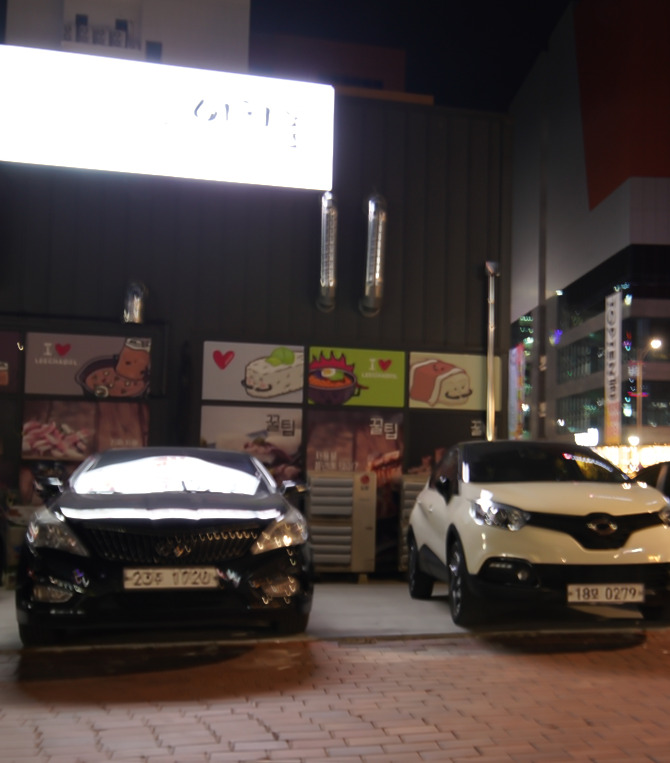}
     \end{subfigure}
     \hfill
     \begin{subfigure}[b]{\gpwr\textwidth}
         \centering
         \includegraphics[trim={450 130 0 450},clip,width=\textwidth]{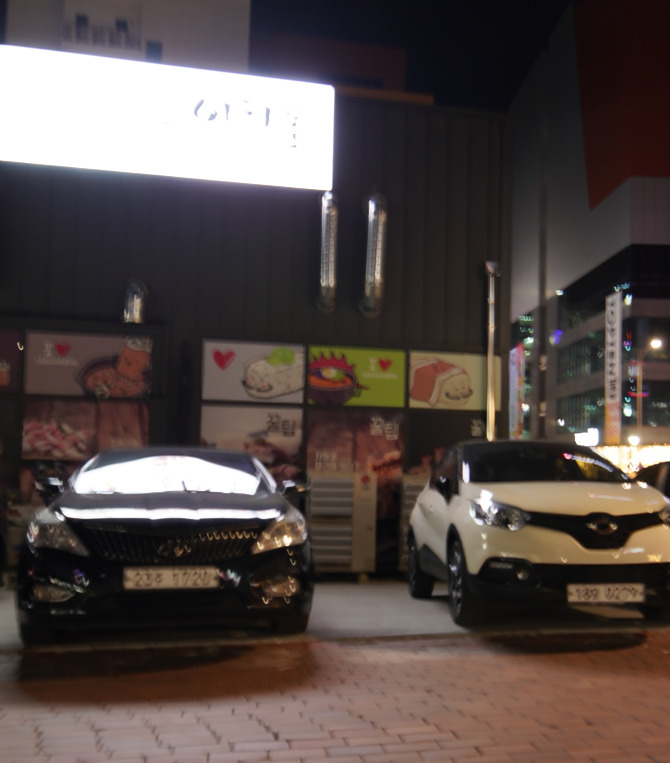}
     \end{subfigure}
     \hfill
     \begin{subfigure}[b]{\gpwr\textwidth}
         \centering
         \includegraphics[trim={450 130 0 450},clip,width=\textwidth]{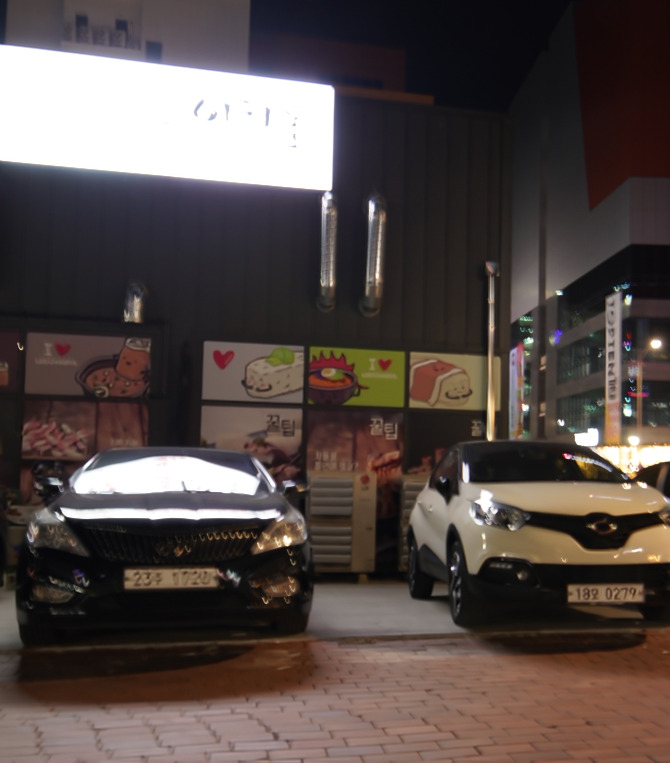}
     \end{subfigure} \hfill \\
     \begin{subfigure}[b]{\gpwr\textwidth}
         \centering
         \includegraphics[trim={200 250 200 250},clip,width=\textwidth]{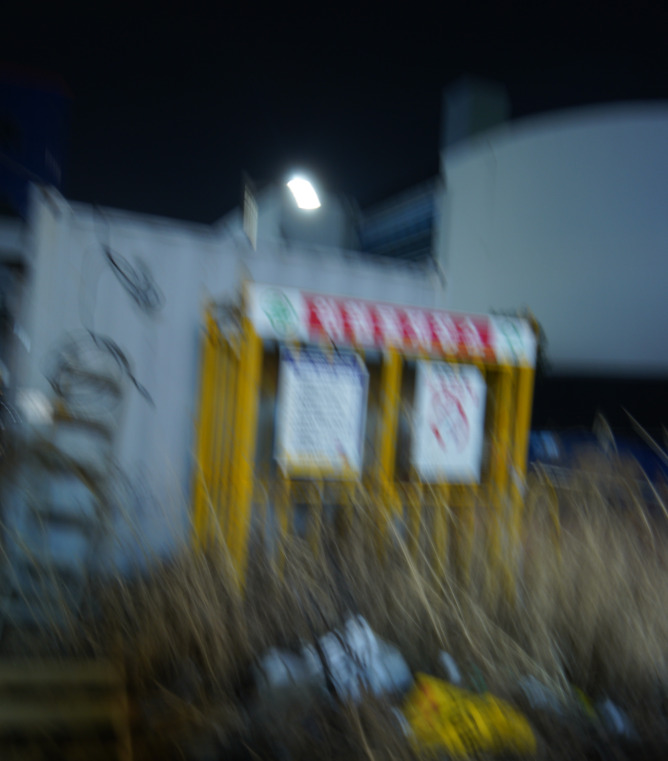}
     \end{subfigure}
     \hfill
     \begin{subfigure}[b]{\gpwr\textwidth}
         \centering
         \includegraphics[trim={200 250 200 250},clip,width=\textwidth]{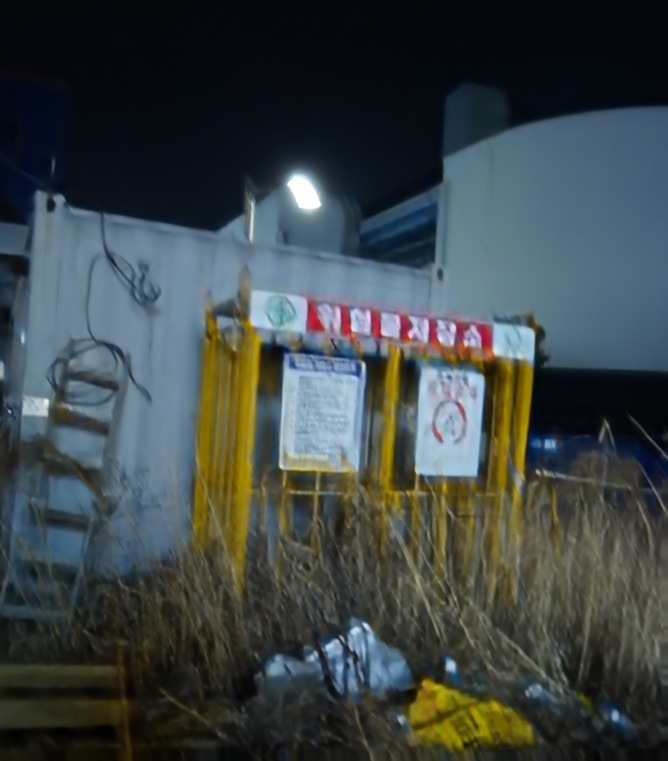}
     \end{subfigure}
     \hfill
     \begin{subfigure}[b]{\gpwr\textwidth}
         \centering
         \includegraphics[trim={200 250 200 250},clip,width=\textwidth]{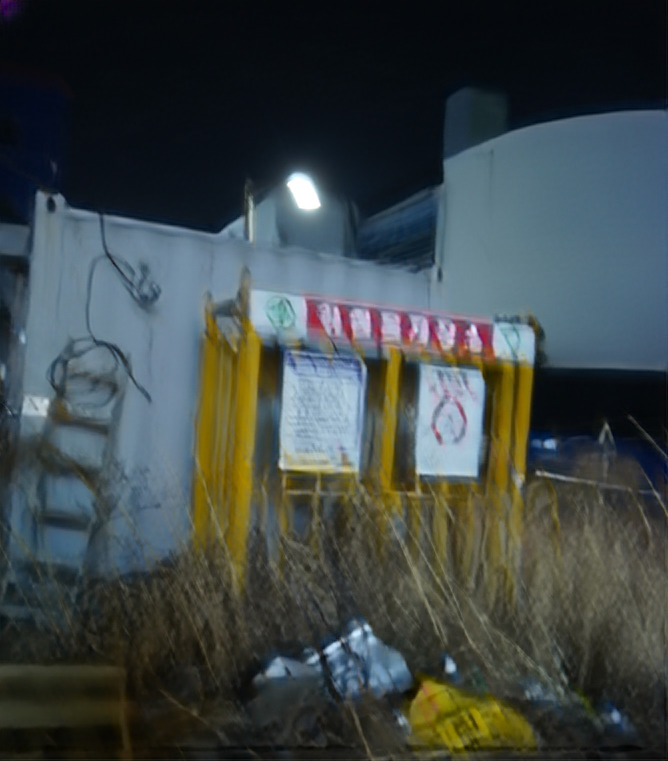}
     \end{subfigure}
     \hfill
     \begin{subfigure}[b]{\gpwr\textwidth}
         \centering
         \includegraphics[trim={200 250 200 250},clip,width=\textwidth]{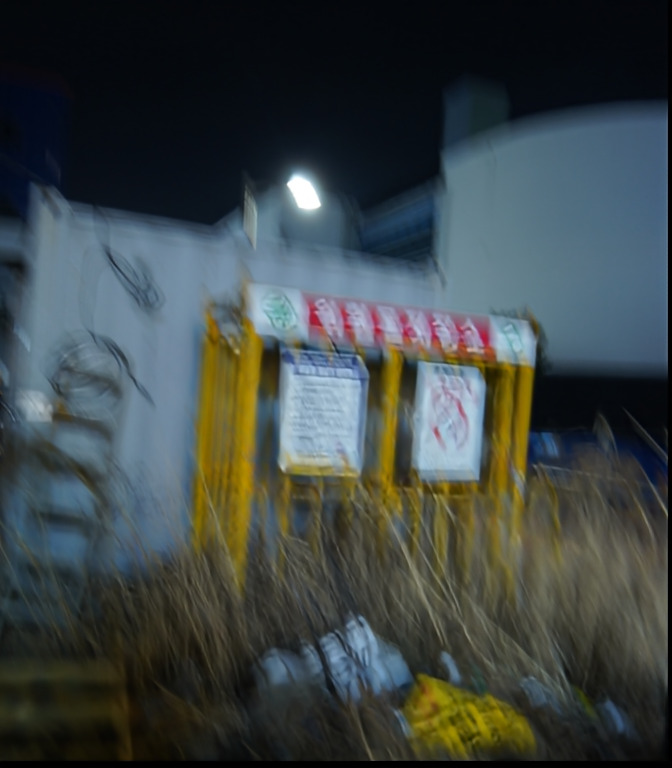}
     \end{subfigure}
     \hfill
     \begin{subfigure}[b]{\gpwr\textwidth}
         \centering
         \includegraphics[trim={200 250 200 250},clip,width=\textwidth]{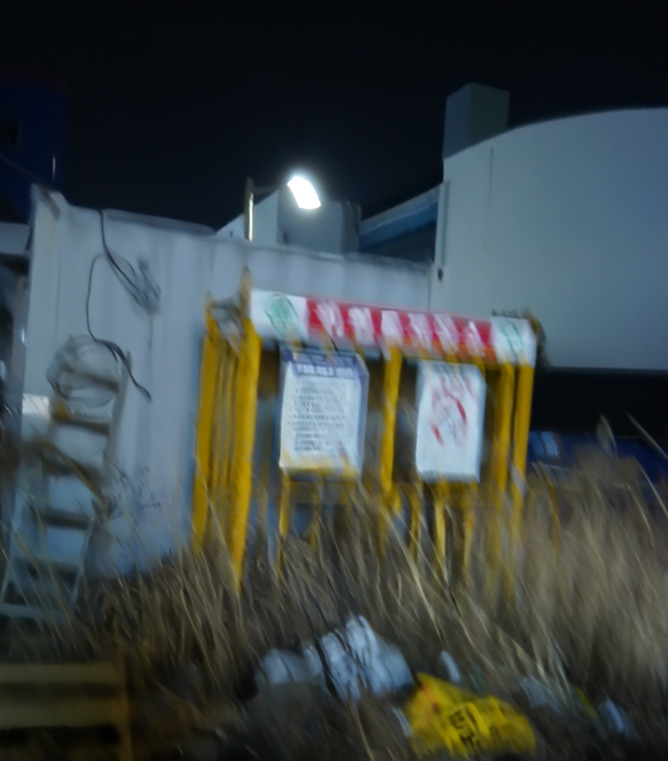}
     \end{subfigure}
     \hfill
     \begin{subfigure}[b]{\gpwr\textwidth}
         \centering
         \includegraphics[trim={200 250 200 250},clip,width=\textwidth]{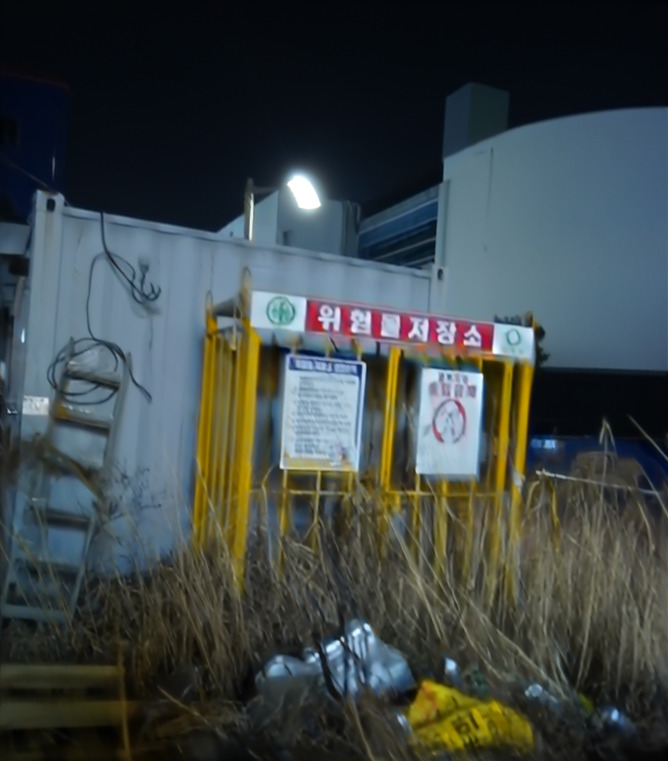}
     \end{subfigure}
     \hfill
     \begin{subfigure}[b]{\gpwr\textwidth}
         \centering
         \includegraphics[trim={200 250 200 250},clip,width=\textwidth]{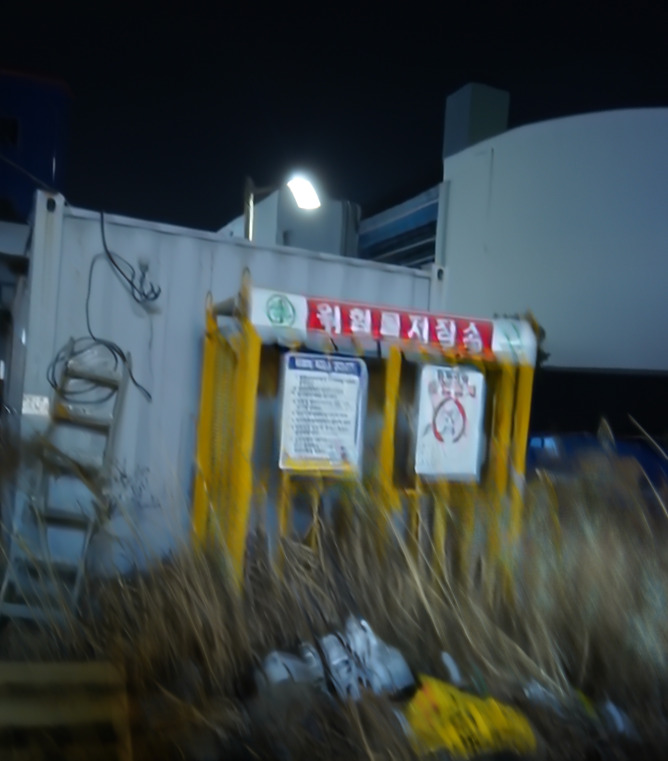}
     \end{subfigure}
     \hfill
     \begin{subfigure}[b]{\gpwr\textwidth}
         \centering
         \includegraphics[trim={200 250 200 250},clip,width=\textwidth]{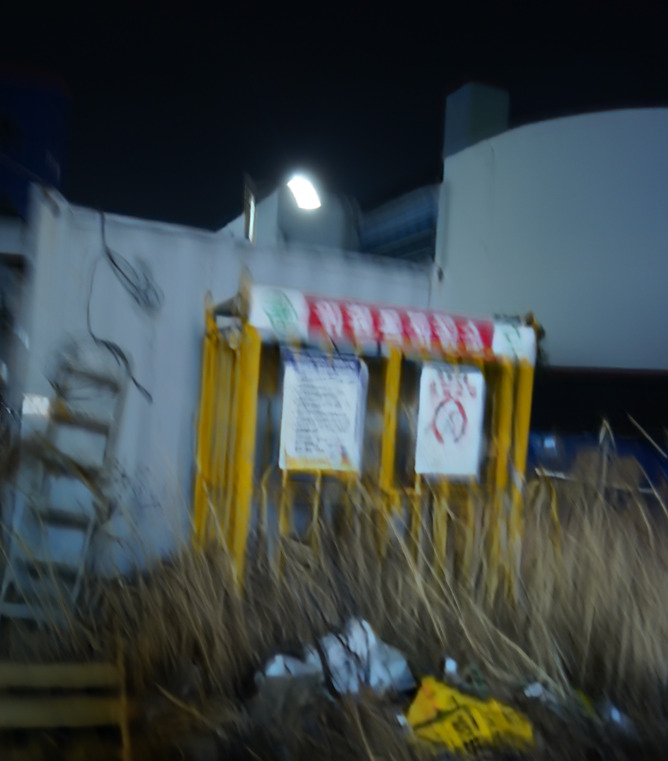}
     \end{subfigure}
     \hfill
     \begin{subfigure}[b]{\gpwr\textwidth}
         \centering
         \includegraphics[trim={200 250 200 250},clip,width=\textwidth]{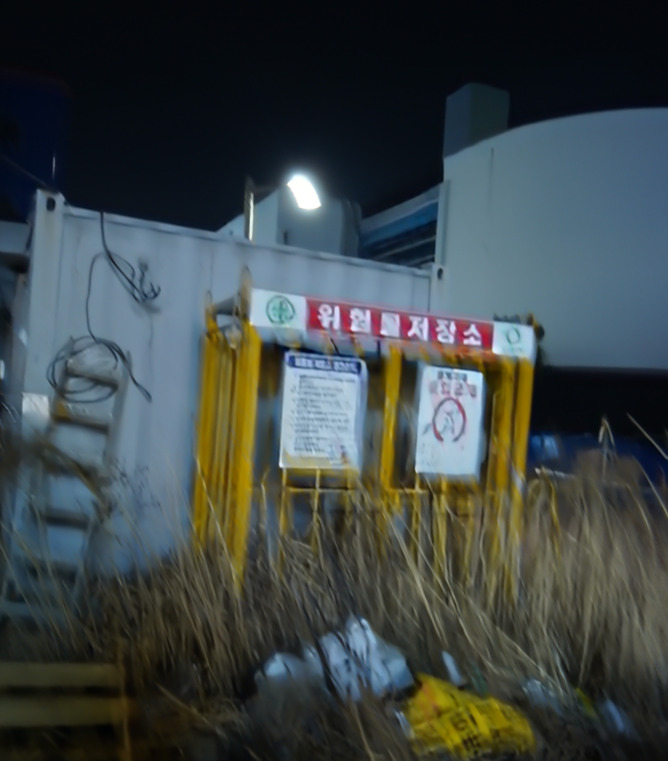}
     \end{subfigure} \\
     \begin{subfigure}[b]{\gpwr\textwidth}
         \centering
         \includegraphics[trim={0 400 300 0},clip,width=\textwidth]{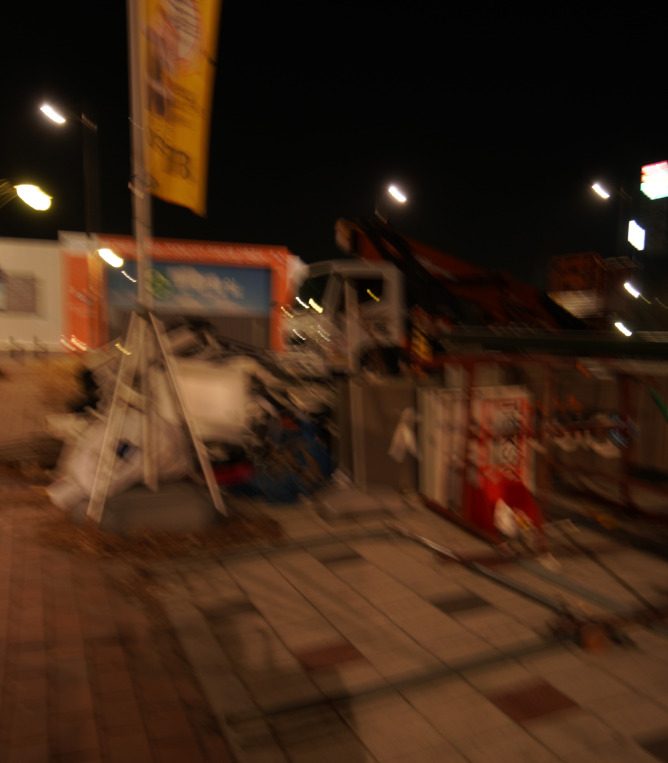}
         \caption{IP}
     \end{subfigure}
     \hfill
     \begin{subfigure}[b]{\gpwr\textwidth}
         \centering
         \includegraphics[trim={0 400 300 0},clip,width=\textwidth]{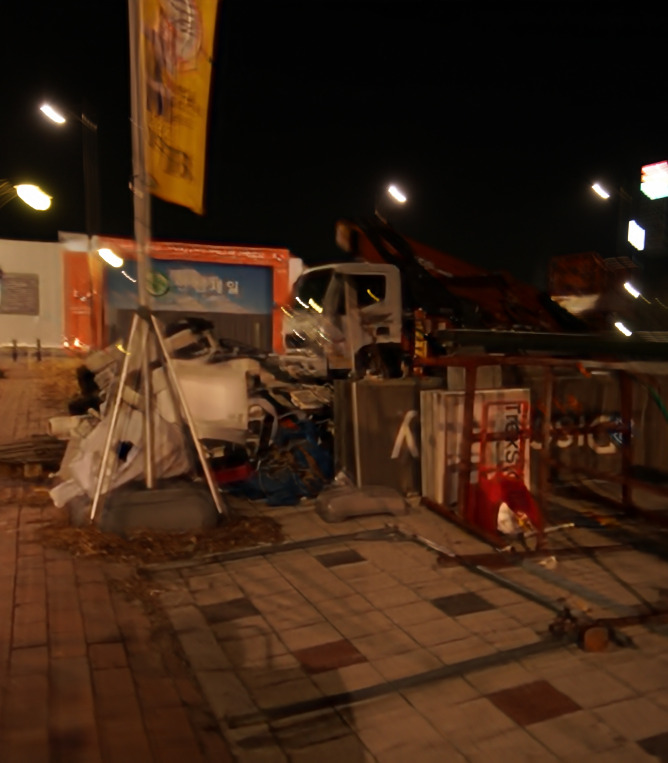}
         \caption{SRN}
     \end{subfigure}
     \hfill
     \begin{subfigure}[b]{\gpwr\textwidth}
         \centering
         \includegraphics[trim={0 400 300 0},clip,width=\textwidth]{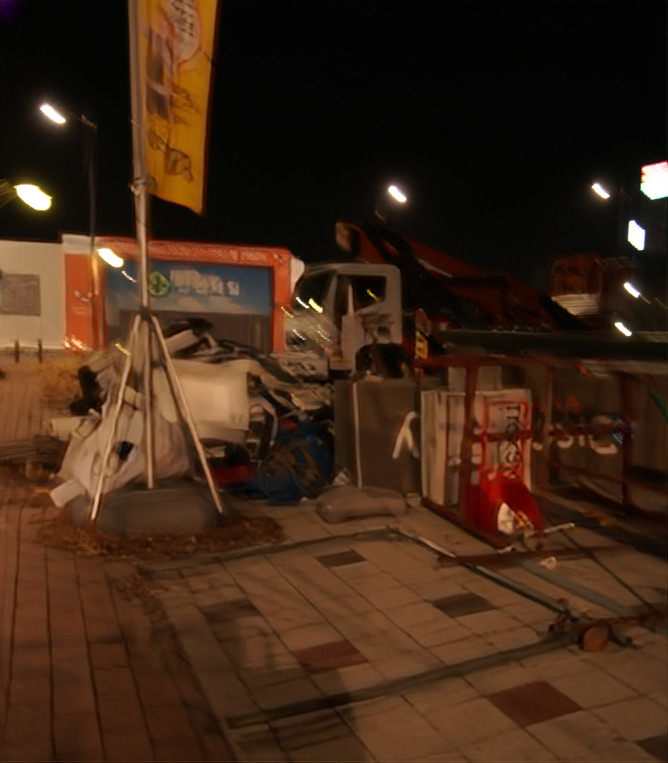}
         \caption{DG2}
     \end{subfigure}
     \hfill
     \begin{subfigure}[b]{\gpwr\textwidth}
         \centering
         \includegraphics[trim={0 400 300 0},clip,width=\textwidth]{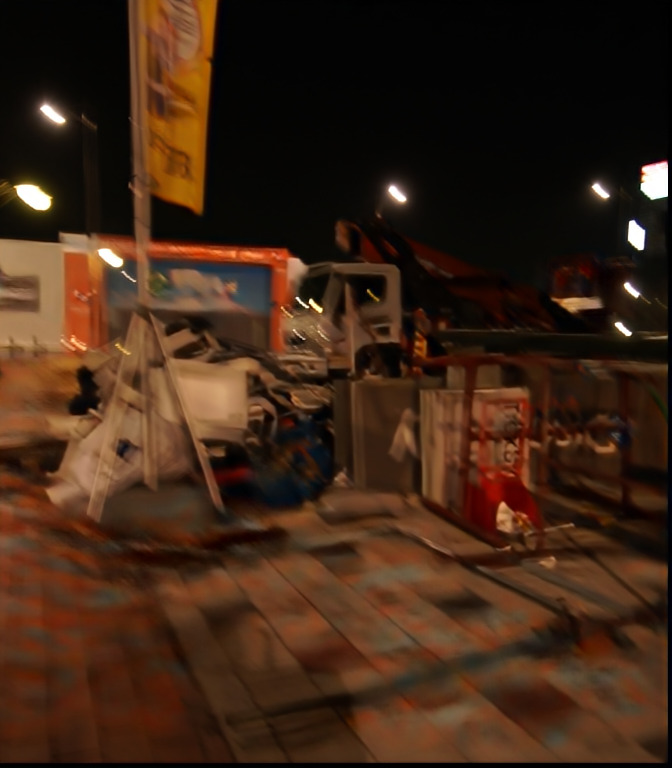}
         \caption{DMPHN}
     \end{subfigure}
     \hfill
     \begin{subfigure}[b]{\gpwr\textwidth}
         \centering
         \includegraphics[trim={0 400 300 0},clip,width=\textwidth]{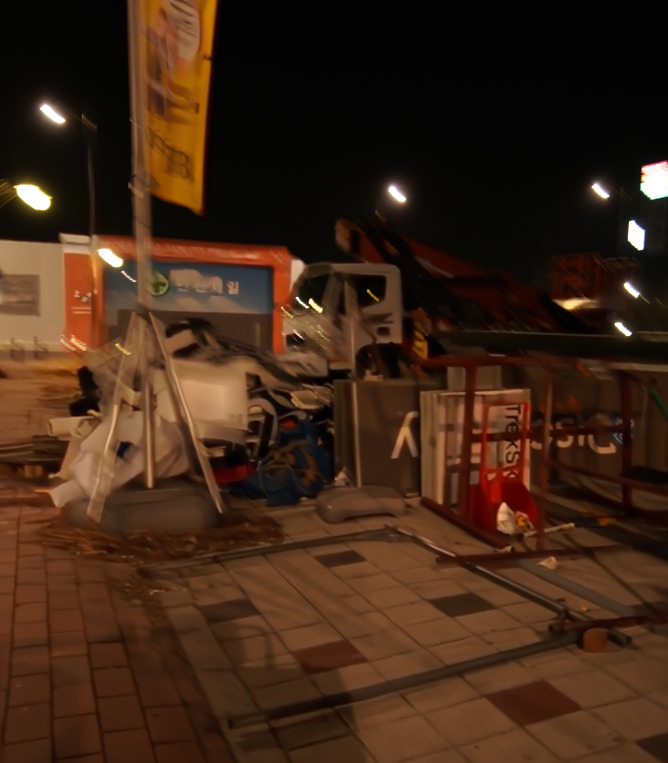}
         \caption{MPR}
     \end{subfigure}
     \hfill 
     \begin{subfigure}[b]{\gpwr\textwidth}
         \centering
         \includegraphics[trim={0 400 300 0},clip,width=\textwidth]{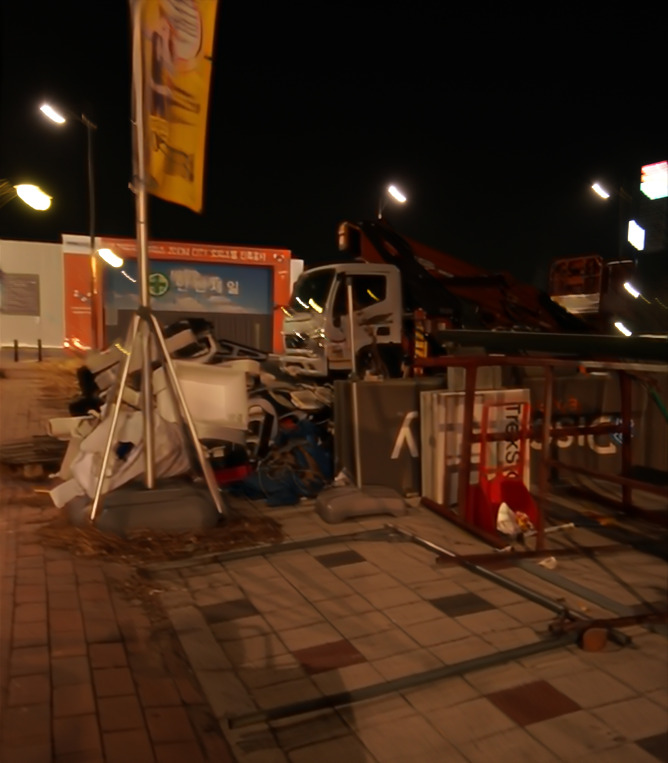}
         \caption{Uform.}
     \end{subfigure}
     \hfill 
     \begin{subfigure}[b]{\gpwr\textwidth}
         \centering
         \includegraphics[trim={0 400 300 0},clip,width=\textwidth]{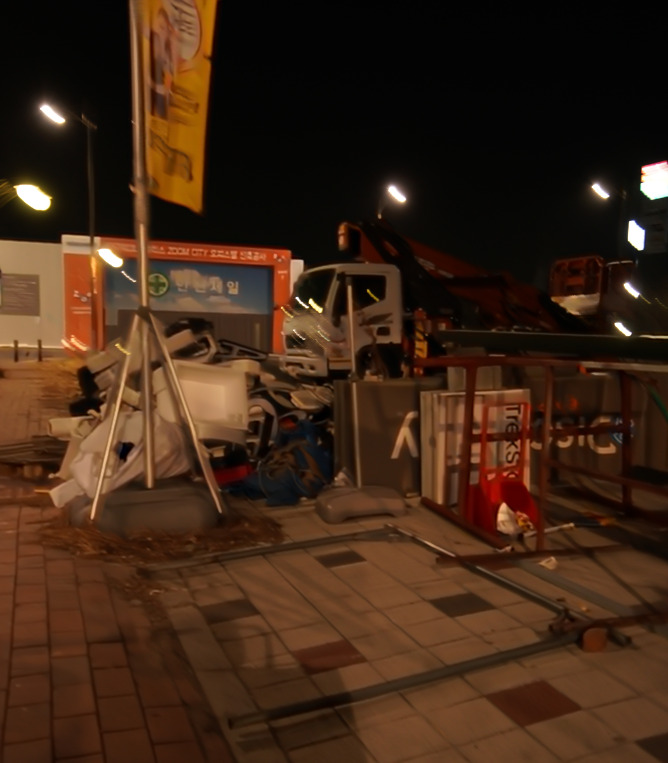}
         \caption{Rest.}
     \end{subfigure}
     \hfill 
     \begin{subfigure}[b]{\gpwr\textwidth}
         \centering
         \includegraphics[trim={0 400 300 0},clip,width=\textwidth]{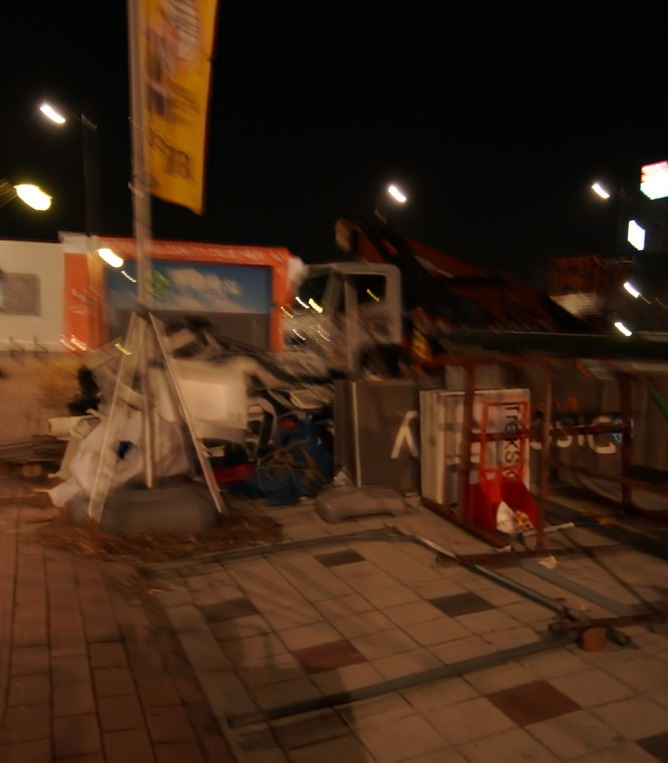}
         \caption{Ours$_U$}
     \end{subfigure}
     \hfill
     \begin{subfigure}[b]{\gpwr\textwidth}
         \centering
         \includegraphics[trim={0 400 300 0},clip,width=\textwidth]{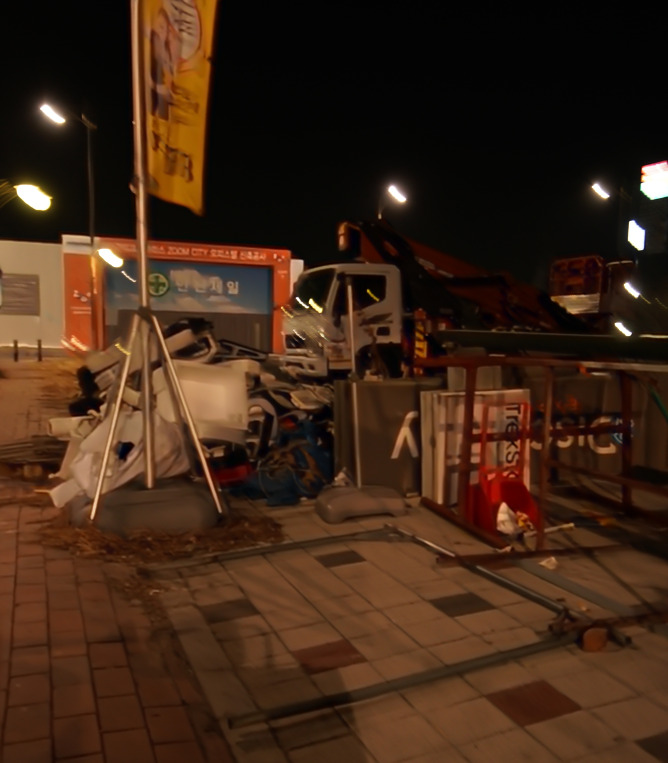}
         \caption{Ours$_{NU}$}
     \end{subfigure} 

        \caption{Visual comparisons of deblurring results on images from the RealBlur-J test set.}
        \label{fig:qual_realj}
        \vspace{-4mm}
\end{figure*}


\begin{figure*}[!hb]
\captionsetup[subfigure]{labelformat=empty}
     \centering
     \begin{subfigure}[b]{\gpwr\textwidth}
         \centering
         \includegraphics[trim={350 30 0 400},clip,width=\textwidth]{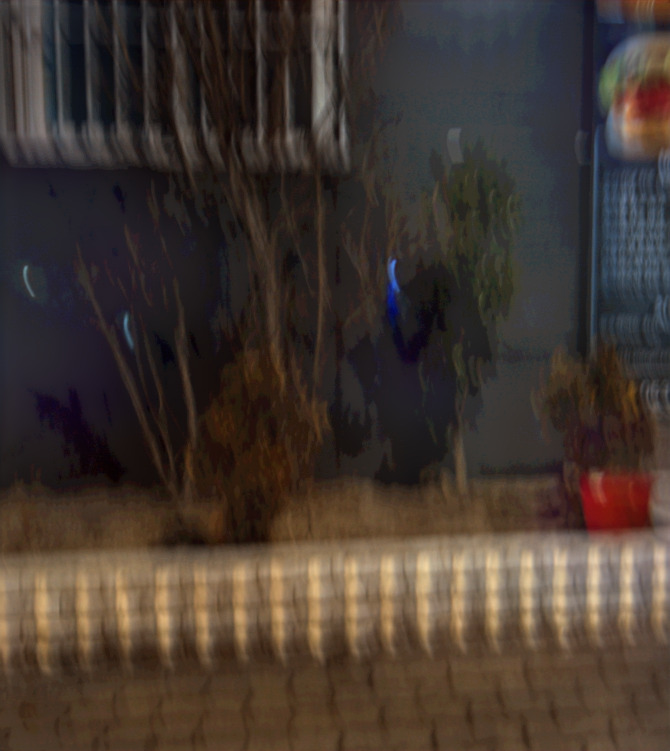}
     \end{subfigure}%
     \hfill
     \begin{subfigure}[b]{\gpwr\textwidth}
         \centering
         \includegraphics[trim={350 30 0 400},clip,width=\textwidth]{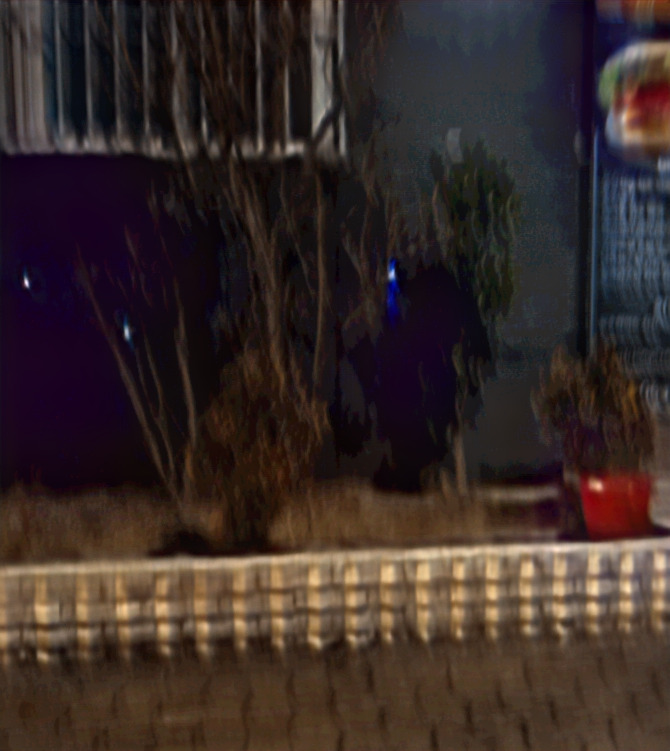}
     \end{subfigure}
     \hfill
     \begin{subfigure}[b]{\gpwr\textwidth}
         \centering
         \includegraphics[trim={350 30 0 400},clip,width=\textwidth]{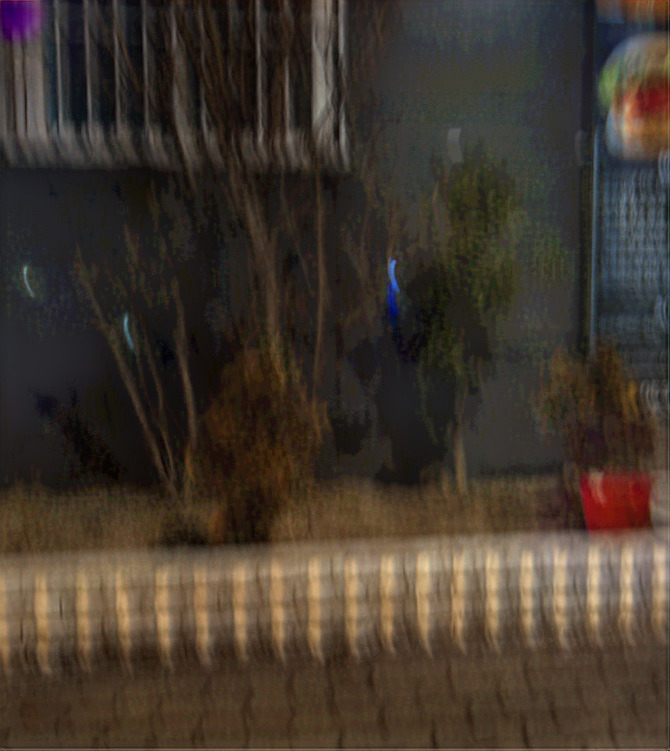}
     \end{subfigure}
     \hfill
     \begin{subfigure}[b]{\gpwr\textwidth}
         \centering
         \includegraphics[trim={350 30 0 415},clip,width=\textwidth]{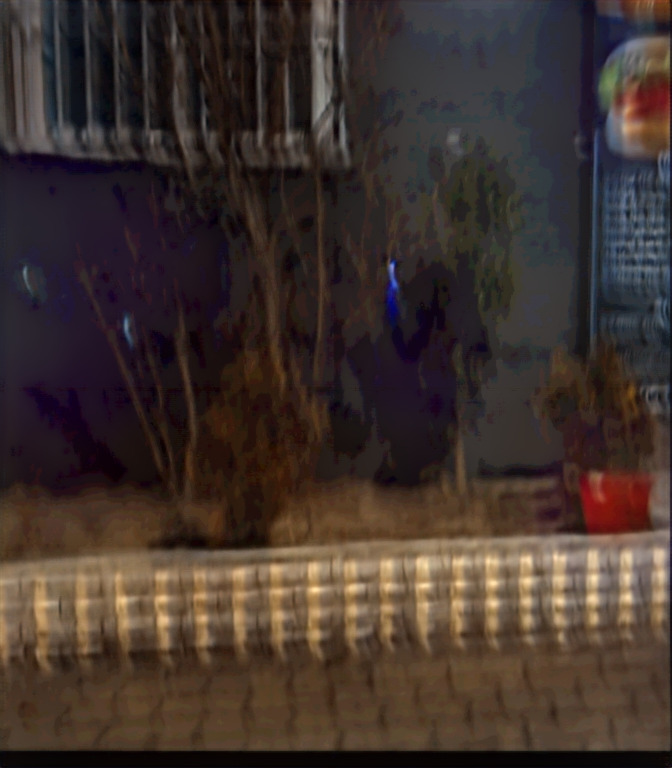}
     \end{subfigure}
     \hfill
     \begin{subfigure}[b]{\gpwr\textwidth}
         \centering
         \includegraphics[trim={350 30 0 400},clip,width=\textwidth]{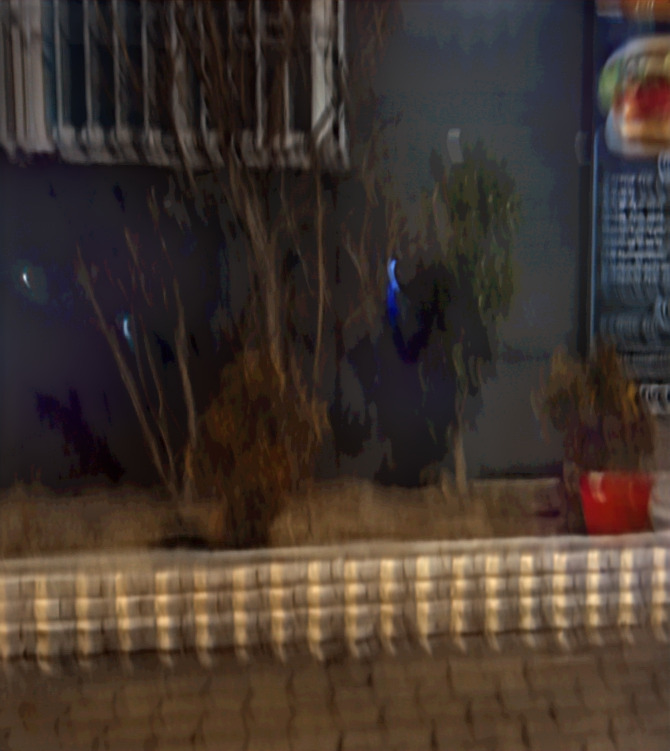}
     \end{subfigure}
     \hfill 
     \begin{subfigure}[b]{\gpwr\textwidth}
         \centering
         \includegraphics[trim={350 30 0 400},clip,width=\textwidth]{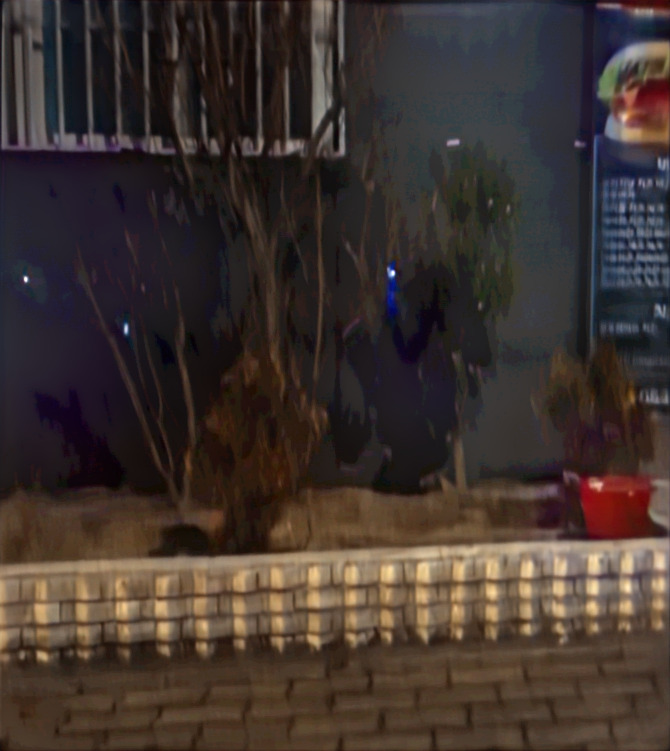}
     \end{subfigure}
     \hfill 
     \begin{subfigure}[b]{\gpwr\textwidth}
         \centering
         \includegraphics[trim={350 30 0 400},clip,width=\textwidth]{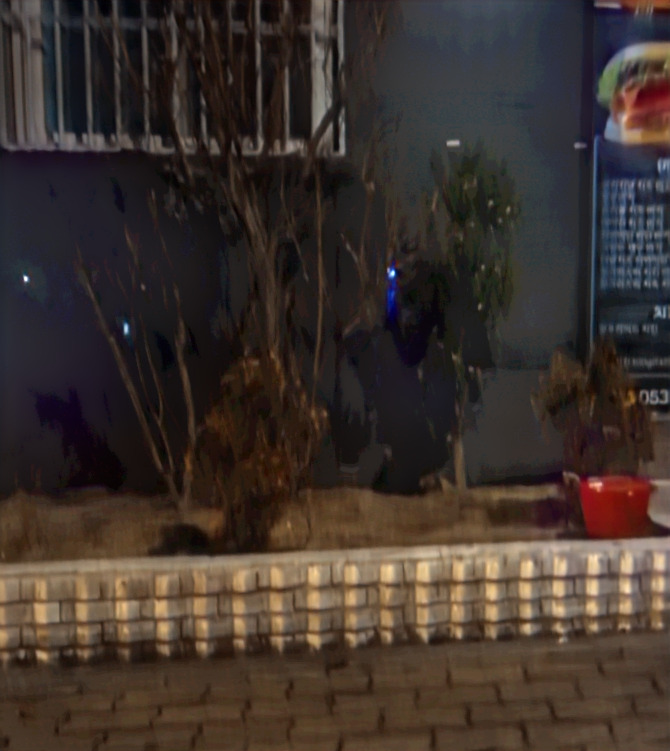}
     \end{subfigure}
     \hfill 
     \begin{subfigure}[b]{\gpwr\textwidth}
         \centering
         \includegraphics[trim={350 30 0 400},clip,width=\textwidth]{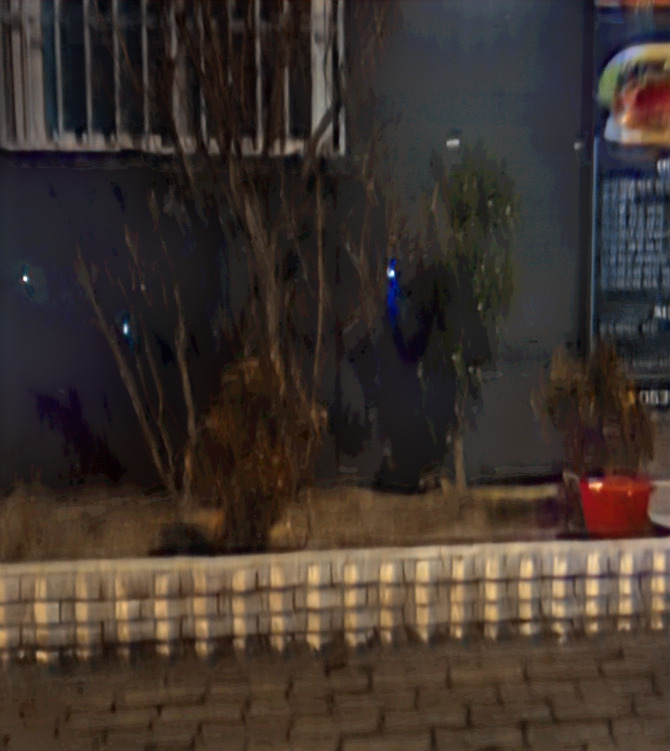}  
     \end{subfigure}
     \hfill
     \begin{subfigure}[b]{\gpwr\textwidth}
         \centering
         \includegraphics[trim={350 30 0 400},clip,width=\textwidth]{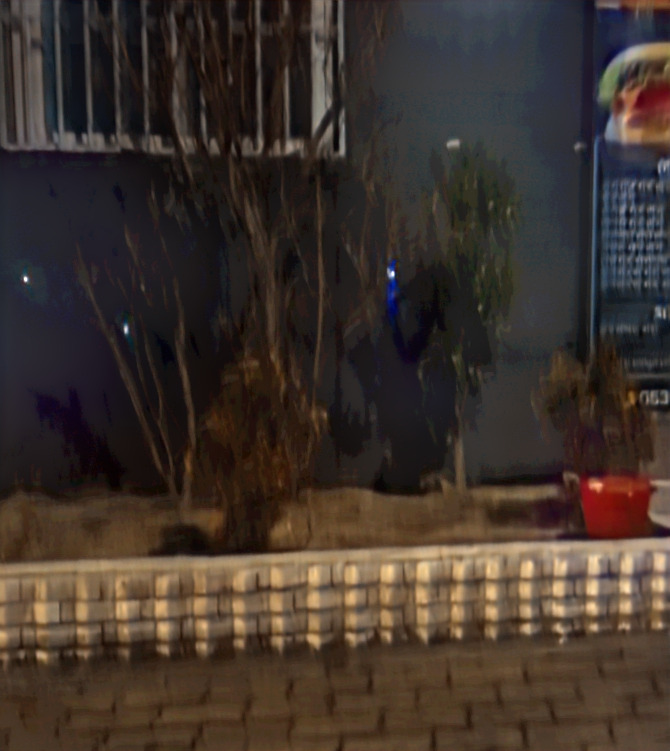}
     \end{subfigure}
     \hfill
     \begin{subfigure}[b]{\gpwr\textwidth}
         \centering
         \includegraphics[trim={400 450 0 50},clip,width=\textwidth]{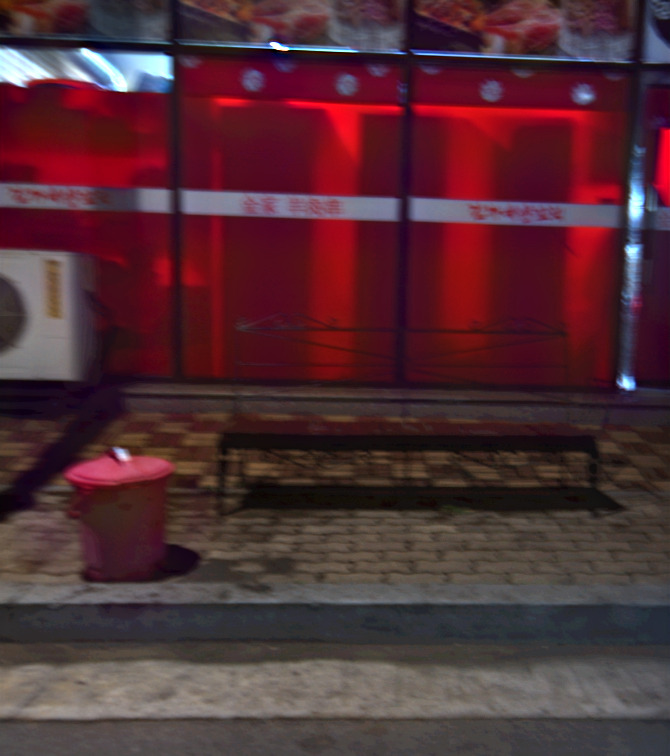}
     \end{subfigure}%
     \hfill
     \begin{subfigure}[b]{\gpwr\textwidth}
         \centering
         \includegraphics[trim={400 450 0 50},clip,width=\textwidth]{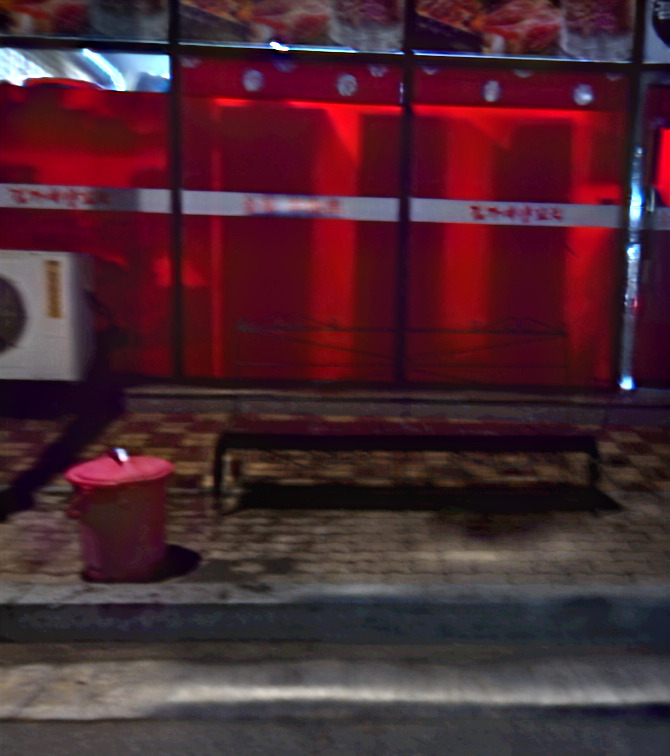}
     \end{subfigure}
     \hfill
     \begin{subfigure}[b]{\gpwr\textwidth}
         \centering
         \includegraphics[trim={400 450 0 50},clip,width=\textwidth]{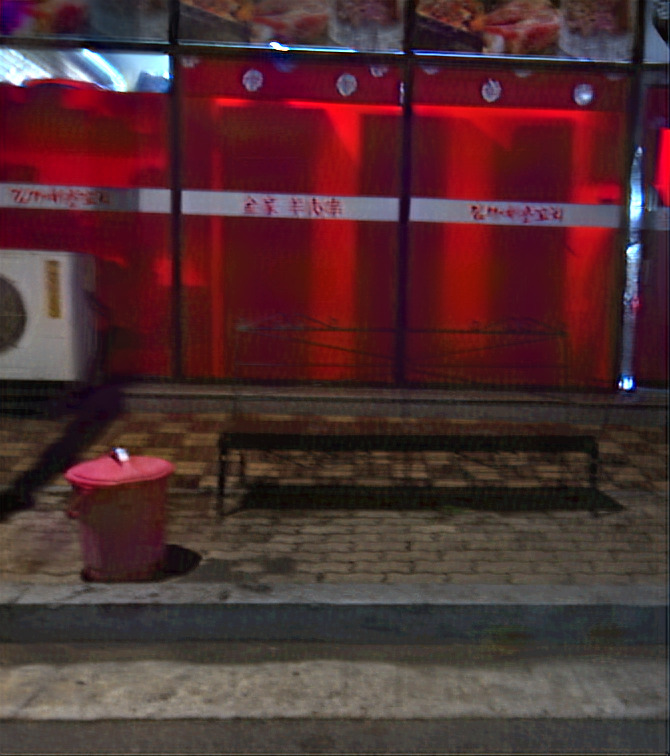}
     \end{subfigure}
     \hfill
     \begin{subfigure}[b]{\gpwr\textwidth}
         \centering
         \includegraphics[trim={400 450 0 60},clip,width=\textwidth]{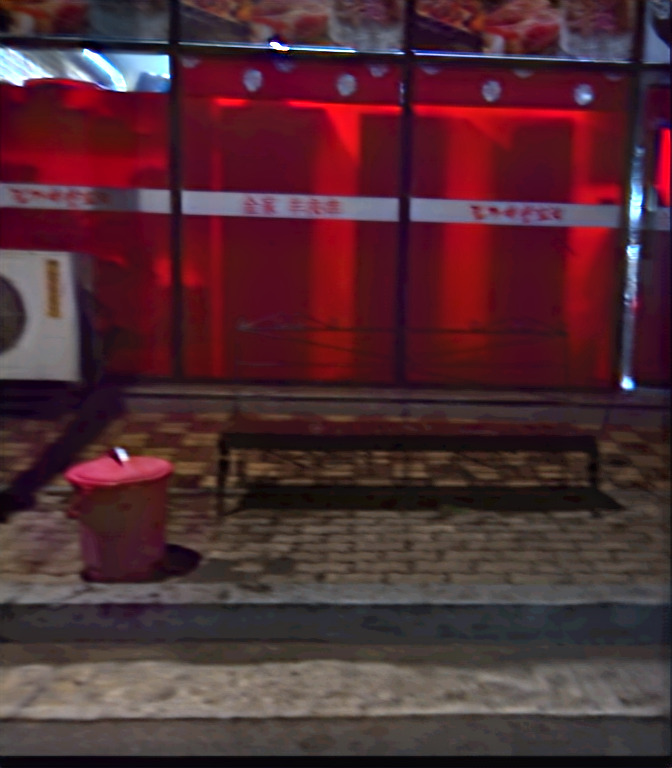}
     \end{subfigure}
     \hfill
     \begin{subfigure}[b]{\gpwr\textwidth}
         \centering
         \includegraphics[trim={400 450 0 50},clip,width=\textwidth]{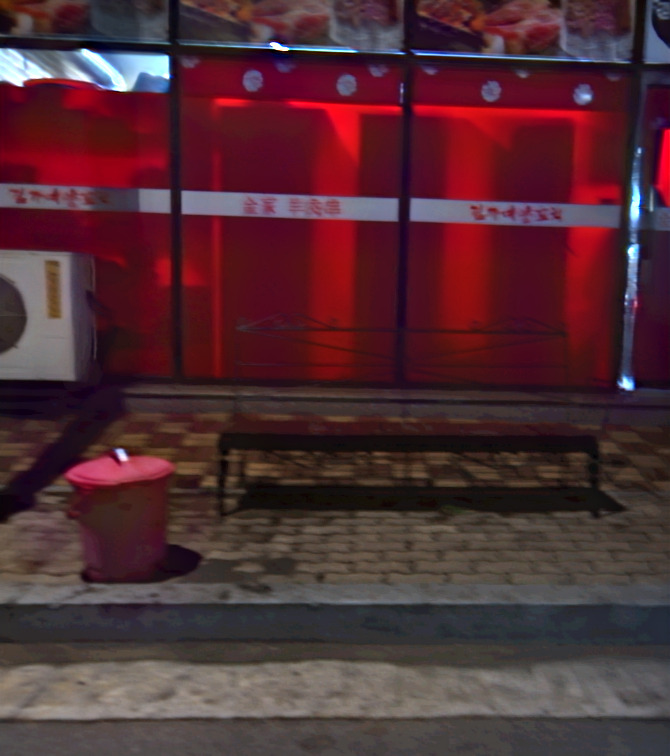}
     \end{subfigure}
     \hfill 
     \begin{subfigure}[b]{\gpwr\textwidth}
         \centering
         \includegraphics[trim={400 450 0 50},clip,width=\textwidth]{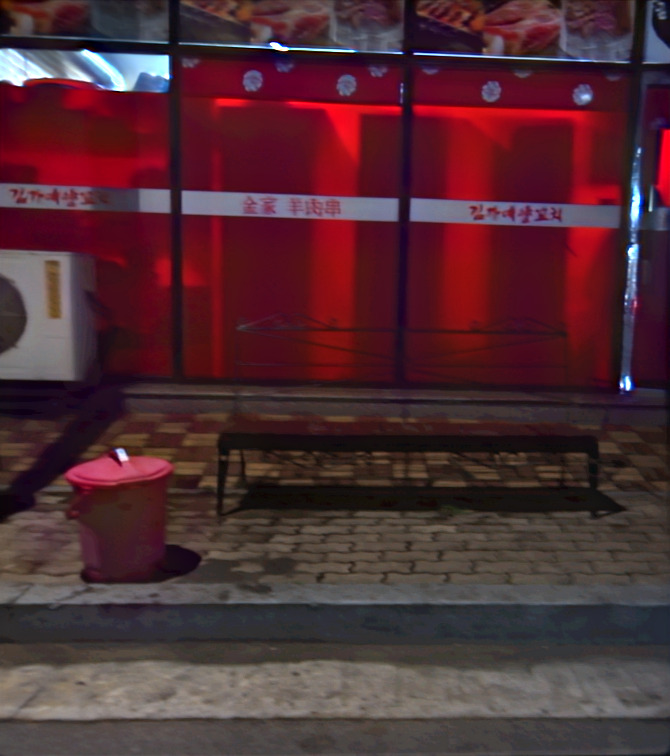}
     \end{subfigure}
     \hfill 
     \begin{subfigure}[b]{\gpwr\textwidth}
         \centering
         \includegraphics[trim={400 450 0 50},clip,width=\textwidth]{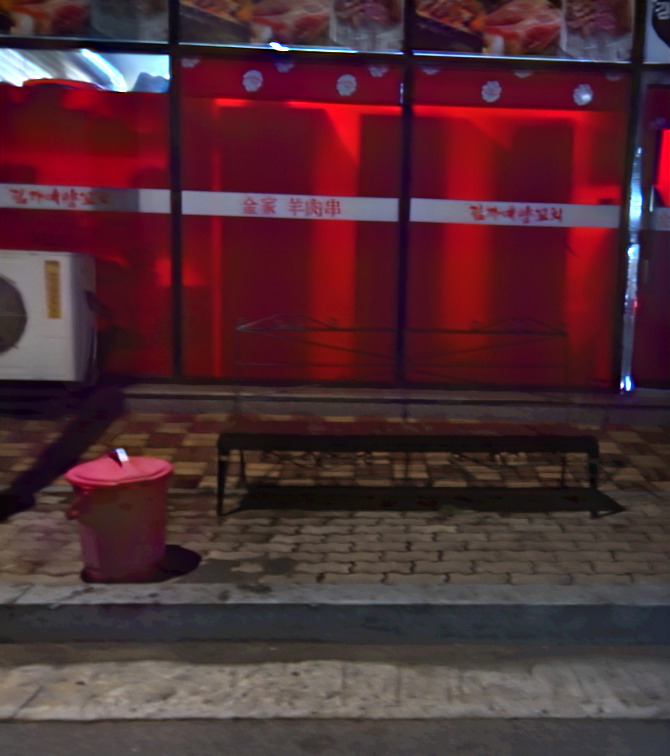}
     \end{subfigure}
     \hfill 
     \begin{subfigure}[b]{\gpwr\textwidth}
         \centering
         \includegraphics[trim={400 450 0 50},clip,width=\textwidth]{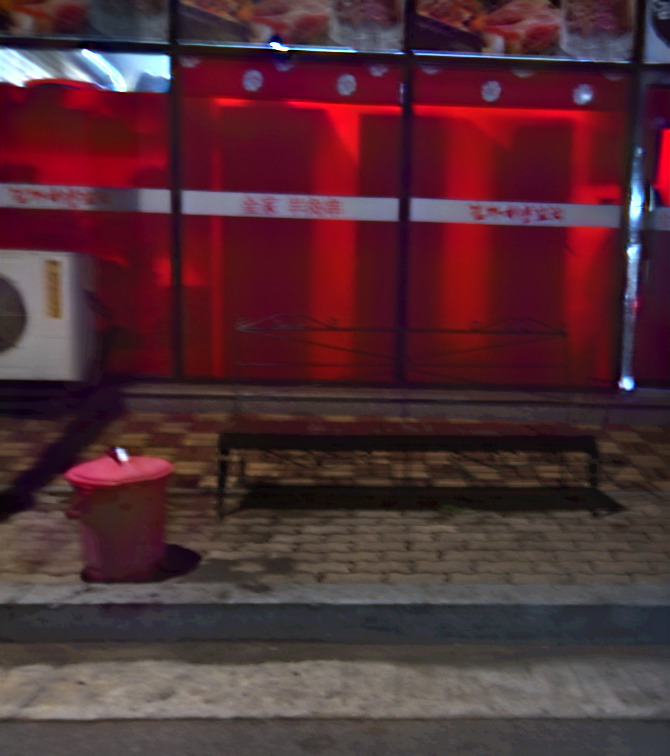}  
     \end{subfigure}
     \hfill
     \begin{subfigure}[b]{\gpwr\textwidth}
         \centering
         \includegraphics[trim={400 450 0 50},clip,width=\textwidth]{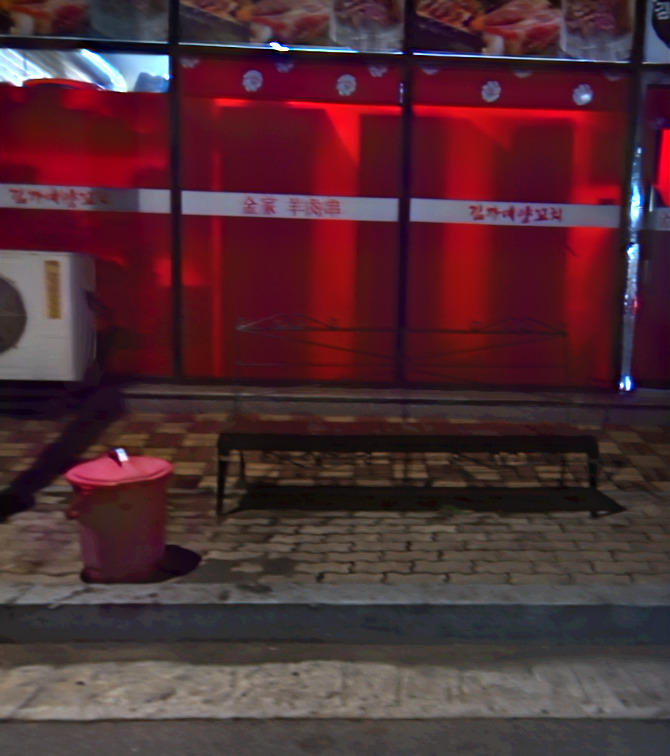}
     \end{subfigure}
     \hfill
     \begin{subfigure}[b]{\gpwr\textwidth}
         \centering
         \includegraphics[trim={0 400 370 100},clip,width=\textwidth]{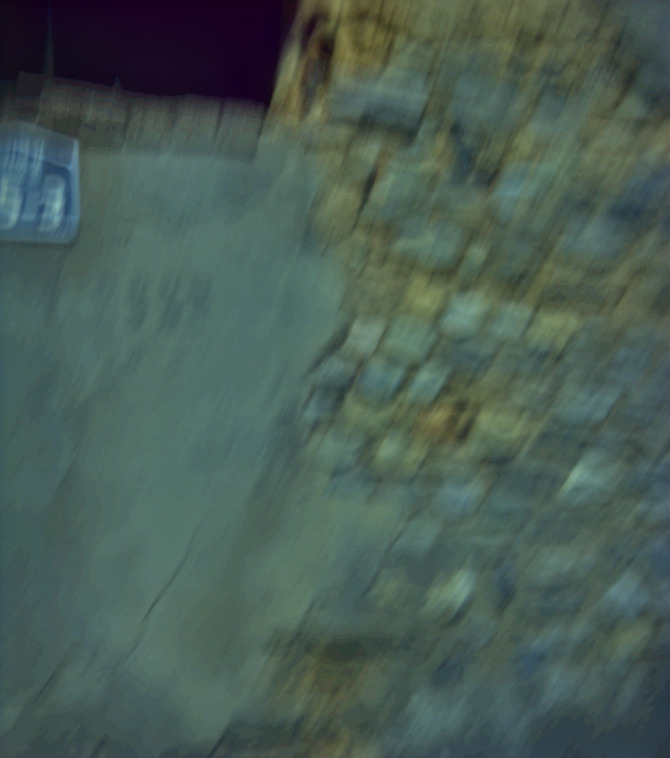}
        \caption{IP}
     \end{subfigure}%
     \hfill
     \begin{subfigure}[b]{\gpwr\textwidth}
         \centering
         \includegraphics[trim={0 400 370 100},clip,width=\textwidth]{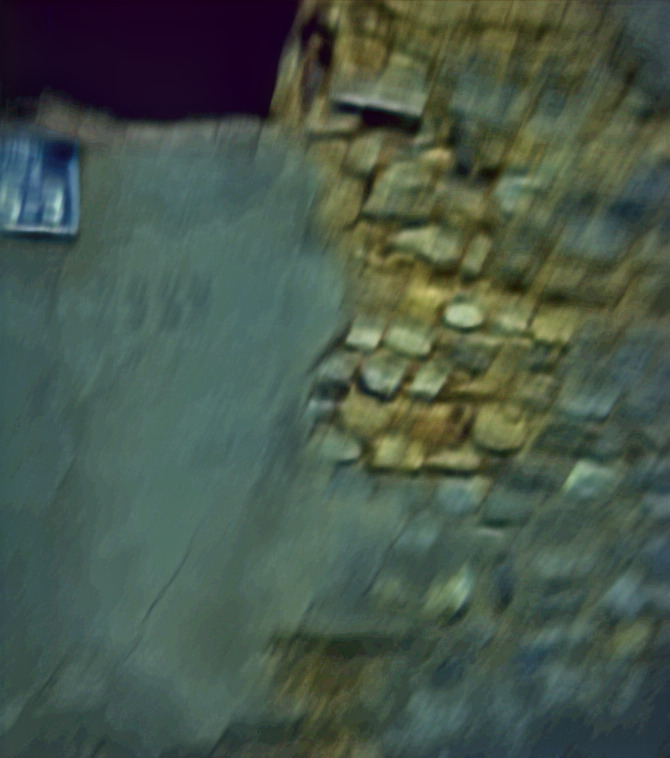}
        \caption{SRN}
     \end{subfigure}
     \hfill
     \begin{subfigure}[b]{\gpwr\textwidth}
         \centering
         \includegraphics[trim={0 400 370 100},clip,width=\textwidth]{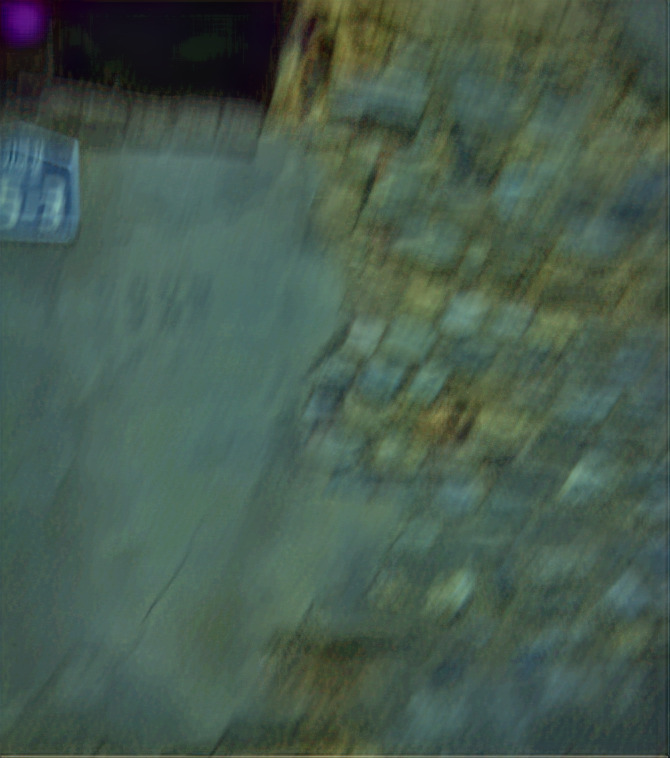}
        \caption{DG2}
     \end{subfigure}
     \hfill
     \begin{subfigure}[b]{\gpwr\textwidth}
         \centering
         \includegraphics[trim={0 400 360 100},clip,width=\textwidth]{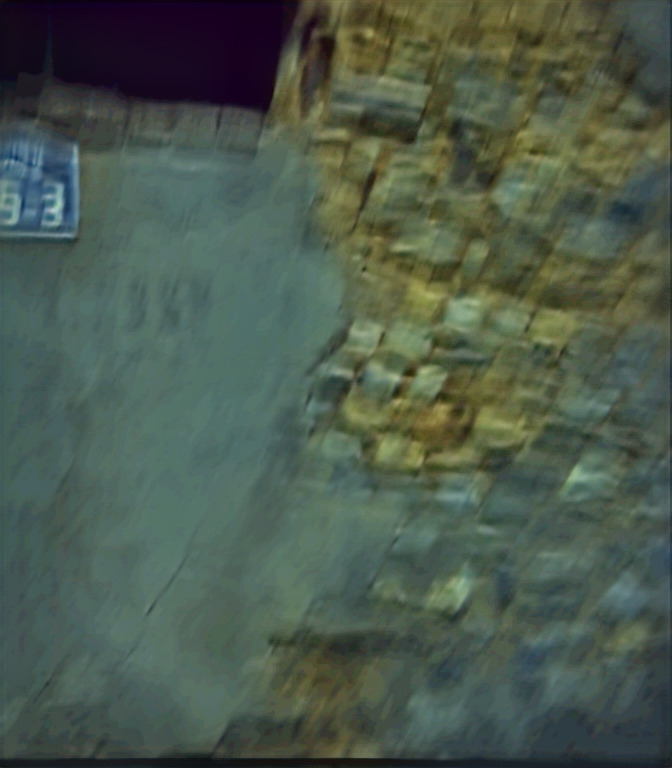}
        \caption{DMPHN}
     \end{subfigure}
     \hfill
     \begin{subfigure}[b]{\gpwr\textwidth}
         \centering
         \includegraphics[trim={0 400 370 100},clip,width=\textwidth]{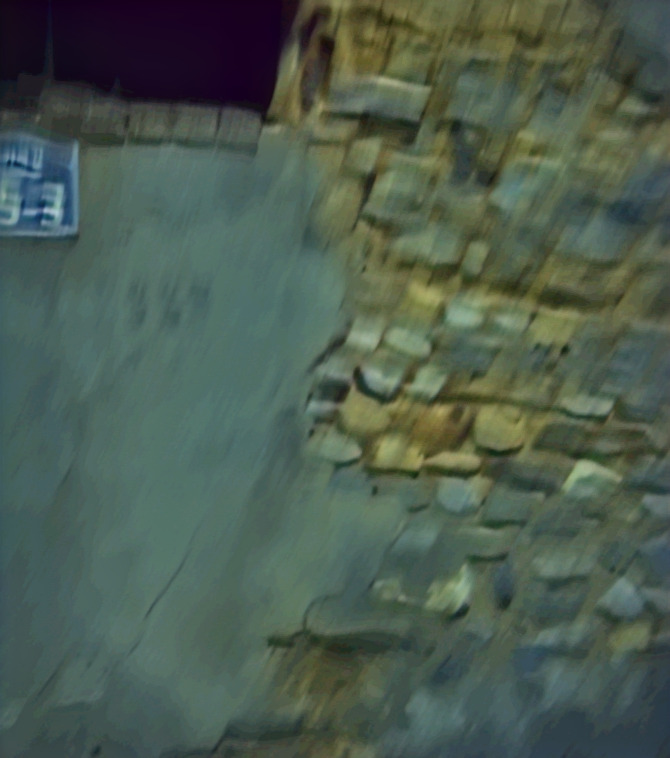}
        \caption{MPR}
     \end{subfigure}
     \hfill 
     \begin{subfigure}[b]{\gpwr\textwidth}
         \centering
         \includegraphics[trim={0 400 370 100},clip,width=\textwidth]{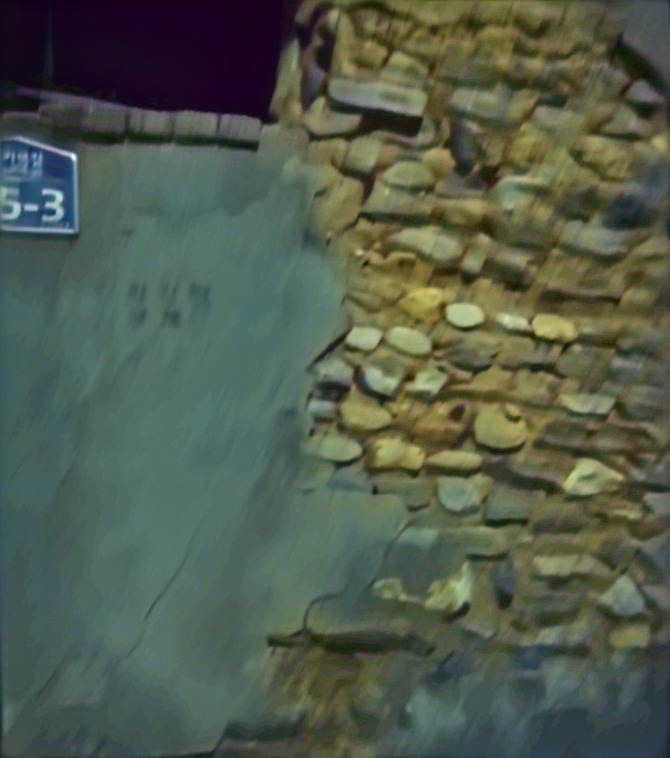}
        \caption{Uform.}
     \end{subfigure}
     \hfill 
     \begin{subfigure}[b]{\gpwr\textwidth}
         \centering
         \includegraphics[trim={0 400 370 100},clip,width=\textwidth]{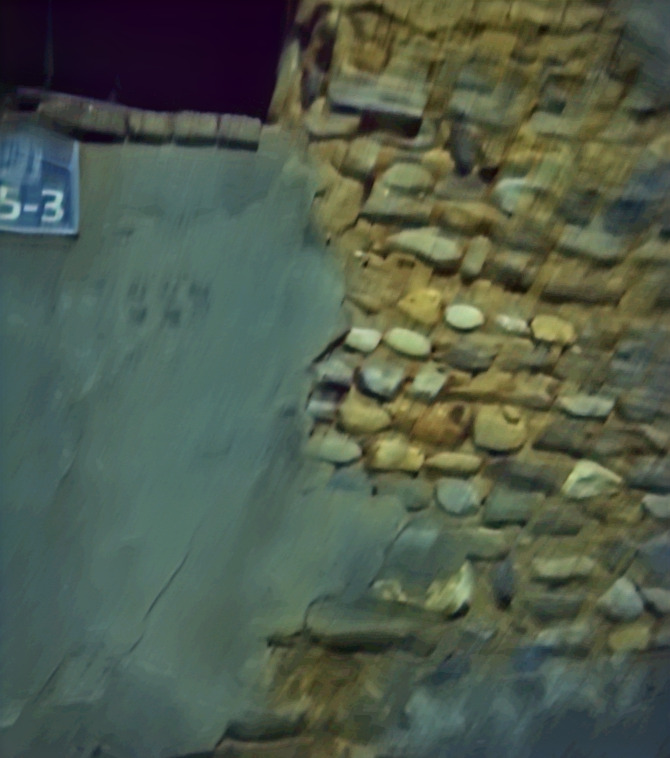}
        \caption{Rest.}
     \end{subfigure}
     \hfill 
     \begin{subfigure}[b]{\gpwr\textwidth}
         \centering
         \includegraphics[trim={0 400 370 100},clip,width=\textwidth]{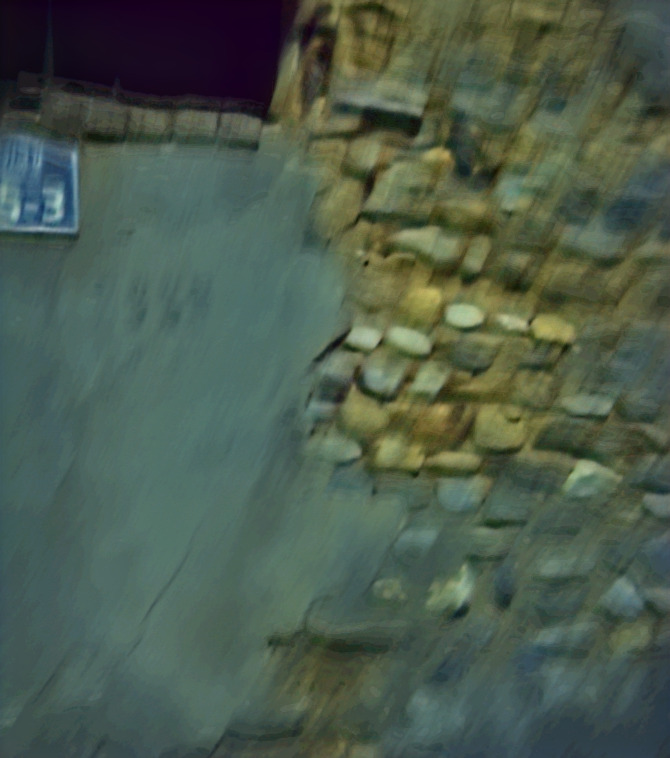}
        \caption{Ours$_u$}
     \end{subfigure}
     \hfill
     \begin{subfigure}[b]{\gpwr\textwidth}
         \centering
         \includegraphics[trim={0 400 370 100},clip,width=\textwidth]{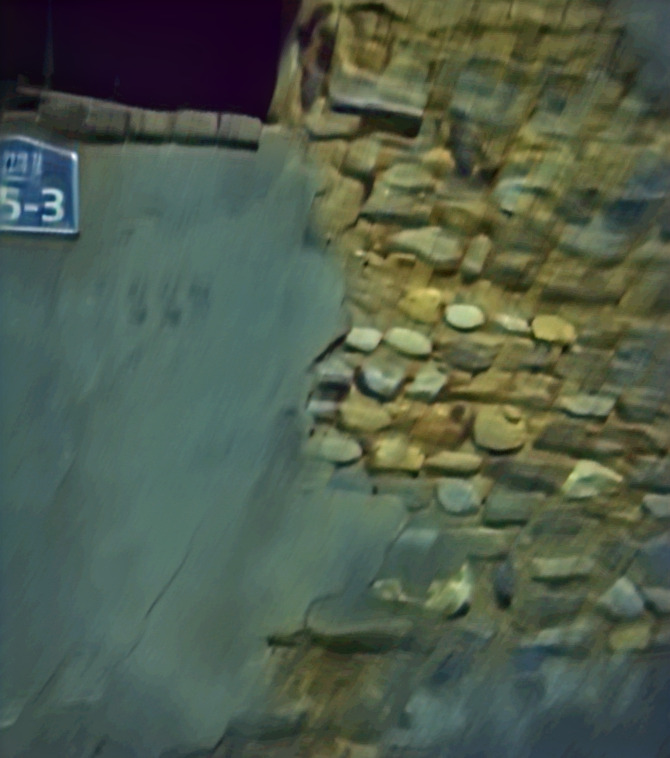}
        \caption{Ours$_{nu}$}
     \end{subfigure}
        \caption{Visual comparisons of deblurring results on images from the RealBlur-R test set. We have post-processed the images using \cite{jiang2021enlightengan} for easier visual comparison.}
        \label{fig:qual_realr}
\end{figure*}
\vspace{-4mm}
\subsection{Network Analysis}
\label{sec:ablation}
\begin{table}[t]
\centering
\caption{Quantitative comparison of different ablations of our network on GoPro testset. $Prog$., $SA$. $CA$, $CLA$, $V$, $\Delta$, $PAS$ represent progressive restoration, self-attention, cross-attention, cross-level-attention, adaptive kernel, adaptive offset and non-uniform pixel-adaptive-sampling, respectively.
}
\label{table:ablation}
\resizebox{0.4\textwidth}{!}{
\begin{tabular}{|c|c|c|c|c|c|c|c|c|c|}
\hline
Design & $Prog.$ & $SA$ & $CA$ & $CLA$ & Mask & $V$ & $\Delta$ & $PAS$ & PSNR \\
\hline
Baseline & \xmark & \xmark & \xmark & \xmark & \xmark & \xmark  & \xmark & \xmark & 30.25\\
Net2 & \xmark & \xmark & \xmark & \xmark & \xmark & \cmark & \xmark & \xmark & 30.81\\
Net3 & \xmark & \cmark & \xmark & \xmark & \cmark & \xmark & \xmark & \xmark  & 30.76\\
Net4 & \xmark & \cmark & \cmark & \xmark  & \cmark & \xmark  & \xmark & \xmark  & 30.93\\
Net5 & \xmark & \cmark & \xmark & \cmark  & \cmark & \xmark  & \xmark & \xmark  & 31.04\\ 
Net6 & \xmark & \cmark & \cmark & \cmark  & \cmark & \cmark & \xmark & \xmark & 31.38\\
Net7 & \xmark & \cmark & \cmark & \xmark  & \cmark & \cmark & \cmark & \xmark  & 31.85\\ 
Net8 & \xmark & \cmark & \cmark & \xmark  & \xmark & \cmark & \cmark & \xmark  & 31.85\\ 
\underline{Net9} & \xmark & \cmark & \cmark & \cmark  & \cmark & \cmark & \cmark & \xmark  & \underline{32.02} \\
Net10 & \cmark & \cmark & \cmark & \cmark  & \cmark & \cmark & \cmark & \xmark  & 32.10 \\
\textbf{Net11} & \cmark & \cmark & \cmark & \cmark  & \cmark & \cmark & \cmark & \cmark  & \textbf{32.76} \\
\hline 
\end{tabular}
}
\end{table}

\begin{table}[t]
\centering
\caption{Effect of different sampling strategies on the network performance and complexity. This experiment is performed on the HIDE testset.
} 
\label{table:sampling}
\resizebox{0.35\textwidth}{!}{
\begin{tabular}{|c|c|c|c|c|c|}
\hline
Method & U & NU & NU$^{+10\%}$ & NU$^{+20\%}$ & NU$^{-10\%}$  \\
\hline
PSNR & 29.98 & 31.09 & 31.14 & 31.18 & 31.01 \\
GFLOPs & 136 & 140 & 170 & 210 & 126 \\
Runtime (s) & 0.77 & 0.79 & 0.91 &1.15 & 0.68 \\
\hline 
\end{tabular}
}
\end{table}

\begin{table}[t]
\centering
\caption{Introducing our adaptive sampling strategy to some of the existing works: DMPHN and SRN. For DMPHN, we have selected the DMPHN(1-2-4-8) as the baseline. The significant improvement in the performance demonstrates the effectiveness of our approach.
}
\label{table:extension}
\resizebox{0.35\textwidth}{!}{
\begin{tabular}{|c|c|c|c|c|}
\hline
Method & SRN\_{U} & SRN\_{NU} & DMPHN\_{U}\tiny{(1-2-4-8)} & DMPHN\_{NU}\\
\hline
PSNR & 30.26 & 31.29 & 30.25 & 31.51 \\
Runtime(s) & 1.2 & 1.4 & 0.3 & 0.33\\
\hline
\end{tabular}
}
\end{table}
\newcommand\visw{0.16}
\begin{figure*}[htb]
     \centering
     \begin{subfigure}[b]{\visw\textwidth}
         \centering
         \includegraphics[width=\textwidth]{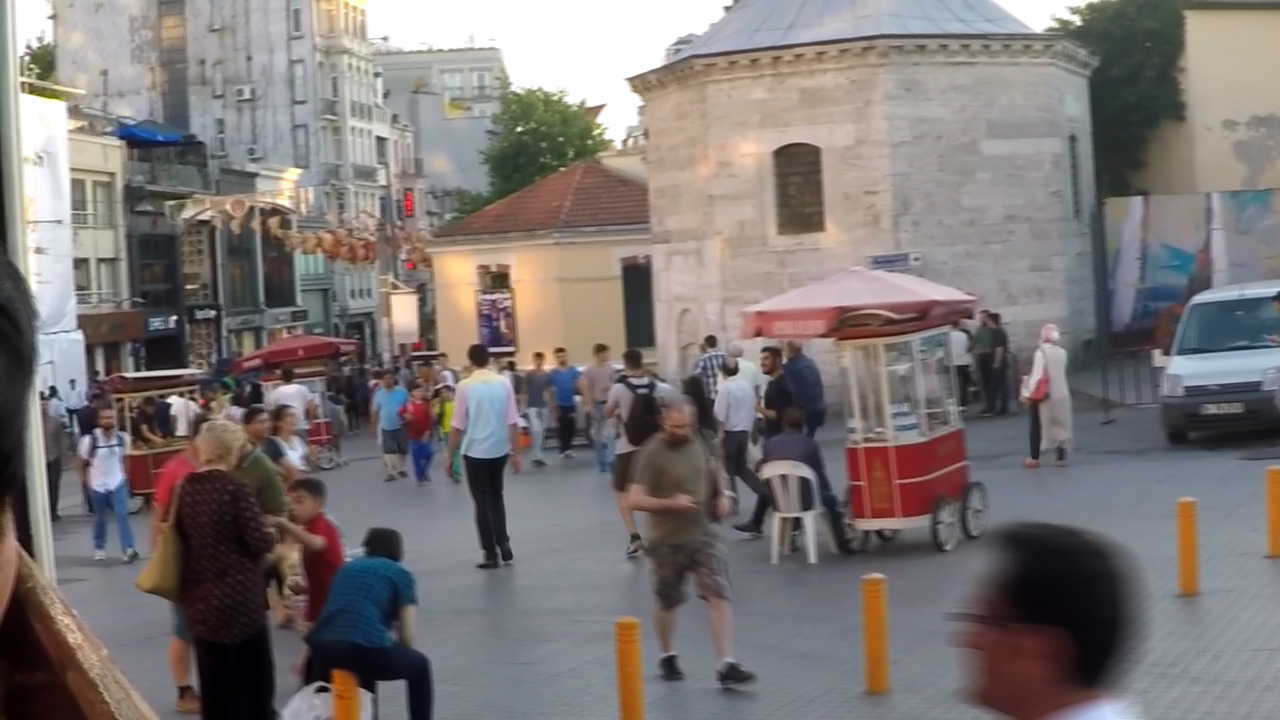}
     \end{subfigure}
     \begin{subfigure}[b]{\visw\textwidth}
         \centering
         \includegraphics[width=\textwidth]{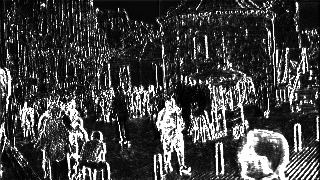}
     \end{subfigure}
     \begin{subfigure}[b]{\visw\textwidth}
         \centering
         \includegraphics[width=\textwidth]{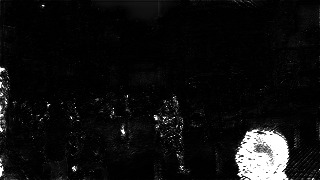}
     \end{subfigure}
     \begin{subfigure}[b]{\visw\textwidth}
         \centering
         \includegraphics[bb=60 80 400 280,clip=True,width=\textwidth]{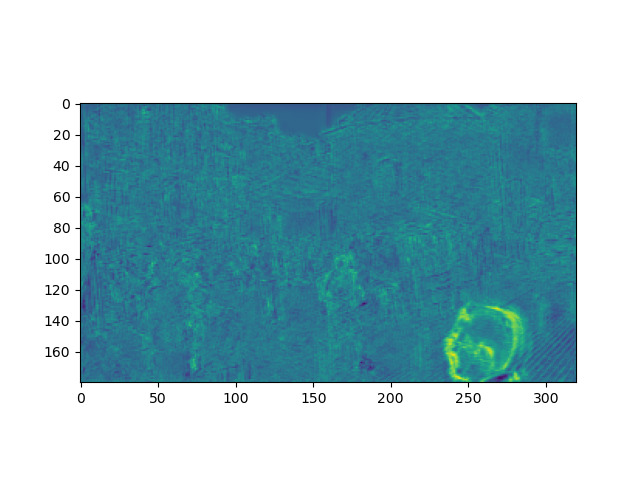}
     \end{subfigure}
     \begin{subfigure}[b]{\visw\textwidth}
         \centering
         \includegraphics[bb=60 80 400 280,clip=True,width=\textwidth]{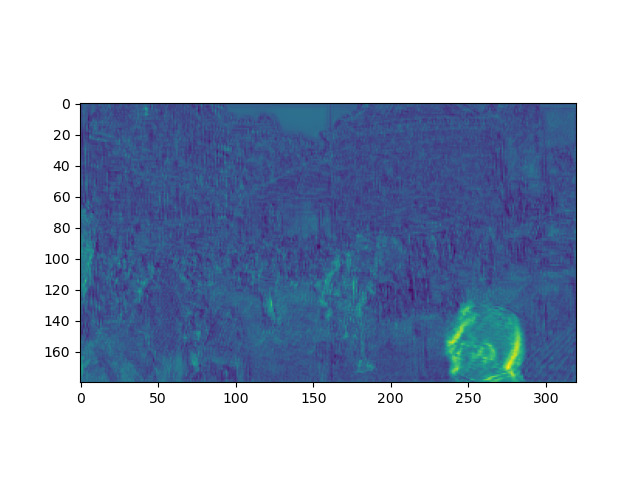}
     \end{subfigure}
     \begin{subfigure}[b]{\visw\textwidth}
         \centering
         \includegraphics[bb=60 80 400 280,clip=True,width=\textwidth]{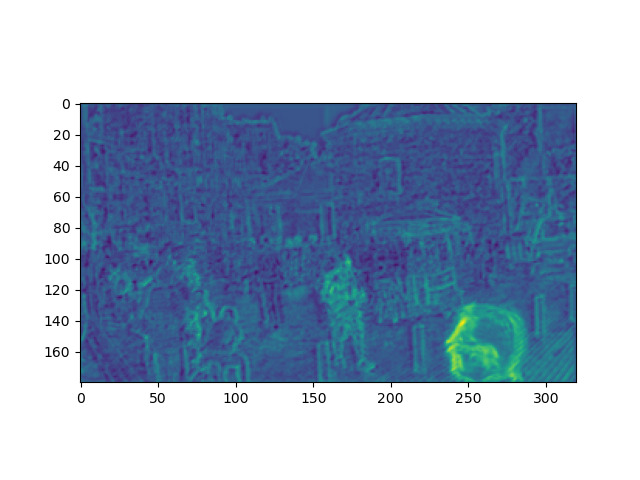}
     \end{subfigure}\\
     
     \begin{subfigure}[b]{\visw\textwidth}
         \centering
         \includegraphics[width=\textwidth]{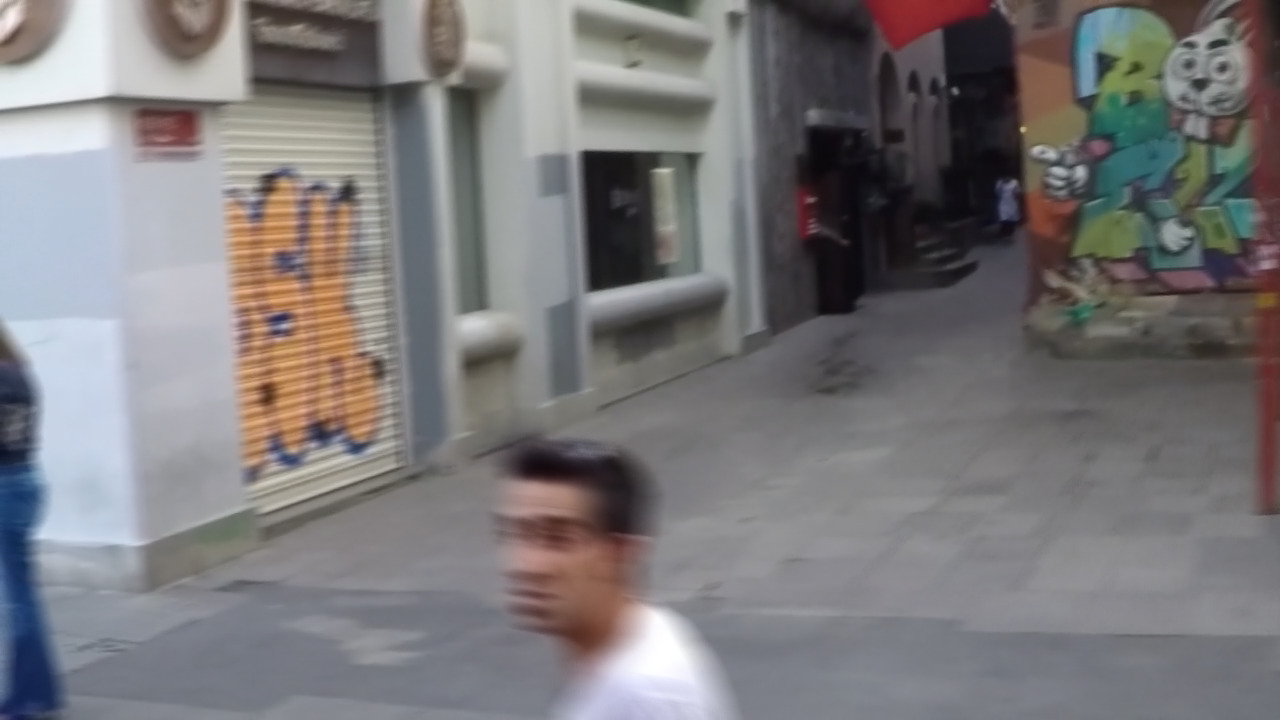}
     \end{subfigure}
     \begin{subfigure}[b]{\visw\textwidth}
         \centering
         \includegraphics[width=\textwidth]{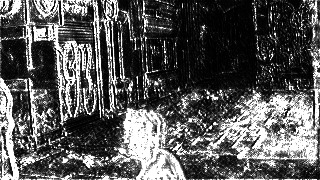}
     \end{subfigure}
     \begin{subfigure}[b]{\visw\textwidth}
         \centering
         \includegraphics[width=\textwidth]{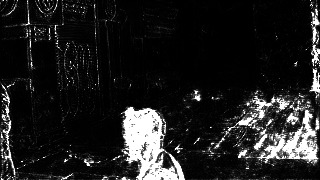}
     \end{subfigure}
     \begin{subfigure}[b]{\visw\textwidth}
         \centering
         \includegraphics[bb=60 80 400 280,clip=True,width=\textwidth]{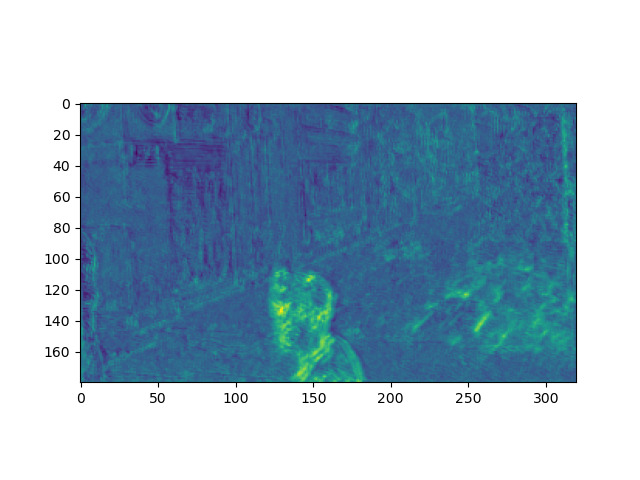}
     \end{subfigure}
     \begin{subfigure}[b]{\visw\textwidth}
         \centering
         \includegraphics[bb=60 80 400 280,clip=True,width=\textwidth]{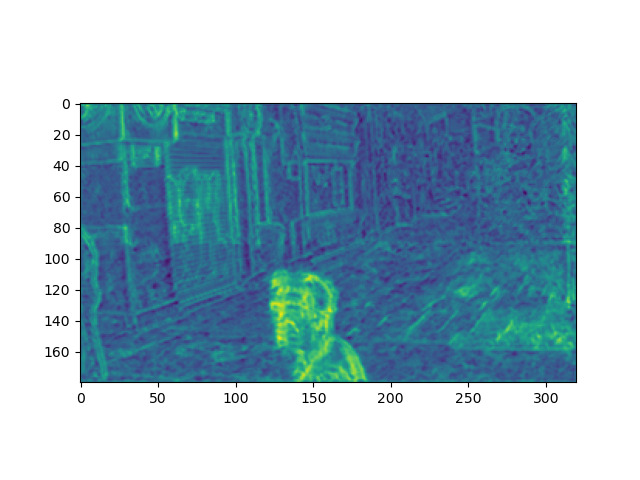}
     \end{subfigure}
     \begin{subfigure}[b]{\visw\textwidth}
         \centering
         \includegraphics[bb=60 80 400 280,clip=True,width=\textwidth]{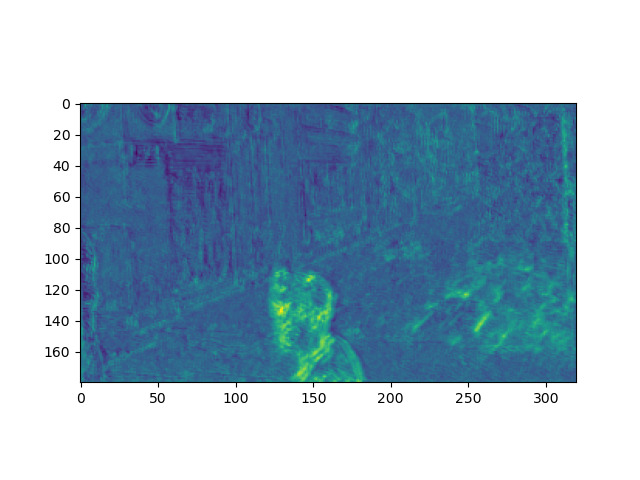}
     \end{subfigure}\\ 
     \begin{subfigure}[b]{\visw\textwidth}
         \centering
         \includegraphics[width=\textwidth]{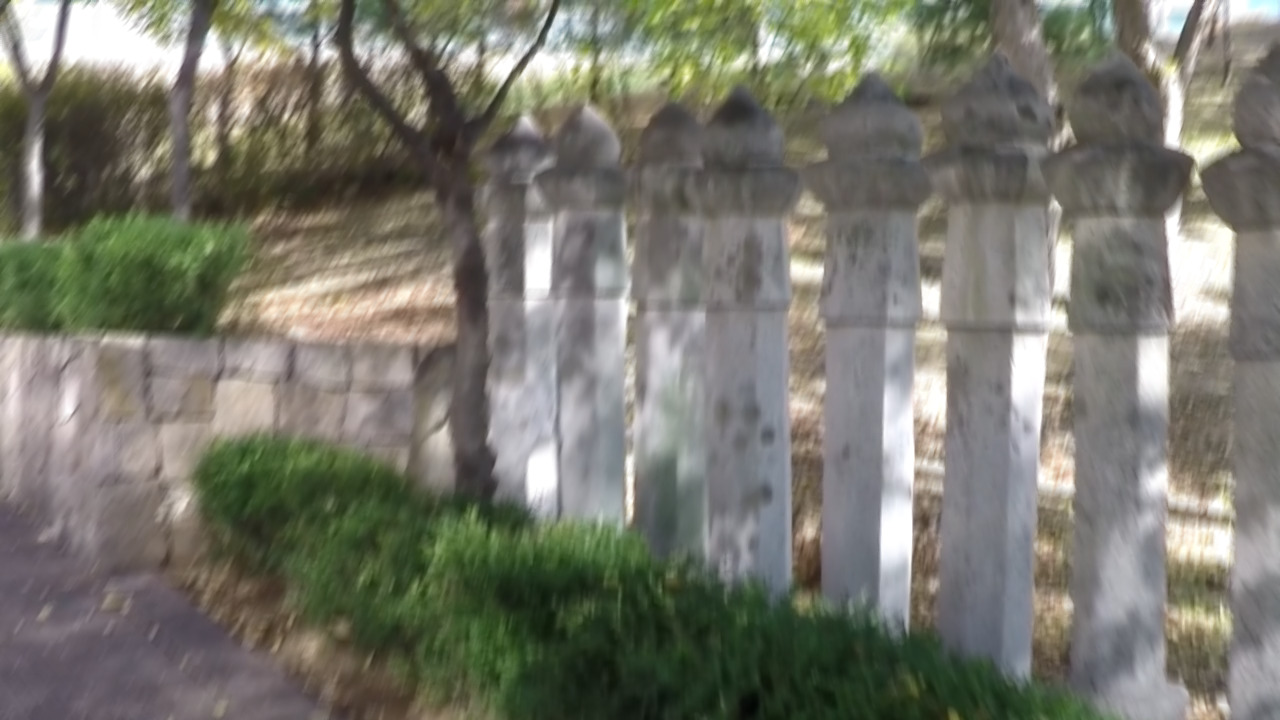}
         \caption{$Input$}
     \end{subfigure}
     \begin{subfigure}[b]{\visw\textwidth}
         \centering
         \includegraphics[width=\textwidth]{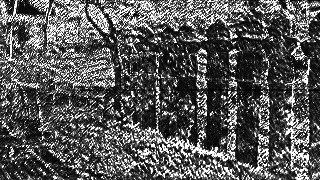}
         \caption{$Att_{Stage_1}$}
     \end{subfigure}
     \begin{subfigure}[b]{\visw\textwidth}
         \centering
         \includegraphics[width=\textwidth]{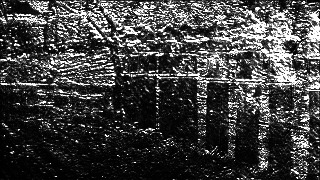}
         \caption{$Att_{Stage_3}$}
     \end{subfigure}
     \begin{subfigure}[b]{\visw\textwidth}
         \centering
         \includegraphics[bb=60 80 400 280,clip=True,width=\textwidth]{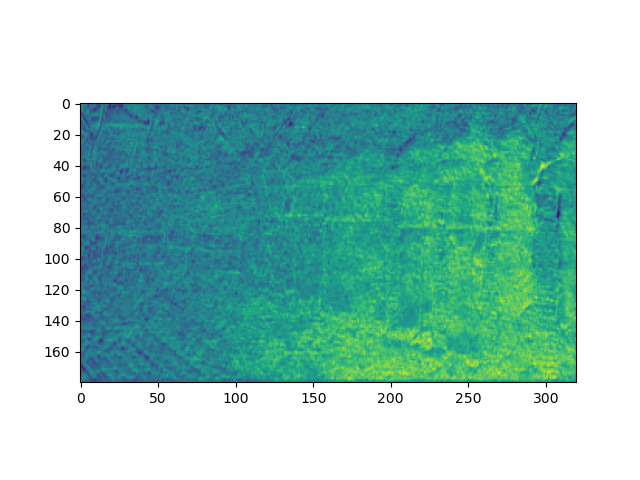}
         \caption{$\Delta_{Stage_1}$}
     \end{subfigure}
     \begin{subfigure}[b]{\visw\textwidth}
         \centering
         \includegraphics[bb=60 80 400 280,clip=True,width=\textwidth]{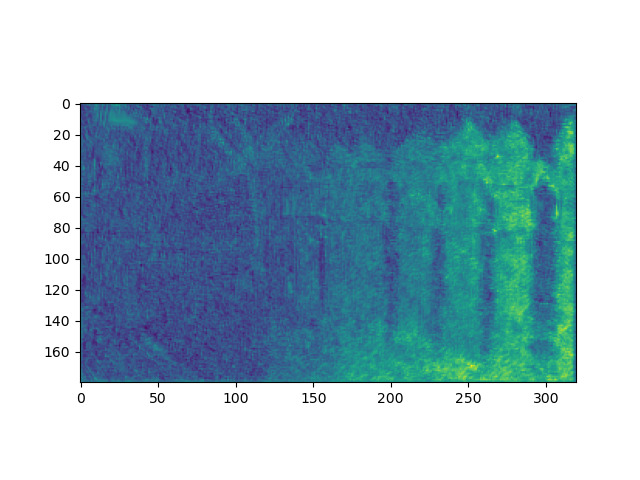}
         \caption{$\Delta_{Stage_3}$}
     \end{subfigure}
     \begin{subfigure}[b]{\visw\textwidth}
         \centering
         \includegraphics[bb=60 80 400 280,clip=True,width=\textwidth]{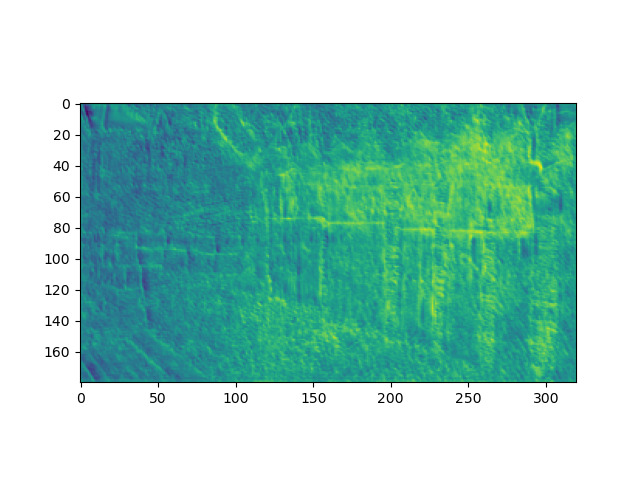}
         \caption{$V$}
     \end{subfigure}
        \caption{The $2^{nd}$ and $3^{rd}$ columns show attention map for the $1^{st}$ and $3^{rd}$ stage of the network. The $4^{th}$ and $5^{th}$ columns show the spatial distribution of the horizontal-offset for the filter. The $6^{th}$ column shows the variance of the predicted kernel values.}
        \label{fig:visualization}
        \vspace{-4mm}
\end{figure*}


In Table \ref{table:ablation}, we analyze the effect of individual modules on our network's performance, using $1111$ test images from GoPro dataset~\cite{nah2017deep}. As shown in Fig. \ref{fig:arch}, we use 3 hierarchical stages, where the bottom stage is the same as in our original network with uniform sampling, and the subsequent stages deploy pixel-adaptive sampling while progressively restoring the image. We take the final output from the top branch. Although the training performance and the quantitative results got better with the increase in the number of stages, beyond 3 the improvement was marginal. This led us to the choice of using 3 stages, keeping a good balance between efficiency and performance.

First, we will analyze the effectiveness of the pixel-adaptive processing modules: self-attention, and pixel-adaptive filtering, followed by their complementary behavior in a global-local fusion strategy. Next, we will discuss the role of pixel-adaptive sampling in improving the capabilities of these attention and the filtering modules for handling non-uniformly distributed image degradation. 
\newline As the use of local convolution and global attention together \cite{bello2019attention} or replacing local convolution with attention \cite{ramachandran2019stand} has been explored recently for image recognition tasks, we further analyze it for image restoration tasks. The baseline network in Table \ref{table:ablation} consists of standard convolutional and down/up-sampling layers with a similar structure and number of parameters as our final model. Net2 and Net3 show the utility of pixel-adaptive kernel and self-attention operation. Similarly, for better information flow between different layers of encoder-decoder and different stages, we analyze the advantages of cross encoder-decoder attention and cross-stage attention (instead of simple addition or skip connection) in Net4 and Net5, respectively (vs. Net3). Next, we observe that the advantages of SA and PDF modules are complementary, and their union leads to better performance (Net5 vs. Net6/7). We also analyze the role of both adaptive weights and the adaptive local neighborhood for the PDF module. As shown quantitatively in Table \ref{table:ablation} (Net7 and Net9) and visualized in Figure \ref{fig:visualization}, adaptiveness of the offsets along with the weights perform better as it satisfies the need for directional local filters. A concurrent work \cite{katharopoulos2020transformers} also explored a similar strategy for reducing the computational overload of self-attention for different generation and classification tasks. But, it does not utilize any spatial or channel adaptability before aggregating or distributing global information to each pixel. In contrast, we use learnable masks ($M_C$ and $M_S$ in Eq. \ref{eq:mask}) to emphasize only the crucial information and suppress the unnecessary ones to mitigate the ill effect of propagating noisy or unwanted information. To validate this, we remove spatial and channel masks and train our network (Net8). The PSNR drop of 0.18 dB (Net9 vs Net8) supports our claim. Further, following \cite{katharopoulos2020transformers}, we train a variant of our network where we replace softmax with `elu' non-linearity for generating the spatial and pixelwise attention maps $Q$ and $P$, respectively. We found the PSNR of the `elu' model to drop by 0.15 dB compared to Net9. We argue that, for image restoration, softmax allows us to generate normalized attention weights over all pixels depending on their relative importance, similar to the original self-attention design in \cite{vaswani2017attention}, which `elu' does not allow implicitly. The normalized weights also make the training process more stable, ultimately obtaining better accuracy.
\newline We also tried to incorporate the non-linear attention mechanism used in \cite{bello2019attention} in our model for a fair comparison. Due to high memory requirements, we were only able to use the non-linear attention module at the lowest spatial resolution. As it already occupied full GPU memory, we were unable to introduce more blocks or cross attention. The resultant PSNR was 30.52 compared to 30.76 of Net3. This experiment demonstrates that for image restoration tasks like deblurring, applying efficient linear attention on relatively higher resolutions and thus preserving the pixel information is far more beneficial than deploying more expressible non-linear operations at significantly downsampled feature maps.

Net9 was our final model with pixel-adaptive processing modules but with uniform sampling. Further, we train two more variants of our network: we add GT image supervision at the end of each stage so that the network learns to restore the image progressively (Net10). For Net11, we introduce non-uniform pixel-adaptive sampling to Net10. We can observe that, although the introduction of progressive restoration losses outperforms Net8, the improvement is relatively small. But, when coupled with the non-uniform sampling (Net11), we achieve a significant (0.73 dB) PSNR improvement over our original network Net8. We argue that selecting the difficult-to-restore locations and processing those over a much finer grid (instead of making redundant predictions uniformly over all points) enables us to fully exploit the superior processing power of the attention and adaptive-filtering module, leading to a significant improvement in performance. We also train a variant of our network, where we predict the cluster of spatial attention maps for context aggregation ($X^i_{u}$ in Eq. \ref{eq:spat_cluster}) from the full-sized feature map. It leads to 32.78 dB PSNR, which is similar to Net 11, but considerably increases the computation and inference time.
\newline In the next section, we will visualize and discuss the changes in the behavior of attention and PDF modules after including pixel-adaptive sampling.


\begin{figure}
   \centering
\begin{tabular}{ccc}
\hspace{-3mm}\includegraphics[width=0.15\textwidth, height = 0.09\textwidth]{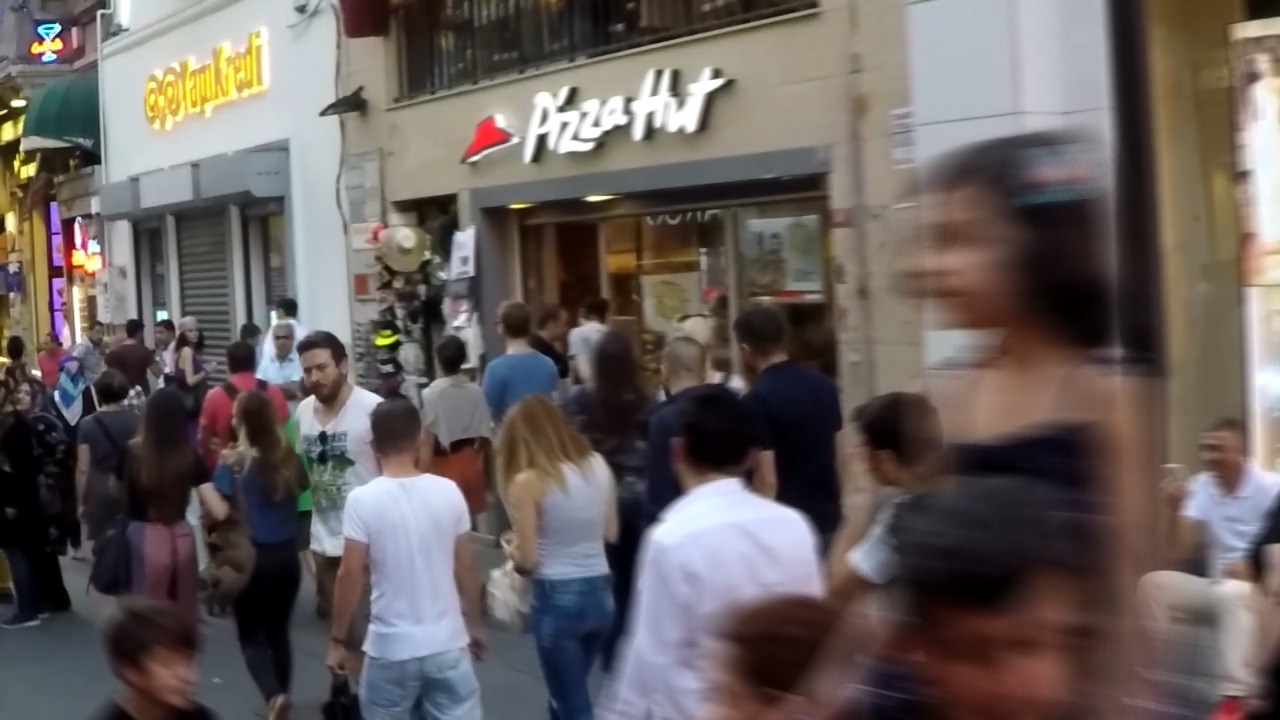}&
\hspace{-3mm}\includegraphics[width=0.15\textwidth, height = 0.09\textwidth]{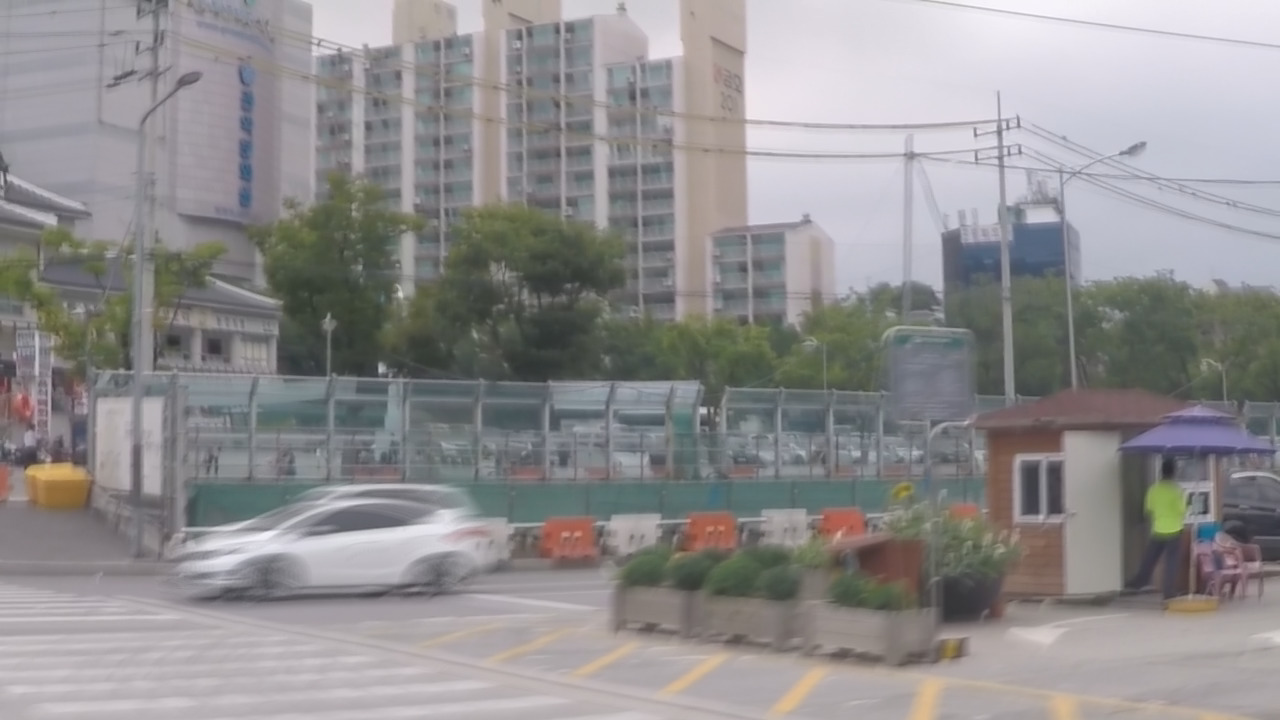}&
\hspace{-3mm}\includegraphics[width=0.15\textwidth, height = 0.09\textwidth]{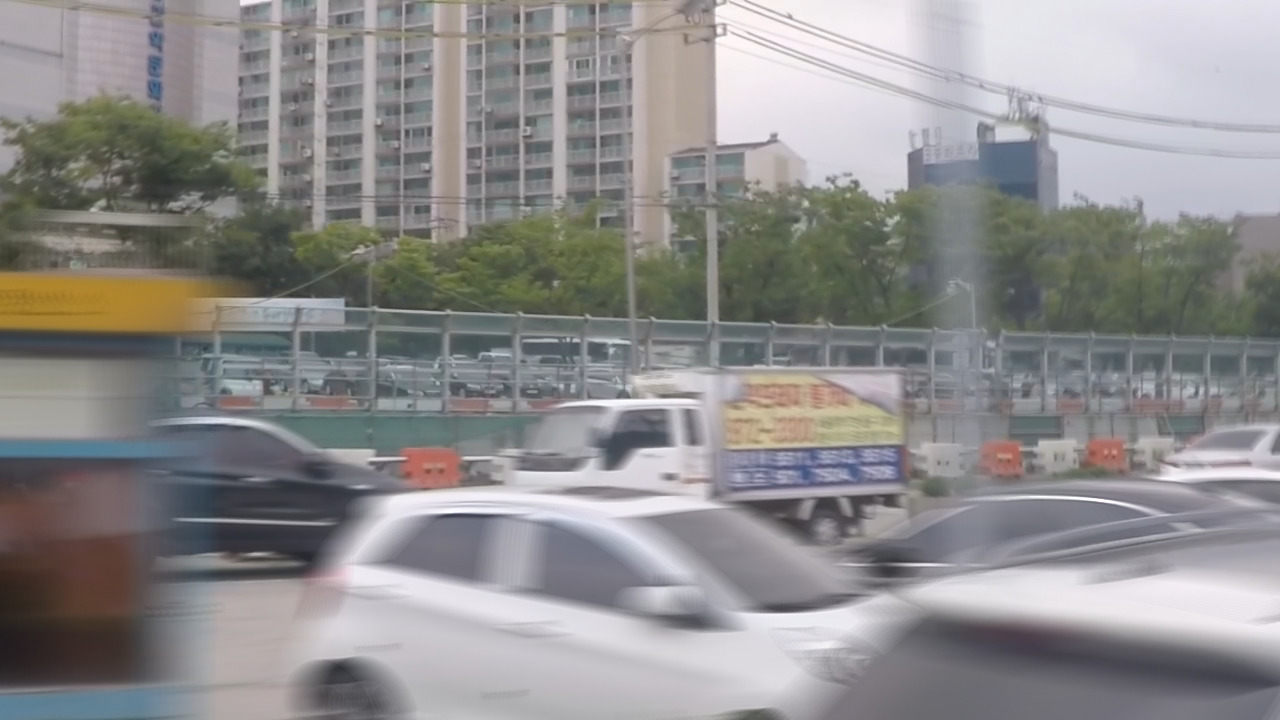}
\\
\hspace{-3mm}\includegraphics[width=0.15\textwidth, height = 0.09\textwidth]{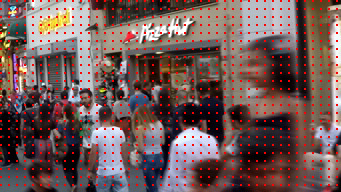}&
\hspace{-3mm}\includegraphics[width=0.15\textwidth, height = 0.09\textwidth]{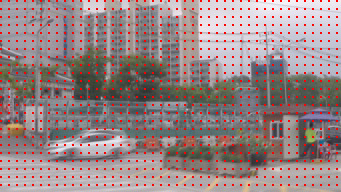}&
\hspace{-3mm}\includegraphics[width=0.15\textwidth, height = 0.09\textwidth]{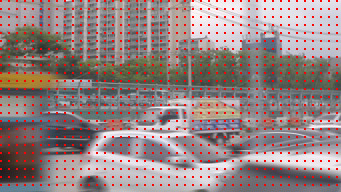}
\\
\hspace{-3mm}\includegraphics[width=0.15\textwidth, height = 0.09\textwidth]{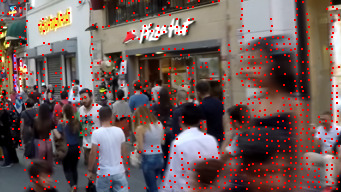}&
\hspace{-3mm}\includegraphics[width=0.15\textwidth, height = 0.09\textwidth]{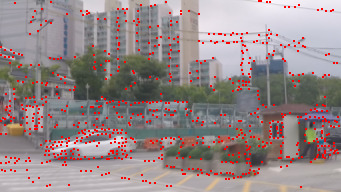}&
\hspace{-3mm}\includegraphics[width=0.15\textwidth, height = 0.09\textwidth]{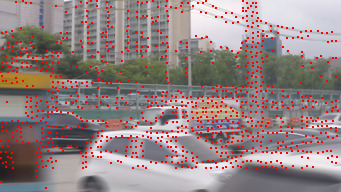}
\\
\hspace{-3mm}\includegraphics[width=0.15\textwidth, height = 0.09\textwidth]{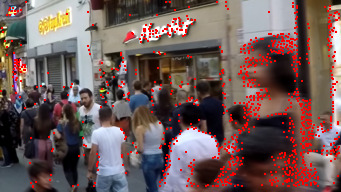}&
\hspace{-3mm}\includegraphics[width=0.15\textwidth, height = 0.09\textwidth]{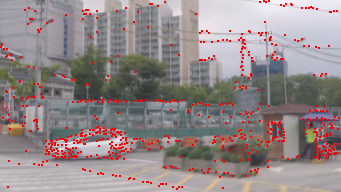}&
\hspace{-3mm}\includegraphics[width=0.15\textwidth, height = 0.09\textwidth]{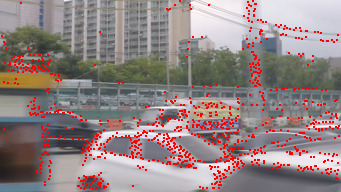}
\end{tabular}

    \caption{Visualization of sampling across stages. Row1 to Row4 represents: Input Image, Uniform Sampling (Stage-1), Non-uniform Sampling (Stage-2) and Non-uniform Sampling (Stage-3). Sampled pixels are shown in red. Best viewed zoomed-in.}
    \label{fig:sampling}
    \vspace{-5mm}
\end{figure}
\vspace{-3mm}
\subsection{Visualization and Analysis}
We have visualized the behavior of our adaptive processing modules in Fig. \ref{fig:visualization}. The first column of Fig. \ref{fig:visualization} contains images from the testing datasets, which suffer from complex blur due to a combination of camera and object motion. In the subsequent columns, we visualize the output of different modules of our network and analyze the behavior while handling different blur levels at different stages of the network. The second and third columns of Fig. \ref{fig:visualization} show one of the attention-maps ($q_i$, $i \in {1,2,...C_2}$) from stage-1 and stage-3. We can observe a high correlation between estimated attention weights and the dominant motion-blurred regions present in the image for both stages. But, the attention in stage-1 is more uniformly distributed over all the blurry areas. In comparison, the attention in stage-3 is more refined and focused on the severely degraded areas. Thus, the adaptive sampling allows the attention module to `zoom' into the challenging regions and process those pixels more accurately in the subsequent stages and can be considered crucial to the observed performance improvement. The fourth, fifth, and sixth columns of Fig. \ref{fig:visualization} show the spatially-varying nature of filter weights and offsets. Observe that a large offset is estimated in the regions with high blur so that the filter shape can spread along the direction of motion. After introducing adaptive sampling, we can observe that for stage-1, the offset variation is relatively more uniform over a large part of the blurry image, whereas, for stage-3, it is more finely distributed over the high-blur areas. Note that, for better visualization of $V$ and $\Delta$, we have transformed the non-uniform pixel grid to a uniform grid and applied interpolation wherever required. We have shown the estimated filter weights in the sixth column. Although the estimated filter wights are not directly interpretable, it can be seen that the variance of the filters correlates with the magnitude of blur.

\par Next, we analyze the behavior of the sampling process. The pixels are sampled uniformly in the first stage. In the second and third stages, we sample adaptively while keeping the computation, i.e., the total number of pixels sampled the same as the uniform case. In Fig. \ref{fig:sampling}, we have visualized the sampled pixels across stages. For better visualization of the non-uniformity of the sampled pixels, we showed the sampled pixels for the lowest spatial resolution. For the first stage, the pixels are uniformly spread over the full spatial scale. For the second and the third stage,  we can observe that the distribution of the sampled pixels directly correlates with the level of blur in the input image. For example, in the second column, we can visualize that pixels are being densely sampled from the boundary of the moving car or edges of the buildings. Not only the edges, which are highly affected by blur in general, our approach is able to detect blurry pixels from other regions as well. For example, in the first column, there are many blurry pixels sampled from the facial region of the two persons in the foreground. In the third column, we can observe many pixels sampled from the high-frequency region of the tree and the shop structure (on the extreme left). On the other hand, homogeneous regions such as the sky are skipped in the final stages. This behavior supports our intuition that, in most cases, a progressively trained network struggles for a handful of regions in the later stages, and the overall computation should be re-distributed to allow a finer refinement of the difficult regions. We also experimented with increasing or decreasing the number of non-uniformly sampled pixels (compared to uniform sampling), to observe its effect on the accuracy in Table \ref{table:sampling}. Although the accuracy improves by increasing the total number of pixels, the complexity also rises sharply. Moreover, processing more pixels often becomes infeasible due to the memory limitation of standard GPUs. Note that the sampling operation is considerably lightweight and increases the inference time by only 0.02s (NU vs. U in Table \ref{table:sampling}). But, increasing the number of sampled pixels significantly increases the runtime as more pixels are processed at every stage and level of the network (equivalent to increasing the spatial resolution of the feature maps). On the other hand, decreasing the number of total pixels hurt the overall accuracy. This strategy can potentially improve other methods that solely focus on efficiency but is beyond the scope of the current work.
\newline Detailed experimental results and layer-wise details are provided in the supplementary material.

\begin{figure}
   \centering
   \resizebox{.5\textwidth}{!}{
\begin{tabular}{cccc}
\hspace{-3mm}\includegraphics[width=0.15\textwidth]{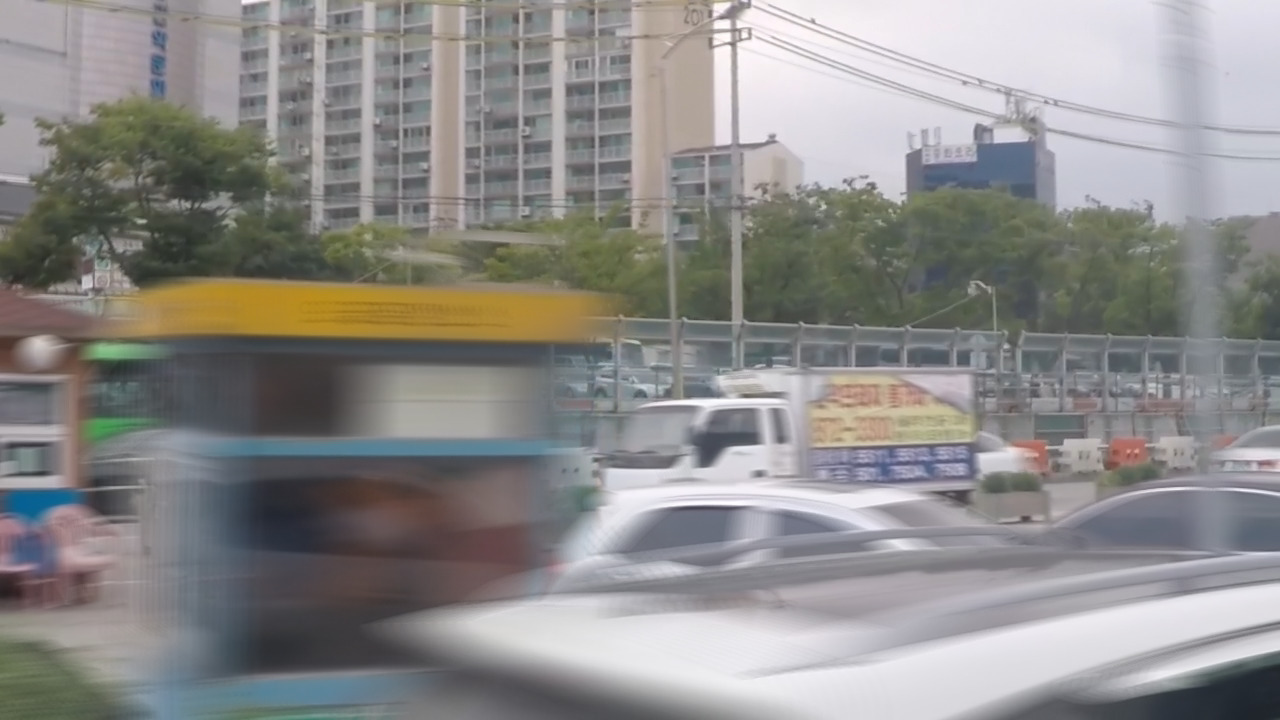}&
\hspace{-3mm}\includegraphics[bb=100 150 630 450,clip=True,width=0.15\textwidth]{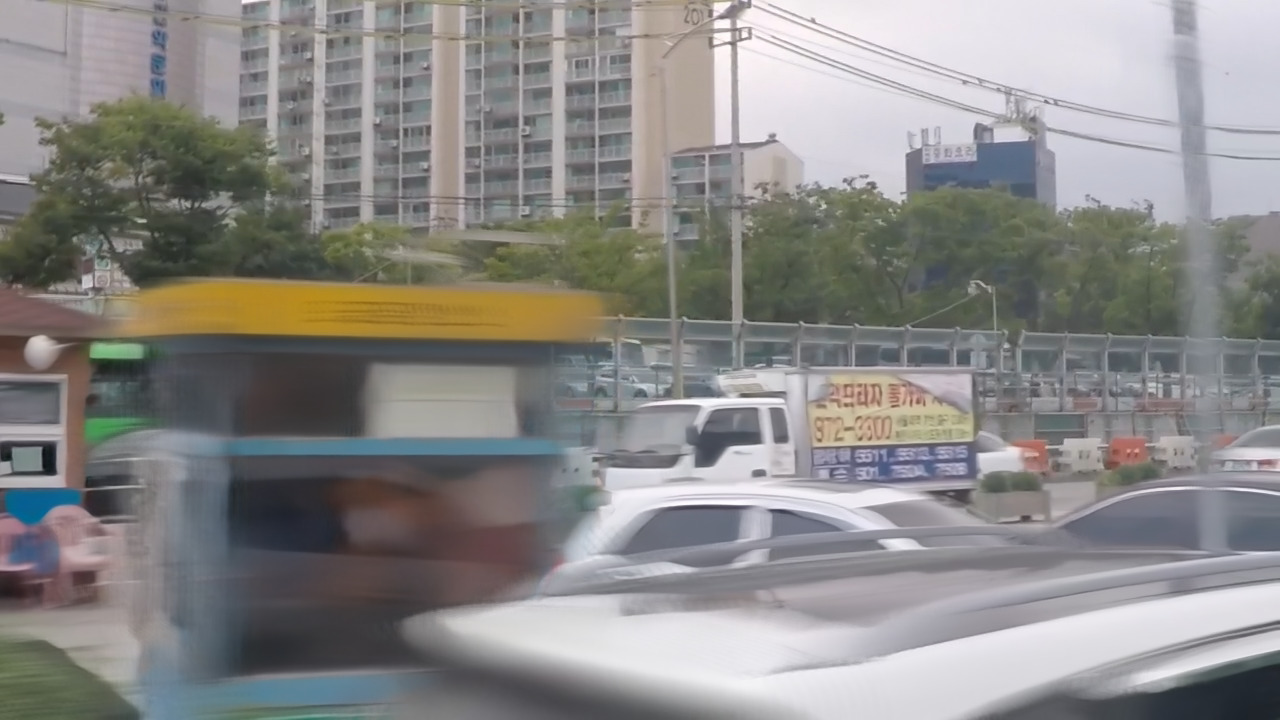}&
\hspace{-3mm}\includegraphics[bb=100 150 630 450,clip=True,width=0.15\textwidth]{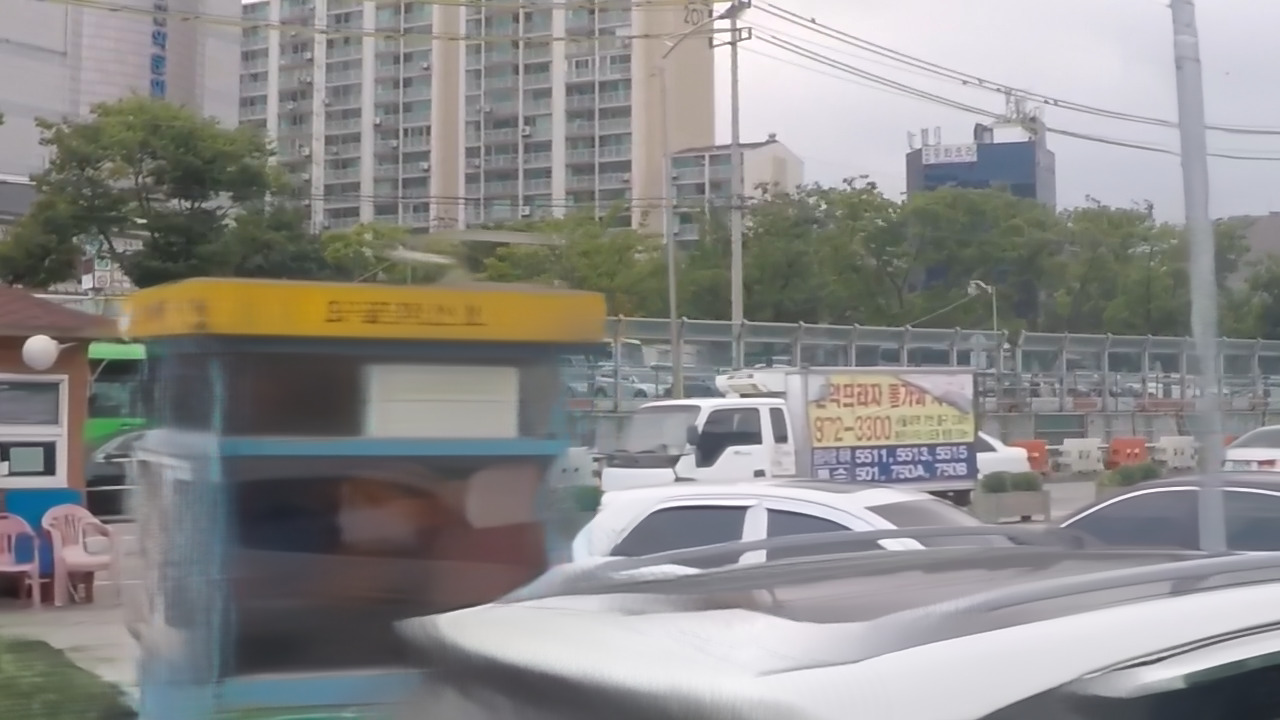}&
\hspace{-3mm}\includegraphics[bb=100 150 630 450,clip=True,width=0.15\textwidth]{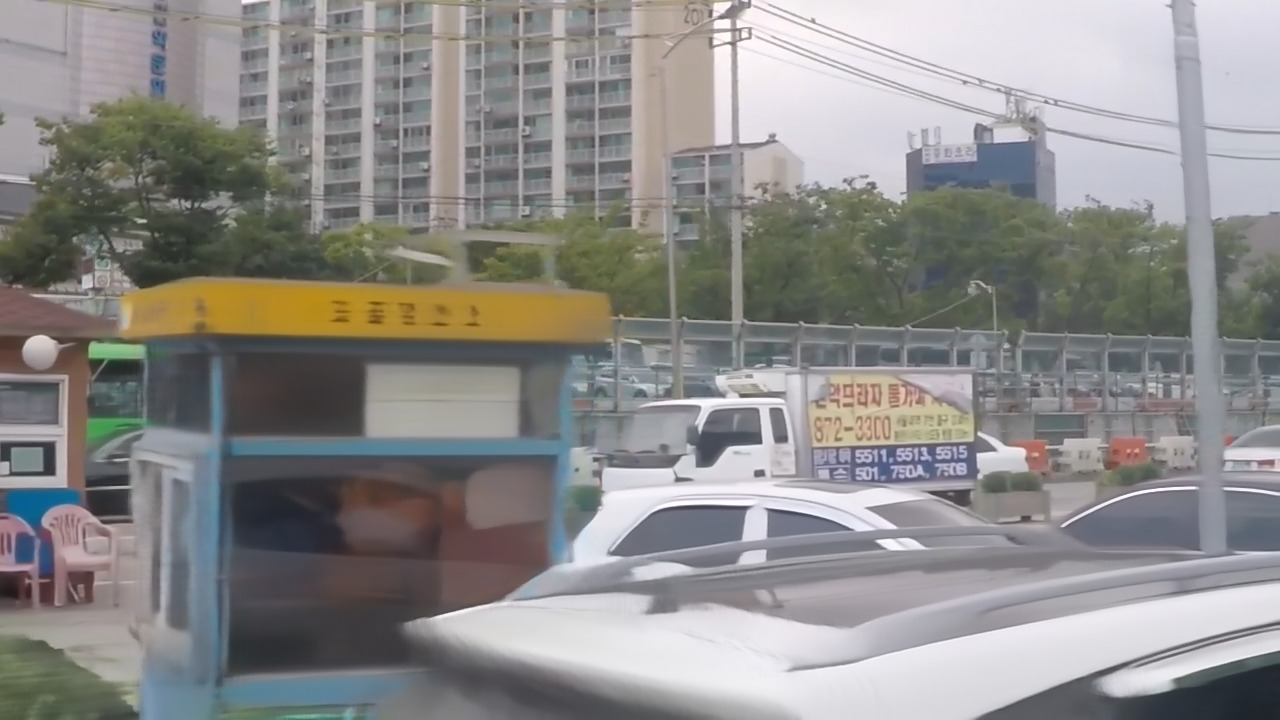} \\
\hspace{-3mm}\includegraphics[width=0.15\textwidth]{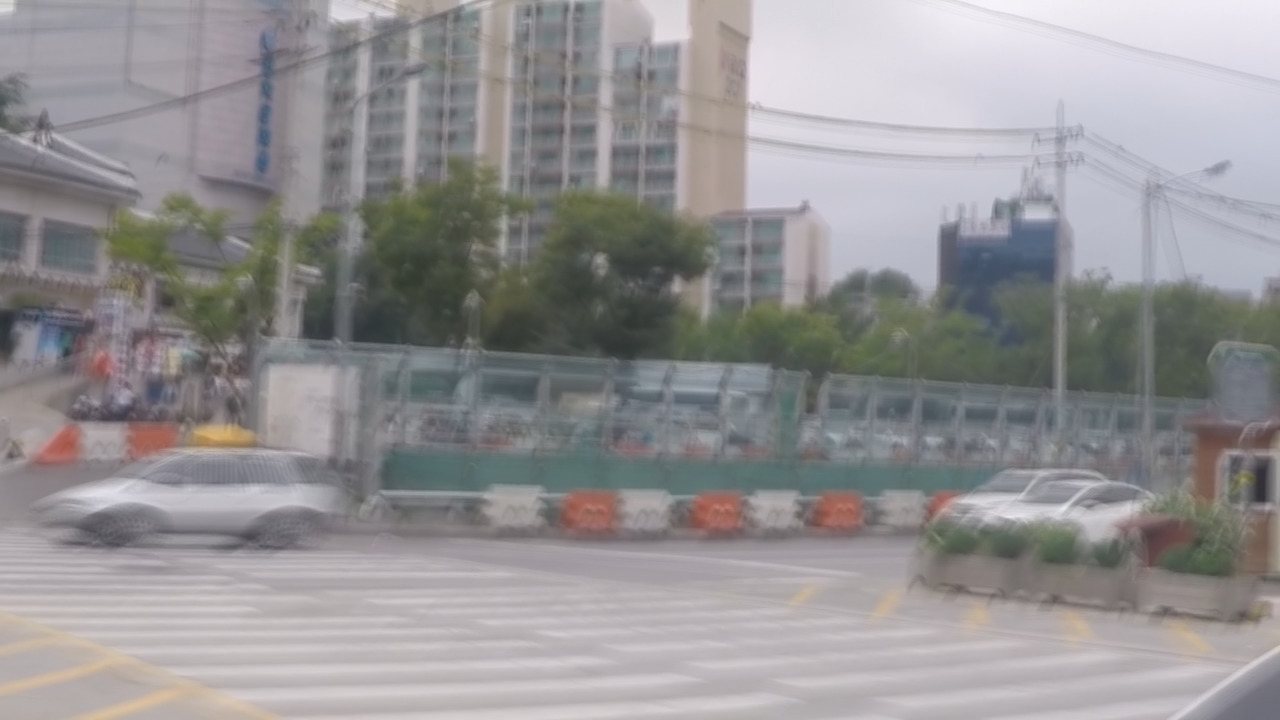}&
\hspace{-3mm}\includegraphics[bb=50 150 400 350,clip=True,width=0.15\textwidth]{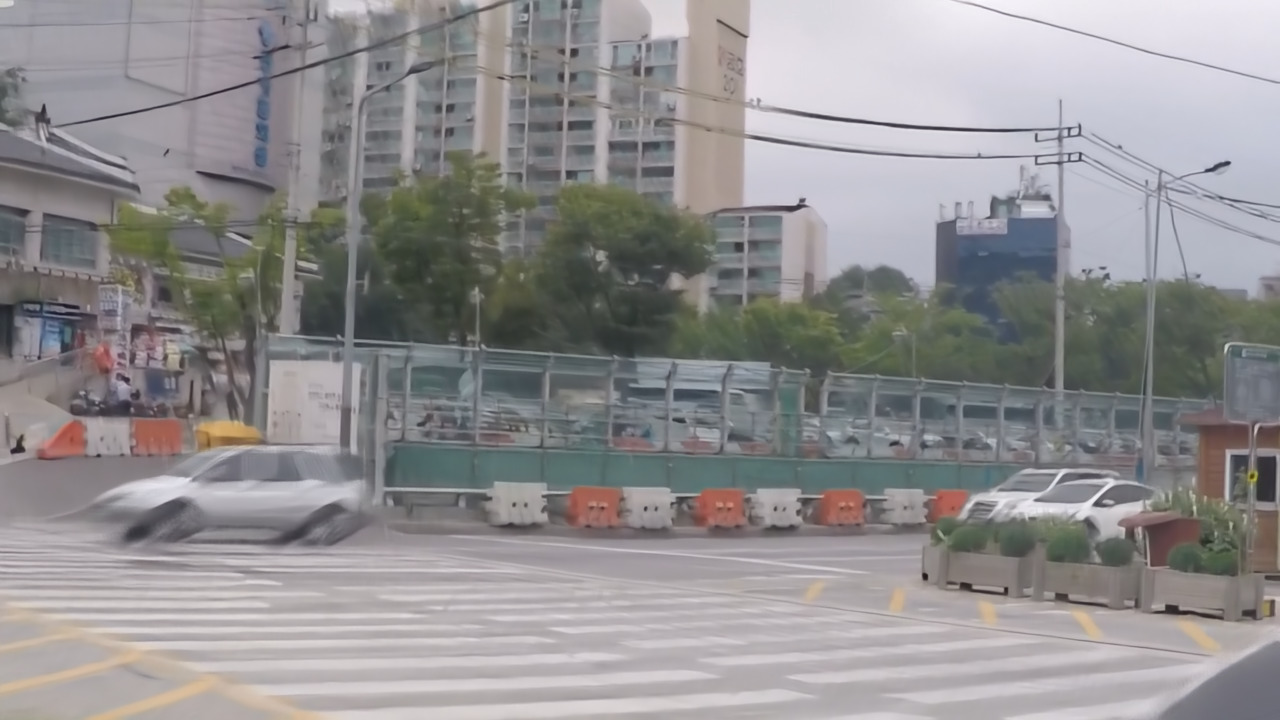}&
\hspace{-3mm}\includegraphics[bb=50 150 400 350,clip=True,width=0.15\textwidth]{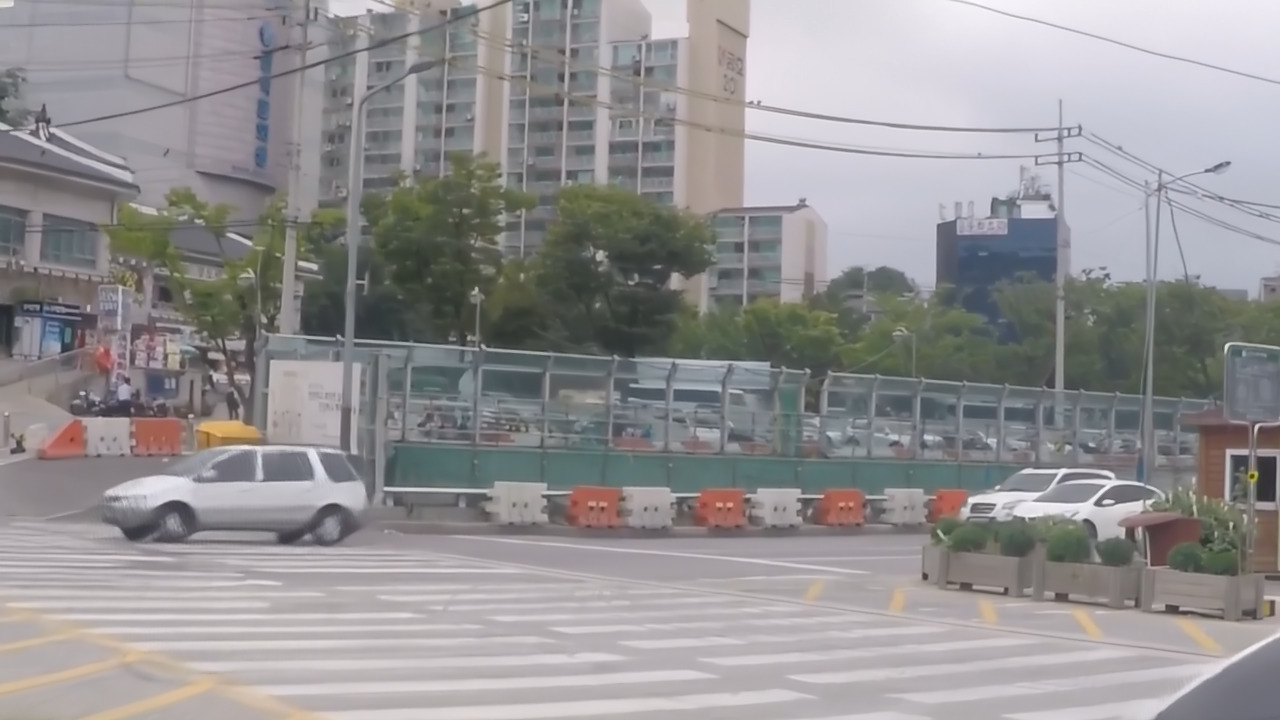}&
\hspace{-3mm}\includegraphics[bb=50 150 400 350,clip=True,width=0.15\textwidth]{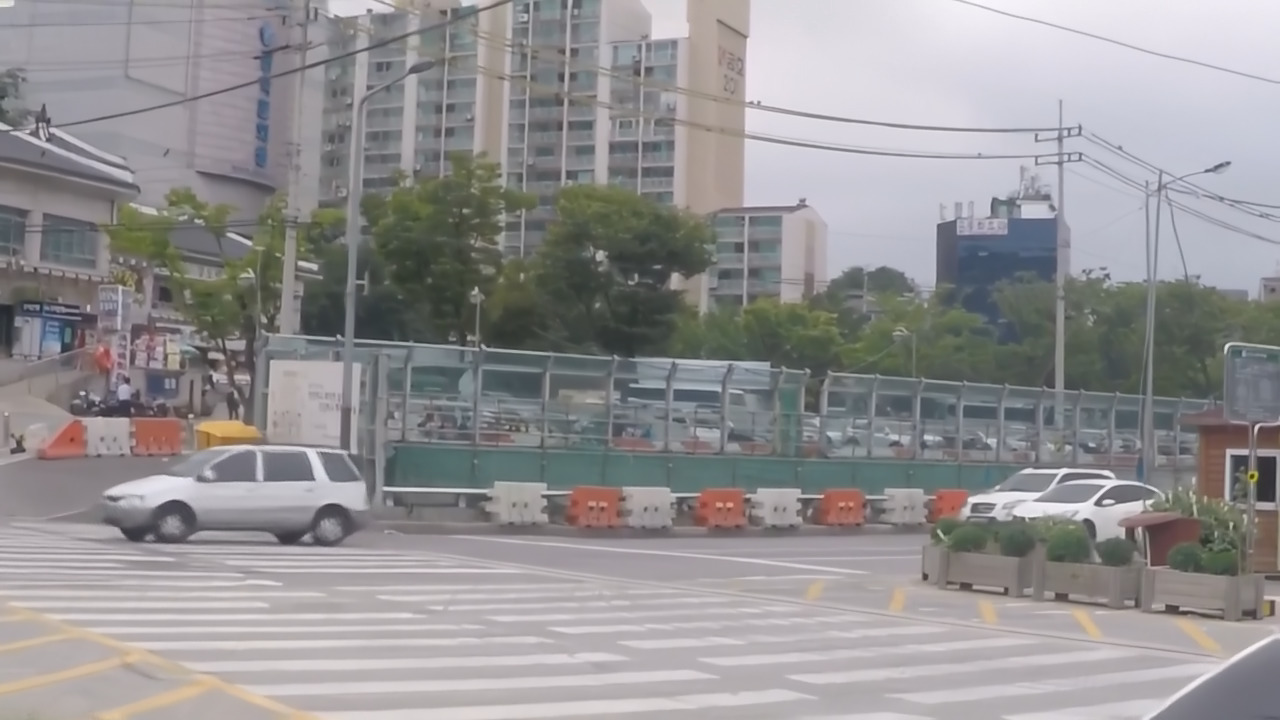} \\
Input & Stage-1 Output & Stage-2 Output & Stage-3 Output
\end{tabular}}

    \caption{Visualization of deblurring results across stages in the progressive refinement framework.}
    \label{fig:progr}
    \vspace{-4mm}
\end{figure}

\vspace{-5mm}
\section{Conclusion}
\label{sec:conc}
We proposed a new content-adaptive architecture design for the challenging task of removing spatially-varying blur in images of dynamic scenes. We first introduce a pixel-adaptive non-uniform sampling strategy to densely sample the heavily blurred pixels that require finer refinement. Efficient self-attention is utilized in all the encoder-decoder to get better representation, whereas cross-attention helps in efficient feature propagation across layers and levels. The proposed dynamic filtering module shows content awareness for local filtering. Unlike existing deep learning-based methods for such applications, the proposed method is more interpretable, which is one of its key strengths. Our experimental results demonstrated that the proposed method achieved better results than state-of-the-art methods with comparable efficiency in most cases. The proposed content-adaptive approach can potentially be applied to other image-processing tasks in the future. 
\ifCLASSOPTIONcaptionsoff
  \newpage
\fi



%
{
\bibliographystyle{IEEEtran}
\bibliography{egbib}

\begin{thebibliography}{10}
\providecommand{\url}[1]{#1}
\csname url@samestyle\endcsname
\providecommand{\newblock}{\relax}
\providecommand{\bibinfo}[2]{#2}
\providecommand{\BIBentrySTDinterwordspacing}{\spaceskip=0pt\relax}
\providecommand{\BIBentryALTinterwordstretchfactor}{4}
\providecommand{\BIBentryALTinterwordspacing}{\spaceskip=\fontdimen2\font plus
\BIBentryALTinterwordstretchfactor\fontdimen3\font minus \fontdimen4\font\relax}
\providecommand{\BIBforeignlanguage}[2]{{%
\expandafter\ifx\csname l@#1\endcsname\relax
\typeout{** WARNING: IEEEtran.bst: No hyphenation pattern has been}%
\typeout{** loaded for the language `#1'. Using the pattern for}%
\typeout{** the default language instead.}%
\else
\language=\csname l@#1\endcsname
\fi
#2}}
\providecommand{\BIBdecl}{\relax}
\BIBdecl

\bibitem{nimisha2017blur}
T.~Nimisha, A.~K. Singh, and A.~Rajagopalan, ``Blur-invariant deep learning for blind-deblurring,'' in \emph{Proceedings of the IEEE E International Conference on Computer Vision (ICCV)}, 2017.

\bibitem{gong2017motion}
D.~Gong, J.~Yang, L.~Liu, Y.~Zhang, I.~Reid, C.~Shen, A.~Hengel, and Q.~Shi, ``From motion blur to motion flow: a deep learning solution for removing heterogeneous motion blur,'' in \emph{The IEEE conference on computer vision and pattern recognition (CVPR)}, 2017.

\bibitem{nah2017deep}
S.~Nah, T.~H. Kim, and K.~M. Lee, ``Deep multi-scale convolutional neural network for dynamic scene deblurring,'' in \emph{CVPR}, vol.~1, no.~2, 2017, p.~3.

\bibitem{suin2020spatially}
M.~Suin, K.~Purohit, and A.~Rajagopalan, ``Spatially-attentive patch-hierarchical network for adaptive motion deblurring,'' in \emph{Proceedings of the IEEE/CVF Conference on Computer Vision and Pattern Recognition}, 2020, pp. 3606--3615.

\bibitem{Zamir_2021_CVPR_mprnet}
S.~W. Zamir, A.~Arora, S.~Khan, M.~Hayat, F.~S. Khan, M.-H. Yang, and L.~Shao, ``Multi-stage progressive image restoration,'' in \emph{CVPR}, 2021.

\bibitem{rim_2020_realblur}
J.~Rim, H.~Lee, J.~Won, and S.~Cho, ``Real-world blur dataset for learning and benchmarking deblurring algorithms,'' in \emph{ECCV}, 2020.

\bibitem{fergus2006removing}
R.~Fergus, B.~Singh, A.~Hertzmann, S.~T. Roweis, and W.~T. Freeman, ``Removing camera shake from a single photograph,'' in \emph{ACM transactions on graphics (TOG)}, vol.~25, no.~3.\hskip 1em plus 0.5em minus 0.4em\relax ACM, 2006, pp. 787--794.

\bibitem{cho2009fast}
S.~Cho and S.~Lee, ``Fast motion deblurring,'' in \emph{ACM Transactions on Graphics (TOG)}, vol.~28, no.~5.\hskip 1em plus 0.5em minus 0.4em\relax ACM, 2009, p. 145.

\bibitem{gong2016blind}
D.~Gong, M.~Tan, Y.~Zhang, A.~Van~den Hengel, and Q.~Shi, ``Blind image deconvolution by automatic gradient activation,'' in \emph{Proceedings of the IEEE Conference on Computer Vision and Pattern Recognition}, 2016, pp. 1827--1836.

\bibitem{xu2010two}
L.~Xu and J.~Jia, ``Two-phase kernel estimation for robust motion deblurring,'' in \emph{European Conference on Computer Vision}.\hskip 1em plus 0.5em minus 0.4em\relax Springer, 2010, pp. 157--170.

\bibitem{hu2012good}
Z.~Hu and M.-H. Yang, ``Good regions to deblur,'' in \emph{European conference on computer vision}.\hskip 1em plus 0.5em minus 0.4em\relax Springer, 2012, pp. 59--72.

\bibitem{hyun2013dynamic}
T.~Hyun~Kim, B.~Ahn, and K.~Mu~Lee, ``Dynamic scene deblurring,'' in \emph{Proceedings of the IEEE International Conference on Computer Vision}, 2013, pp. 3160--3167.

\bibitem{hyun2014segmentation}
T.~Hyun~Kim and K.~Mu~Lee, ``Segmentation-free dynamic scene deblurring,'' in \emph{Proceedings of the IEEE Conference on Computer Vision and Pattern Recognition}, 2014, pp. 2766--2773.

\bibitem{ren2016image}
W.~Ren, X.~Cao, J.~Pan, X.~Guo, W.~Zuo, and M.-H. Yang, ``Image deblurring via enhanced low-rank prior,'' \emph{IEEE Transactions on Image Processing}, vol.~25, no.~7, pp. 3426--3437, 2016.

\bibitem{pan2014motion}
J.~Pan, R.~Liu, Z.~Su, and G.~Liu, ``Motion blur kernel estimation via salient edges and low rank prior,'' in \emph{2014 IEEE International Conference on Multimedia and Expo (ICME)}.\hskip 1em plus 0.5em minus 0.4em\relax IEEE, 2014, pp. 1--6.

\bibitem{pan2016blind}
J.~Pan, D.~Sun, H.~Pfister, and M.-H. Yang, ``Blind image deblurring using dark channel prior,'' in \emph{Proceedings of the IEEE Conference on Computer Vision and Pattern Recognition}, 2016, pp. 1628--1636.

\bibitem{yan2017image}
Y.~Yan, W.~Ren, Y.~Guo, R.~Wang, and X.~Cao, ``Image deblurring via extreme channels prior,'' in \emph{Proceedings of the IEEE Conference on Computer Vision and Pattern Recognition}, 2017, pp. 4003--4011.

\bibitem{li2018learning}
L.~Li, J.~Pan, W.-S. Lai, C.~Gao, N.~Sang, and M.-H. Yang, ``Learning a discriminative prior for blind image deblurring,'' in \emph{Proceedings of the IEEE Conference on Computer Vision and Pattern Recognition}, 2018, pp. 6616--6625.

\bibitem{lai2015blur}
W.-S. Lai, J.-J. Ding, Y.-Y. Lin, and Y.-Y. Chuang, ``Blur kernel estimation using normalized color-line prior,'' in \emph{Proceedings of the IEEE Conference on Computer Vision and Pattern Recognition}, 2015, pp. 64--72.

\bibitem{sun2015learning}
J.~Sun, W.~Cao, Z.~Xu, and J.~Ponce, ``Learning a convolutional neural network for non-uniform motion blur removal,'' in \emph{Proceedings of the IEEE Conference on Computer Vision and Pattern Recognition}, 2015, pp. 769--777.

\bibitem{xu2017learning}
X.~Xu, D.~Sun, J.~Pan, Y.~Zhang, H.~Pfister, and M.-H. Yang, ``Learning to super-resolve blurry face and text images,'' in \emph{Proceedings of the IEEE international conference on computer vision}, 2017, pp. 251--260.

\bibitem{hradivs2015convolutional}
M.~Hradi{\v{s}}, J.~Kotera, P.~Zemc{\'\i}k, and F.~{\v{S}}roubek, ``Convolutional neural networks for direct text deblurring,'' in \emph{Proceedings of BMVC}, vol.~10, 2015, p.~2.

\bibitem{svoboda2016cnn}
P.~Svoboda, M.~Hradi{\v{s}}, L.~Mar{\v{s}}{\'\i}k, and P.~Zemc{\'\i}k, ``Cnn for license plate motion deblurring,'' in \emph{2016 IEEE International Conference on Image Processing (ICIP)}.\hskip 1em plus 0.5em minus 0.4em\relax IEEE, 2016, pp. 3832--3836.

\bibitem{kupyn2018deblurgan}
O.~Kupyn, V.~Budzan, M.~Mykhailych, D.~Mishkin, and J.~Matas, ``Deblurgan: Blind motion deblurring using conditional adversarial networks,'' in \emph{Proceedings of the IEEE conference on computer vision and pattern recognition}, 2018, pp. 8183--8192.

\bibitem{gao2019dynamic}
H.~Gao, X.~Tao, X.~Shen, and J.~Jia, ``Dynamic scene deblurring with parameter selective sharing and nested skip connections,'' in \emph{Proceedings of the IEEE Conference on Computer Vision and Pattern Recognition}, 2019, pp. 3848--3856.

\bibitem{tao2018scale}
X.~Tao, H.~Gao, X.~Shen, J.~Wang, and J.~Jia, ``Scale-recurrent network for deep image deblurring,'' in \emph{Proceedings of the IEEE Conference on Computer Vision and Pattern Recognition}, 2018, pp. 8174--8182.

\bibitem{cho2021rethinking}
S.-J. Cho, S.-W. Ji, J.-P. Hong, S.-W. Jung, and S.-J. Ko, ``Rethinking coarse-to-fine approach in single image deblurring,'' in \emph{Proceedings of the IEEE/CVF International Conference on Computer Vision}, 2021, pp. 4641--4650.

\bibitem{suin2024diffuse}
M.~Suin, N.~G. Nair, C.~P. Lau, V.~M. Patel, and R.~Chellappa, ``Diffuse and restore: A region-adaptive diffusion model for identity-preserving blind face restoration,'' in \emph{Proceedings of the IEEE/CVF Winter Conference on Applications of Computer Vision}, 2024, pp. 6343--6352.

\bibitem{purohit2021spatially}
K.~Purohit, M.~Suin, A.~Rajagopalan, and V.~N. Boddeti, ``Spatially-adaptive image restoration using distortion-guided networks,'' in \emph{Proceedings of the IEEE/CVF International Conference on Computer Vision}, 2021, pp. 2309--2319.

\bibitem{zhang2019deep}
H.~Zhang, Y.~Dai, H.~Li, and P.~Koniusz, ``Deep stacked hierarchical multi-patch network for image deblurring,'' in \emph{Proceedings of the IEEE Conference on Computer Vision and Pattern Recognition}, 2019, pp. 5978--5986.

\bibitem{zhang2018dynamic}
J.~Zhang, J.~Pan, J.~Ren, Y.~Song, L.~Bao, R.~W. Lau, and M.-H. Yang, ``Dynamic scene deblurring using spatially variant recurrent neural networks,'' in \emph{Proceedings of the IEEE Conference on Computer Vision and Pattern Recognition}, 2018, pp. 2521--2529.

\bibitem{xu2020unified}
Y.-S. Xu, S.-Y.~R. Tseng, Y.~Tseng, H.-K. Kuo, and Y.-M. Tsai, ``Unified dynamic convolutional network for super-resolution with variational degradations,'' in \emph{Proceedings of the IEEE/CVF Conference on Computer Vision and Pattern Recognition}, 2020, pp. 12\,496--12\,505.

\bibitem{niu2020single}
B.~Niu, W.~Wen, W.~Ren, X.~Zhang, L.~Yang, S.~Wang, K.~Zhang, X.~Cao, and H.~Shen, ``Single image super-resolution via a holistic attention network,'' in \emph{European conference on computer vision}.\hskip 1em plus 0.5em minus 0.4em\relax Springer, 2020, pp. 191--207.

\bibitem{yi2019progressive}
P.~Yi, Z.~Wang, K.~Jiang, J.~Jiang, and J.~Ma, ``Progressive fusion video super-resolution network via exploiting non-local spatio-temporal correlations,'' in \emph{Proceedings of the IEEE/CVF international conference on computer vision}, 2019, pp. 3106--3115.

\bibitem{luo2016understanding}
W.~Luo, Y.~Li, R.~Urtasun, and R.~Zemel, ``Understanding the effective receptive field in deep convolutional neural networks,'' in \emph{Advances in neural information processing systems}, 2016, pp. 4898--4906.

\bibitem{sutton1999policy}
R.~S. Sutton, D.~McAllester, S.~Singh, and Y.~Mansour, ``Policy gradient methods for reinforcement learning with function approximation,'' \emph{Advances in neural information processing systems}, vol.~12, 1999.

\bibitem{sutton2018reinforcement}
R.~S. Sutton and A.~G. Barto, \emph{Reinforcement learning: An introduction}.\hskip 1em plus 0.5em minus 0.4em\relax MIT press, 2018.

\bibitem{foerster2017stabilising}
J.~Foerster, N.~Nardelli, G.~Farquhar, T.~Afouras, P.~H. Torr, P.~Kohli, and S.~Whiteson, ``Stabilising experience replay for deep multi-agent reinforcement learning,'' in \emph{International conference on machine learning}.\hskip 1em plus 0.5em minus 0.4em\relax PMLR, 2017, pp. 1146--1155.

\bibitem{foerster2016learning}
J.~Foerster, I.~A. Assael, N.~De~Freitas, and S.~Whiteson, ``Learning to communicate with deep multi-agent reinforcement learning,'' \emph{Advances in neural information processing systems}, vol.~29, 2016.

\bibitem{wang2018non}
X.~Wang, R.~Girshick, A.~Gupta, and K.~He, ``Non-local neural networks,'' in \emph{Proceedings of the IEEE Conference on Computer Vision and Pattern Recognition}, 2018, pp. 7794--7803.

\bibitem{bello2019attention}
I.~Bello, B.~Zoph, A.~Vaswani, J.~Shlens, and Q.~V. Le, ``Attention augmented convolutional networks,'' \emph{arXiv preprint arXiv:1904.09925}, 2019.

\bibitem{vaswani2017attention}
A.~Vaswani, N.~Shazeer, N.~Parmar, J.~Uszkoreit, L.~Jones, A.~N. Gomez, {\L}.~Kaiser, and I.~Polosukhin, ``Attention is all you need,'' in \emph{Advances in Neural Information Processing Systems}, 2017, pp. 5998--6008.

\bibitem{parmar2018image}
N.~Parmar, A.~Vaswani, J.~Uszkoreit, {\L}.~Kaiser, N.~Shazeer, A.~Ku, and D.~Tran, ``Image transformer,'' \emph{arXiv preprint arXiv:1802.05751}, 2018.

\bibitem{liu2018non}
D.~Liu, B.~Wen, Y.~Fan, C.~C. Loy, and T.~S. Huang, ``Non-local recurrent network for image restoration,'' in \emph{Advances in Neural Information Processing Systems}, 2018, pp. 1673--1682.

\bibitem{ramachandran2019stand}
P.~Ramachandran, N.~Parmar, A.~Vaswani, I.~Bello, A.~Levskaya, and J.~Shlens, ``Stand-alone self-attention in vision models,'' \emph{arXiv preprint arXiv:1906.05909}, 2019.

\bibitem{jia2016dynamic}
X.~Jia, B.~De~Brabandere, T.~Tuytelaars, and L.~V. Gool, ``Dynamic filter networks,'' in \emph{Advances in Neural Information Processing Systems}, 2016, pp. 667--675.

\bibitem{li2018video}
Y.~Li, M.~R. Min, D.~Shen, D.~Carlson, and L.~Carin, ``Video generation from text,'' in \emph{Thirty-Second AAAI Conference on Artificial Intelligence}, 2018.

\bibitem{zhang2018crowd}
L.~Zhang, M.~Shi, and Q.~Chen, ``Crowd counting via scale-adaptive convolutional neural network,'' in \emph{2018 IEEE Winter Conference on Applications of Computer Vision (WACV)}.\hskip 1em plus 0.5em minus 0.4em\relax IEEE, 2018, pp. 1113--1121.

\bibitem{su2019pixel}
H.~Su, V.~Jampani, D.~Sun, O.~Gallo, E.~Learned-Miller, and J.~Kautz, ``Pixel-adaptive convolutional neural networks,'' in \emph{Proceedings of the IEEE Conference on Computer Vision and Pattern Recognition}, 2019, pp. 11\,166--11\,175.

\bibitem{kupyn2019deblurgan}
O.~Kupyn, T.~Martyniuk, J.~Wu, and Z.~Wang, ``Deblurgan-v2: Deblurring (orders-of-magnitude) faster and better,'' in \emph{Proceedings of the IEEE International Conference on Computer Vision}, 2019, pp. 8878--8887.

\bibitem{deblurgan}
O.~Kupyn, V.~Budzan, M.~Mykhailych, D.~Mishkin, and J.~Matas, ``{DeblurGAN}: Blind motion deblurring using conditional adversarial networks,'' in \emph{CVPR}, 2018.

\bibitem{gopro2017}
S.~Nah, T.~Hyun~Kim, and K.~Mu~Lee, ``Deep multi-scale convolutional neural network for dynamic scene deblurring,'' in \emph{CVPR}, 2017.

\bibitem{shen2019human}
Z.~Shen, W.~Wang, X.~Lu, J.~Shen, H.~Ling, T.~Xu, and L.~Shao, ``Human-aware motion deblurring,'' in \emph{Proceedings of the IEEE International Conference on Computer Vision}, 2019, pp. 5572--5581.

\bibitem{kingma2014adam}
D.~P. Kingma and J.~Ba, ``Adam: A method for stochastic optimization,'' \emph{arXiv preprint arXiv:1412.6980}, 2014.

\bibitem{paszke2017automatic}
A.~Paszke, S.~Gross, S.~Chintala, G.~Chanan, E.~Yang, Z.~DeVito, Z.~Lin, A.~Desmaison, L.~Antiga, and A.~Lerer, ``Automatic differentiation in pytorch,'' 2017.

\bibitem{xu2013unnatural}
L.~Xu, S.~Zheng, and J.~Jia, ``Unnatural l0 sparse representation for natural image deblurring,'' in \emph{Proceedings of the IEEE conference on computer vision and pattern recognition}, 2013, pp. 1107--1114.

\bibitem{whyte2012non}
O.~Whyte, J.~Sivic, A.~Zisserman, and J.~Ponce, ``Non-uniform deblurring for shaken images,'' \emph{International journal of computer vision}, vol.~98, no.~2, pp. 168--186, 2012.

\bibitem{zamir2022restormer}
S.~W. Zamir, A.~Arora, S.~Khan, M.~Hayat, F.~S. Khan, and M.-H. Yang, ``Restormer: Efficient transformer for high-resolution image restoration,'' in \emph{Proceedings of the IEEE/CVF Conference on Computer Vision and Pattern Recognition}, 2022, pp. 5728--5739.

\bibitem{wang2022uformer}
Z.~Wang, X.~Cun, J.~Bao, W.~Zhou, J.~Liu, and H.~Li, ``Uformer: A general u-shaped transformer for image restoration,'' in \emph{Proceedings of the IEEE/CVF Conference on Computer Vision and Pattern Recognition}, 2022, pp. 17\,683--17\,693.

\bibitem{deblurganv2}
O.~Kupyn, T.~Martyniuk, J.~Wu, and Z.~Wang, ``{DeblurGAN-v2}: Deblurring (orders-of-magnitude) faster and better,'' in \emph{ICCV}, 2019.

\bibitem{dmphn2019}
H.~Zhang, Y.~Dai, H.~Li, and P.~Koniusz, ``Deep stacked hierarchical multi-patch network for image deblurring,'' in \emph{CVPR}, 2019.

\bibitem{jiang2021enlightengan}
Y.~Jiang, X.~Gong, D.~Liu, Y.~Cheng, C.~Fang, X.~Shen, J.~Yang, P.~Zhou, and Z.~Wang, ``Enlightengan: Deep light enhancement without paired supervision,'' \emph{IEEE Transactions on Image Processing}, vol.~30, pp. 2340--2349, 2021.

\bibitem{katharopoulos2020transformers}
A.~Katharopoulos, A.~Vyas, N.~Pappas, and F.~Fleuret, ``Transformers are rnns: Fast autoregressive transformers with linear attention,'' in \emph{International Conference on Machine Learning}.\hskip 1em plus 0.5em minus 0.4em\relax PMLR, 2020, pp. 5156--5165.

\end{thebibliography}
}




%
\begin{IEEEbiography}[{\includegraphics[width=1in,height=1.25in,clip,keepaspectratio]{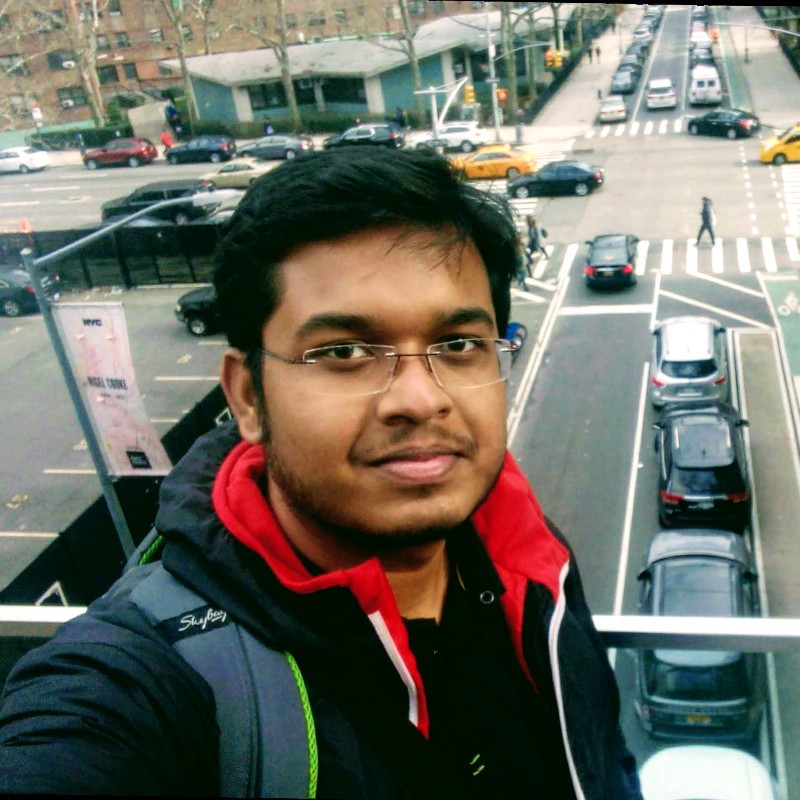}}]{Maitreya Suin}
Maitreya Suin is a postdoctoral fellow at JHU, USA. Earlier, he completed his MS and PhD from IIT Madras. His research interests lie at the intersection of Computer Vision, Deep Learning, and Deep Reinforcement Learning. 
\end{IEEEbiography}
\begin{IEEEbiography}[{\includegraphics[width=1in,height=1.25in,clip,keepaspectratio]{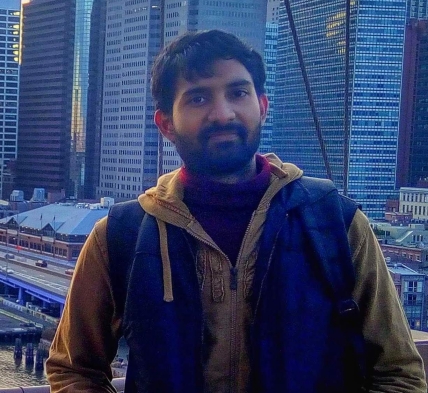}}]{Kuldeep Purohit}
Kuldeep Purohit is currently an AI Research Scientist at Phiar, California, USA. Earlier, he worked as a postdoctoral researcher at MSU, USA. He received his M.S. and Ph.D. degrees from the IIT Madras, India in 2020. His research interests include deep learning, image restoration and enhancement, computational imaging, and general image-to-image mapping tasks.
\end{IEEEbiography}

\begin{IEEEbiography}[{\includegraphics[width=1in,height=1.25in,clip,keepaspectratio]{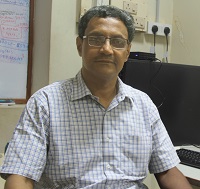}}]{
A. N. Rajagopalan:}
Dr. A.N. Rajagopalan received the Ph.D degree in Electrical Engineering from IIT Bombay. During the summer of 1998, he was a visiting scientist at the Image Communication Lab, University of Erlangen, Germany. From 1998 to 2000, he was a research faculty at the Center for Automation Research, University of Maryland. He subsequently joined the Department of Electrical Engineering, IIT Madras and currently serves as Sterlite Technologies Chair Professor. His areas of interest include deep learning, image restoration for deblurring, superresolution, inpainting and dehazing, 3D structure from blur, light field imaging, underwater imaging, image forensics, and multimodal learning. He was an Associate Editor for IEEE Transactions PAMI from 2007 to 2011, Associate Editor for IEEE Transactions on Image Processing from 2012 to 2016, and is serving as Senior Area Editor for IEEE Transactions on Image Processing since 2016. He has been Area Chair for CVPR 2012, 2022, ACCV 2022, and ICPR 2012, Program Co-chair for ICVGIP 2010, and General Chair for NCVPRIPG 2017 and 2019. He has co-authored two books. He is a Fellow of the Alexander von Humboldt Foundation, Germany, and a Fellow of the Indian National Academy of Engineering. He is a recipient of the DAE-SRC Outstanding Investigator award in 2012, the VASVIK award in 2013, the Mid-Career Research \& Development award from IIT Madras in 2014, Google India AI/ML Research award for faculty in 2018, and Qualcomm Faculty award in 2022.

\end{IEEEbiography}




\end{document}